\documentclass[final, 3p, authoryear]{elsarticle}

\usepackage{amsmath}
\usepackage{amssymb}
\usepackage{amsthm}
\usepackage{mathtools}

\usepackage{balance} 

\usepackage{diagbox} 

\usepackage{arydshln}
\usepackage{booktabs}

\usepackage{adjustbox}
\usepackage[ruled, linesnumbered]{algorithm2e}
\usepackage{bbm}
\usepackage{bm}
\usepackage{framed} 
\usepackage{graphicx}
\usepackage{graphics}
\usepackage{multirow}
\usepackage{float}
\usepackage{microtype}      
\usepackage{rotating}
\usepackage[subrefformat=parens,labelformat=parens]{subfig}
\usepackage{xcolor}
\usepackage[T1,hyphens]{url}

\usepackage{breakcites}
\usepackage[toc,page]{appendix}

\usepackage{hyperref}       

\usepackage[capitalize]{cleveref}
\crefname{section}{Sec.}{Secs.}
\Crefname{section}{Section}{Sections}
\Crefname{table}{Table}{Tables}
\crefname{table}{Tab.}{Tabs.}

\newcommand{\notes}[1]{\textcolor{red}{#1}}

\newtheorem{theorem}{Theorem}

\usepackage{amsthm}

\newtheorem{remark}{Remark}

\usepackage{enumitem}
\newlist{AL}{enumerate}{1}
\setlist[AL]{label=\textbf{A\arabic*}.}

\AtBeginDocument{%
  \providecommand\BibTeX{{%
    \normalfont B\kern-0.5em{\scshape i\kern-0.25em b}\kern-0.8em\TeX}}}

\allowdisplaybreaks 

\newcommand{\rev}[1]{\textcolor{black}{#1}}

\def\x{{\mathbf x}}

\usepackage{lineno}

\journal{Expert Systems with Applications}

\begin{document}
\date{}
\begin{frontmatter}



\title{Distilling and Transferring Knowledge via cGAN-generated Samples \\ for Image Classification and Regression} 

\author[contact1]{Xin Ding}
\ead{dingx92ubc@126.com}
\author[contact2]{Yongwei Wang\corref{cor1}}
\ead{yongweiw@ece.ubc.ca}
\author[contact1]{Zuheng Xu}
\ead{zuheng.xu@stat.ubc.ca}
\author[contact2]{Z. Jane Wang}
\ead{zjanew@ece.ubc.ca}
\author[contact1]{William J. Welch}
\ead{will@stat.ubc.ca}

\cortext[cor1]{Corresponding author.}

\address[contact1]{Department of Statistics, University of British Columbia, Vancouver, BC, V6T 1Z4, Canada\\}
\address[contact2]{Department
	of Electrical and Computer Engineering, University of British Columbia, Vancouver,
	BC, V6T 1Z4, Canada\\}

\begin{abstract}

\textit{Knowledge distillation} (KD) has been actively studied for image classification tasks in deep learning, aiming to improve the performance of a student model based on the knowledge from a teacher model. However, applying KD in image regression with a scalar response variable is also important (e.g., age estimation) yet has been rarely studied. Besides, existing KD methods often require a practitioner to carefully select or adjust the teacher and student architectures, making these methods less flexible in practice. To address the above problems in a unified way, we propose a comprehensive KD framework based on \textit{conditional generative adversarial networks} (cGANs), termed cGAN-KD. Fundamentally different from existing KD methods, cGAN-KD distills and transfers knowledge from a teacher model to a student model via specifically processed cGAN-generated samples. This novel mechanism makes cGAN-KD suitable for both classification and regression tasks, compatible with other KD methods, and insensitive to the teacher and student architectures. An error bound for a student model trained in the cGAN-KD framework is derived in this work, providing a theory for why cGAN-KD is effective as well as guiding the practical implementation of cGAN-KD. Extensive experiments on CIFAR-100 and ImageNet-100 (a subset of ImageNet with only 100 classes) datasets show that the cGAN-KD framework can leverage state-of-the-art KD methods to yield a new state of the art. Moreover, experiments on Steering Angle and UTKFace datasets demonstrate the effectiveness of cGAN-KD in image regression tasks. Notably, in classification, incorporating cGAN-KD into training improves the state-of-the-art SSKD by an average of 1.32\% in test accuracy on ImageNet-100 across five different teacher-student pairs. In regression, cGAN-KD decreases the test mean absolute error of a WRN16$\times$1 student model from 5.74 to 1.79 degrees (i.e., $68.82\%$ drop) on Steering Angle.

\end{abstract}


\end{frontmatter}

\section{Introduction}\label{sec:introduction}

The high precision of a heavyweight learning model, often a deep, overparameterized neural network or an ensemble of multiple deep neural networks, usually comes with associated costs.  First, the large model size (i.e., many learnable parameters) implies a high memory cost.  Secondly, the complexity leads to a low inference speed, i.e., the number of images processed per second. Thus, limited computational resources to evaluate a trained model (e.g., deploying neural networks on mobile devices), require the deployment of a lightweight model that is memory-efficient and fast in inference. The \textit{small model capacity} of a lightweight model would often lead to lower precision, however, motivating recent increased attention on leveraging an accurate heavyweight model to improve the performance of a lightweight model.

\textit{Knowledge distillation} (KD), first proposed by \citet{bucilua2006model} and then developed by \citet{hinton2015distilling}, is a popular method to improve the performance of a lightweight model by utilizing the knowledge distilled from an accurate, heavyweight model. The heavyweight and lightweight models in KD are often known respectively as a \textit{teacher} model and a \textit{student} model. After \citet{hinton2015distilling} introduced \textit{baseline knowledge distillation} (BLKD), many KD methods have been proposed for the image classification task \citep{wang2021knowledge, gou2021knowledge}. These methods can be categorized as logit-based KD \citep{hinton2015distilling, mirzadeh2020improved}, feature-based KD \citep{romero2014fitnets,zagoruyko2016paying, kim2018paraphrasing,ahn2019variational, heo2019knowledge,chen2021distilling, chen2021cross,wang2022semckd, chen2022knowledge}, relation-based KD \citep{passalis2018learning,tung2019similarity, park2019relational}, self-supervised KD \citep{tian2019contrastive, xu2020knowledge}, etc. To transfer knowledge, these KD methods often need to define new loss functions or design new auxiliary training tasks (e.g., the self-supervised learning task). Consequently, we often need to carefully choose teacher models or adjust the network architectures of the teacher and student models, making the implementation of these KD methods complicated.

Unlike the image classification task, the application of KD in image regression with a scalar response variable (e.g., the angle and age applications studied in Section~\ref{sec:cGAN-KD_experiment}) has rarely been studied. \citet{zhao2020distilling} propose a KD method specially designed to estimate ages from human face images. However, this method does not apply to general image regression tasks with a scalar response because some techniques of the proposed framework are only applicable for age estimation. \citet{saputra2019distilling} propose a KD framework to transfer knowledge from a large pose estimation network to a small one. However, the response variable in pose estimation is multivariate which distinguishes from a scalar variable in isometric characteristics. Besides, the proposed KD method in \citet{saputra2019distilling} is only applicable to some specific network architectures. There is no practical KD method general enough for image regression tasks with a scalar response to our best knowledge. Moreover, all the above methods are designed specifically for either image classification or image regression; there exists no unified KD framework suitable for both tasks yet.

\textit{Generative adversarial networks} (GANs) are state-of-the-art generative models for image synthesis \citep{goodfellow2014generative, mirza2014conditional, odena2017conditional, miyato2018spectral, zhang2019self, brock2018large, miyato2018cgans, karras2019style, karras2020analyzing, ding2021ccgan, ding2020continuous, zhou2021survey, li2021theoretical}. Some modern GAN models such as BigGAN \citep{brock2018large} and StyleGAN \citep{karras2019style, karras2020analyzing} are able to generate high-resolution, even photo-realistic images. \textit{Conditional generative adversarial networks} (cGANs) are a type of GANs that can generate images in terms of certain conditions.  Most cGANs are designed for categorical conditions such as class labels \citep{mirza2014conditional, miyato2018cgans, odena2017conditional, miyato2018spectral, zhang2019self, brock2018large, xu2021conditional}, and cGANs with class labels as conditions are also known as \textit{class-conditional GANs}. Recently, \citet{ding2021ccgan, ding2020continuous} propose a new cGAN framework, termed \textit{continuous conditional GANs} (CcGANs). CcGANs can generate images conditional on continuous, scalar variables (termed \textit{regression labels}). In the scenario with limited training data, the performance of GANs often deteriorates. To alleviate this problem for unconditional GANs and class-conditional GANs, DiffAugment \citep{zhao2020differentiable} proposes to conduct online transformation on images during the GAN training. Our experiments show that it also applies to CcGANs. Besides the advances in GAN theory, \citet{frid2018synthetic}, \citet{sixt2018rendergan}, \citet{wu2018conditional}, \citet{zhu2018emotion}, \citet{mariani2018bagan}, and \citet{ali2019mfc} use GAN-generated data for data augmentation in image classification tasks with insufficient training data. However, even state-of-the-art GANs may generate low-quality samples, which may negatively affect the classification task. Fortunately, some recently proposed subsampling methods \citep{ding2020subsampling, ding2021subsampling} may be applied to eliminate these low-quality samples. Additionally, some works \citep{xu2018training, wang2018kdgan, shen2019meal, liu2020ktan} propose to incorporate the adversarial loss into KD, but their performance is not state of the art.

Motivated by the limitations of existing KD methods and the recent advances of cGANs, we propose a general and flexible KD framework applicable for both image classification and regression (with a scalar response). Our contributions can be summarized as follows:
{
\begin{itemize}[noitemsep,topsep=0pt,parsep=0pt,partopsep=0pt]
	\item In \Cref{sec:cGAN-KD_method}, we introduce the proposed cGAN-KD framework, which distills and transfers knowledge via cGAN-generated samples. Compared with other methods, cGAN-KD is a unified KD framework suitable for both classification and regression tasks (with a scalar response). It is also compatible with state-of-the-art KD methods, where cGAN-KD can be incorporated into these methods to reach a new state of the art. Moreover, cGAN-KD is insensitive to architectural differences between teacher and student networks.
	
	\item In \Cref{sec:cGAN-KD_theory}, we derive the error bound of a student model trained in the cGAN-KD framework, which not only helps us understand how cGAN-KD takes effect but also guides the implementation of cGAN-KD in practice. Such an analysis is often omitted in knowledge distillation papers. The error bound suggests that we should generate as many processed fake samples as possible and choose a teacher model with high precision. 
	
	\item In \Cref{sec:cGAN-KD_experiment}, we conduct extensive experiments on CIFAR-100 \citep{krizhevsky2009learning}, ImageNet-100 \citep{cao2017hashnet}, Steering Angle \citep{steeringangle, steeringangle2}, and UTKFace \citep{utkface} to demonstrate the effectiveness of the cGAN-KD framework over state-of-the-art KD methods in both classification and regression tasks. We carefully design an ablation study to investigate the influcence of different (sub-)modules of cGAN-KD. Several sensitivity analyses are also conducted to research the effects of cGAN-KD's hyper-parameters. 
\end{itemize}
}

\section{Related work}\label{sec:related_work}
\rev{In this section, we first provide a comprehensive review of representative knowledge distillation methods proposed for image classification and regression tasks. We then briefly introduce conditional generative adversarial networks (cGANs) and the subsampling techniques for cGANs.  }
\subsection{Knowledge distillation}\label{sec:cGAN-KD_related_KD}

\rev{
	\textbf{Logits-based KD.} Hinton et al. proposed an effective logits-based knowledge distillation method (aka BLKD) that transfers knowledge from the teacher model to the student model by matching the \textit{logits} (i.e., the output of the last layer in a neural network) between these two models. BLKD does not need to change the teacher and student models' architectures, and it has been widely applied in visual recognition applications, e.g., image classification and face recognition \citep{gou2021knowledge}. 
}

\begin{figure}[!ht]
	\centering
	\includegraphics[width=0.4\textwidth]{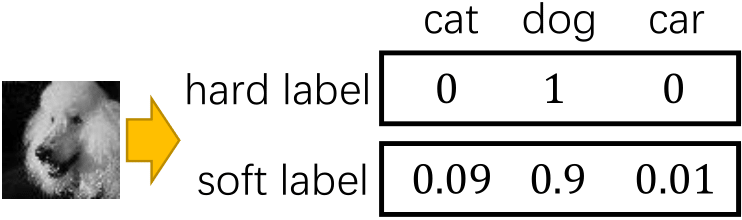}
	\caption{Example of hard and soft labels of a dog image in a 3-class classification task.}
	\label{fig:hard_vs_soft_label}
\end{figure}

Denote by $\bm{l}$ the logits of an image $\bm{x}$ from a neural network, where $\bm{l}=[l_1,\dots,l_C]^\intercal$ is a $C$ by 1 vector and $C$ is the number of classes. With softmax function, we can calculate the probability that the image $\bm{x}$ belonging to class $c$ as follows:
\begin{equation}
	\label{eq:cGAN-KD_softmax_function}
	p_c = \frac{\exp(l_c/T)}{\sum_{k=1}^C\exp(l_k/T)},
\end{equation}
where $c=1,\dots,C$ and $T$ is the temperature factor. The $C$ by 1 vector $\bm{p}=[p_1,\dots,p_C]^\intercal$ is also known as the \textit{soft label} of image $\bm{x}$. A higher $T$ leads to a softer probability distribution over classes. On the contrary, the one-hot encoded class label is also known as the \textit{hard label}. An example of hard labels and soft labels is shown in Fig.\ \ref{fig:hard_vs_soft_label}. Usually, the soft label is more informative than the hard label because it can reflect the similarity between classes and the confidence of prediction. The logits of the same image $\bm{x}$ from the teacher model $f_t$ and the student model $f_s$ are denoted by $\bm{l}^t$ and $\bm{l}^s$ respectively. Then, the corresponding soft labels are denoted respectively by $\bm{p}^t$ and $\bm{p}^s$. The student model $f_s$ is trained to minimize the cross entropy between $\bm{p}^t$ and $\bm{p}^s$ as follows:
\begin{equation}
	\label{eq:cGAN-KD_student_loss_KD}
	\mathcal{L}_{KD}=\textstyle\sum_{c=1}^C\{- p^t_c \log p^s_c \}.
\end{equation}
The student model is also trained to minimize the cross entropy between the one-hot encoded class label $y$ and the soft label $\bm{p}^s$ as follows:
\begin{equation}
	\label{eq:cGAN-KD_student_loss_standard}
	\mathcal{L}_{s}=\textstyle\sum_{c=1}^C\{- y_c \log p^s_c \}.
\end{equation}
Finally, the overall training loss of $f_s$ is a linear combination of Eqs.\ \eqref{eq:cGAN-KD_student_loss_KD} and \eqref{eq:cGAN-KD_student_loss_standard}, i.e.,
\begin{equation}
	\label{eq:cGAN-KD_student_loss}
	\mathcal{L}_{overall} = (1-\lambda_{KD}) \mathcal{L}_{s} + \lambda_{KD}\mathcal{L}_{KD},
\end{equation}
where $\lambda_{KD}\in[0,1]$ is a hyperparameter controlling the trade-off between two losses. $\mathcal{L}_{s}$ is the standard loss for classification and $\mathcal{L}_{KD}$ encourages the knowledge transfer.

\rev{Despite the simplicity and general effectiveness of BLKD, \citet{mirzadeh2020improved} recently show that BLKD may not perform well if there exists a big performance gap between a teacher model and a student model. To resolve this issue, the authors introduce a \textit{teacher assistant} (TA) model, which often performs better than the student model but worse than the teacher model. BLKD is applied to the teacher-TA and TA-student pairs, respectively, where the knowledge is first transferred from the teacher model to the TA model and then from the TA model to the student model.}

\rev{
	\textbf{Feature-based KD.} Instead of matching logits only, feature-based KD methods also encourage the student to mimic the teacher in terms of intermediate feature representations. FitNet \citep{romero2014fitnets} is the first work that proposes to utilize feature responses as knowledge hints. FitNet works by minimizing the feature map discrepancy in the middle feature level between a teacher network and a student network that is deeper and thinner than its teacher. Based on feature maps, the \textit{attention transfer} (AT) knowledge distillation \citep{zagoruyko2016paying} computes the activation-based and gradient-based attention maps, and transfers such attention knowledge to better guide the student model. Instead of directly utilizing or transforming feature maps, \citet{kim2018paraphrasing} propose the \textit{factorization transfer} (FT), which involves a paraphraser and a translator parameterized by convolutional modules. FT re-interprets and aligns the teacher's and student's feature responses. \textit{Variational information distillation} (VID) \citep{ahn2019variational} and \textit{activation boundary} (AB) \citep{heo2019knowledge} are another two novel feature-based KD methods that respectively maximize the mutual information and match the activation boundary between the teacher and student networks. To resolve the manual selection problem of intermediate layers between the teacher and student networks, \citet{chen2021cross} and \citet{wang2022semckd} propose semantic calibration for cross-layer knowledge distillation (SemCKD). SemCKD develops an attention mechanism to perform automatic layer association assignments. Recently, ReviewKD \citep{chen2021distilling} develops a knowledge review framework which explores the importance of low-level features in a teacher model. ReviewKD distills and transfers knowledge from multi-layers of a teacher model to supervise a student model. \textit{Simple knowledge distillation} (SimKD) \citep{chen2022knowledge} is another very recent feature-based KD method. SimKD reuses the discriminative classifier
	from the pre-trained teacher for student inference and minimizes a $\ell_2$ loss defined in the preceding layer of the final
	classifier.  While feature-based KD methods show much potential in knowledge transfer, the architectural differences between the student and teacher networks may hinder their distillation effectiveness in practice.  
}

\rev{
	\textbf{Relation-based KD.}	The relation-based KD models the relational knowledge based on responses of different training samples. \citet{passalis2018learning} propose a \textit{probabilistic knowledge transfer} (PKT) method. PKT establishes a probabilistic version of the relational matrix using pairwise neighboring samples, then aligns the student with the teacher by minimizing the Kullback-Leibler of conditional probability distribution (i.e., each row of a relational matrix). Based on an observation that semantically similar samples produce similar activation, \citet{tung2019similarity} propose the \textit{similarity preserving} (SP) knowledge distillation method. A similarity matrix is computed from a batch of training samples at the feature level, and then it is used as auxiliary information to guide the student's training. Similar to SP, \citet{park2019relational} propose the \textit{relational knowledge distillation} (RKD). RKD models the interplay between training samples by proposing a two-tuple distance-wise relation and a three-tuple angular-wise relation based on network embeddings. By capturing the structural relations among samples, the relation-based KD can distill additional informative knowledge and transfer it from the teacher to the student model.
}

\rev{
	\textbf{Self-supervision signal-guided KD.}
	Besides distilling representational knowledge from a regularly trained classifier, some recent works investigate the feasibility of extracting and transferring knowledge from self-supervision signals \citep{chen2020simple}. \citet{tian2019contrastive} propose a \textit{contrastive representation distillation} (CRD) method that formulates a contrastive objective for training the teacher and the student models. CRD pushes closer the representations of a teacher and a student for the same sample; while it pushes far away those from different samples. \citet{xu2020knowledge} propose \textit{self-supervised knowledge distillation} (SSKD). SSKD introduces the self-supervised learning scheme into knowledge distillation. Firstly, a pre-trained teacher model is tuned on a self-supervision pretask to learn generic representations in an unsupervised manner. Then, SSKD encourages the student model to mimic the teacher model in terms of both  classification outputs (like BLKD) and self-supervision predictions. 
 }

\rev{
	\textbf{Other related methods.} In addition to the state-of-the-art representative KD methods above, there are two other types of KD methods that have gained attention, i.e., GAN-related KD methods \citep{xu2018training, wang2018kdgan, shen2019meal, liu2020ktan} and data-free KD methods \citep{lopes2017data, chen2019data, yin2020dreaming}. However, there also exist major distinctions to our method though they seem related to our method. Our cGAN-KD is fundamentally different from the GAN-related KDs: (1) cGAN-KD utilizes cGAN-generated samples to distill and transfer knowledge, while GAN-related KDs only incorporate adversarial losses into conventional KD methods (e.g., \citet{hinton2015distilling}), and they cannot achieve the state-of-the-art performance.  (2) Our KD framework applies to both classification and regression tasks, while KD methods in GAN-related KDs can only apply to classification tasks. Our method also has quite different mechanisms to data-free KDs: (1) cGAN-KD uses a cGAN to synthesize fake samples instead of an inverted convolutional neural network; (2) cGAN-KD has a subsampling module and a label adjustment module to push the distribution of fake image-label pairs close to the actual distribution, but data-free KDs do not have these crucial techniques; and (3) The student is trained on both real and fake samples in cGAN-KD while data-free KDs do not have access to real samples. 
}

\begin{figure}[!ht]
	\centering
	\includegraphics[width=0.7\textwidth]{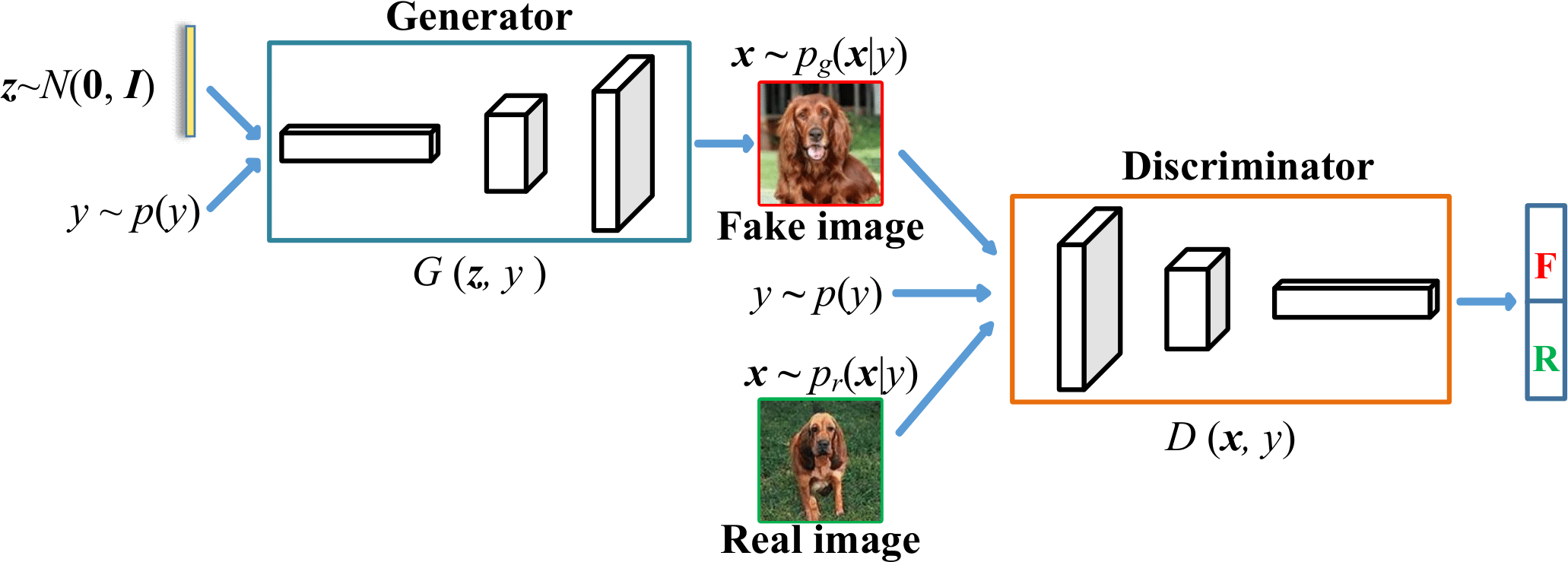} 
	\caption{\textbf{A typical pipeline of cGAN.} The conditioning variable $y$ (e.g., $y$ represents the class label or regression label) is assumed to follow a distribution $p(y)$, which can be easily estimated from the training data.}
	\label{fig:cGAN-KD_cGAN_illustration}
\end{figure}

\subsection{Conditional generative adversarial networks}\label{sec:related_cGANs}

cGANs \citep{mirza2014conditional} aim to estimate the distribution of images conditional on some auxiliary information. A cGAN model includes two neural networks, a generator $G(\bm{z},y)$ and a discriminator $D(\bm{x},y)$. The generator $G(\bm{z},y)$ takes as input a random noise $\bm{z}\sim N(\bm{0},\bm{I})$ and the condition $y$, and outputs a fake image $\bm{x}^g$ which follows the fake conditional image distribution $p_g(\bm{x}|y)$. The discriminator $D(\bm{x},y)$ takes as input an image $\bm{x}$ and the condition $y$, and outputs the probability that the image $\bm{x}$ comes from the true conditional image distribution $p_r(\bm{x}|y)$. A typical pipeline of cGAN is shown in Fig.\ \ref{fig:cGAN-KD_cGAN_illustration}. Mathematically, the cGAN model is trained to minimize the divergence between $p_r(\bm{x}|y)$ and $p_g(\bm{x}|y)$. The condition $y$ is usually a categorical variable such as a class label. cGANs with class labels as conditions are also known as \textit{class-conditional GANs} \citep{mirza2014conditional, miyato2018cgans, odena2017conditional, miyato2018spectral, zhang2019self, brock2018large}. Class-conditional GANs have been widely studied, and the state-of-the-art models such as BigGAN \citep{brock2018large} are already able to generate photo-realistic images. However, GANs conditional on regressions labels (e.g., angles and ages) have been rarely studied because of two problems. First, very few (even zero) images exist for some regression labels, so the empirical cGAN losses may fail. Second, since regression labels are continuous and infinitely many, they cannot be embedded by one-hot encoding like class labels. Recently, \citet{ding2021ccgan, ding2020continuous} propose a new formulation of cGANs, termed \textit{CcGANs}. The CcGAN framework consists of novel empirical cGAN losses and novel label input mechanisms. To solve the first problem, the discriminator is trained by either the \textit{hard vicinal discriminator loss} (HVDL) or the \textit{soft vicinal discriminator loss} (SVDL). A new empirical generator loss is also proposed to alleviate the first problem. To solve the second problem, \citet{ding2021ccgan, ding2020continuous} introduce a \textit{naive label input} (NLI) mechanism and an \textit{improved label input} (ILI) mechanism. Hence, \citet{ding2021ccgan, ding2020continuous} propose four CcGAN models employing different discriminator losses and label input mechanisms, i.e., HVDL+NLI, SVDL+NLI, HVDL+ILI, and SVDL+ILI. The effectiveness of CcGANs has been demonstrated on multiple regression-oriented datasets, e.g., Steering Angle \citep{ding2021ccgan, ding2020continuous} and UTKFace \citep{utkface}. 

The performance of cGANs often deteriorates when training data are insufficient. \textit{DiffAugment} \citep{zhao2020differentiable} is one of some recent works \citep{zhao2020differentiable, karras2020training, tran2020data, zhao2020image} that are designed to stabilize the cGAN training in this setting. Although DiffAugment is designed for unconditional (e.g., styleGAN \citep{karras2019style, karras2020analyzing}) and class-conditional GANs (e.g., BigGAN), our experiment shows that it is also applicable to CcGANs.

\subsection{cDR-RS: Subsampling cGANs}\label{sec:related_subsample_cGANs}

Modern cGANs are demonstrated successful in many applications, but low-quality samples still appear frequently even with state-of-the-art network architectures (e.g., BigGAN) and training setups. To filter out low-quality samples, \citet{ding2021subsampling} proposes a subsampling framework, termed \textit{cDR-RS}, for class-conditional GANs and CcGANs. This framework consists of two components: a \textit{conditional density ratio estimation} (cDRE) method termed \textit{cDRE-F-cSP} and a \textit{rejection sampling }(RS) scheme. cDRE-F-cSP aims to estimate the conditional density ratio function $r^*(\bm{x}|y)\coloneqq p_r(\bm{x}|y)/p_g(\bm{x}|y)$. It trains a density ratio (DR) model (a neural network) to approximate $r^*(\bm{x}|y)$ based on $N^r$ real images $\bm{x}^r_1,\bm{x}^r_2,\dots,\bm{x}^r_{N^r}\sim p_r(\bm{x}|y)$ and $N^g$ fake images $\bm{x}^g_1,\bm{x}^g_2,\dots,\bm{x}^g_{N^g}\sim p_g(\bm{x}|y)$. Based on the estimated conditional density ratios, the rejection sampling scheme is utilized to sample from a trained cGAN. For class-conditional GANs, \citet{ding2021subsampling} demonstrate that cDR-RS can substantially improve the \textit{Fr\'echet inception distance} (FID) \citep{heusel2017gans} and Intra-FID \citep{miyato2018cgans} scores. For CcGANs, cDR-RS not only improves the Intra-FID score but also improves the image diversity and label consistency (i.e., the consistency of generated images with respect to the conditioning label) \citep{ding2021ccgan, devries2019evaluation}.

\section{Proposed method}\label{sec:cGAN-KD_method}

\begin{figure}[!h]
	\centering
	\includegraphics[width=0.9\textwidth]{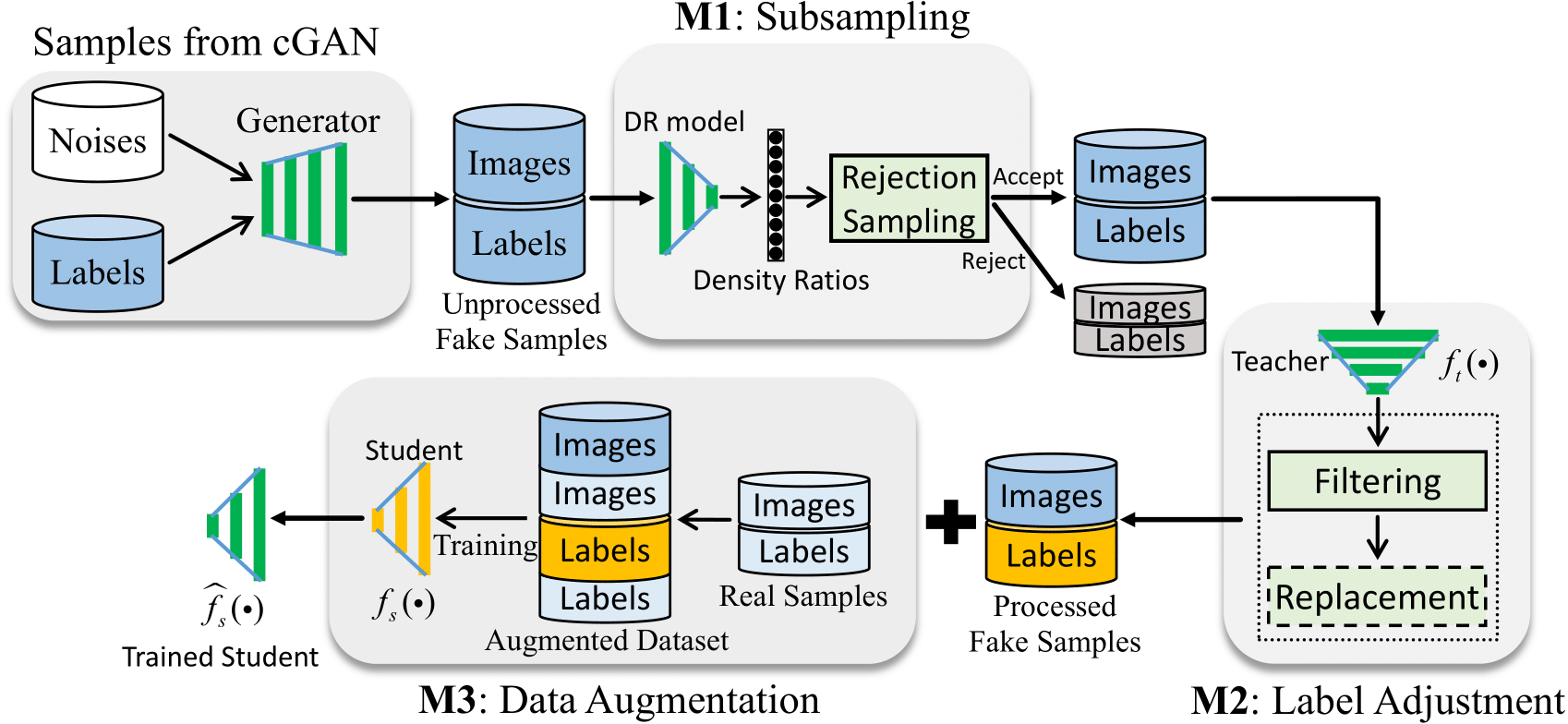}
	\caption{\rev{\textbf{The workflow of the proposed cGAN-KD.} Three important modules are denoted respectively by \textbf{M1}, \textbf{M2}, and \textbf{M3}. \textbf{M2} has two sequential sub-modules, the \textit{filtering} sub-module and the \textit{replacement} sub-module. The replacement sub-module is enabled for regression only. \textbf{M1} aims to drop low-quality fake samples. \textbf{M2} distills knowledge from teacher and embeds it into fake samples. As a side effect, the filtering sub-module in \textbf{M2} can also help drop visually unrealistic images. Processed fake samples are then used to do data augmentation in \textbf{M3}, where the knowledge from the teacher is transferred to the student.}}
	\label{fig:cGAN-KD_workflow}
\end{figure}

While many KD methods have been proposed for image classification, there is only one KD method for image regression (with a scalar response) \citep{zhao2020distilling}. Unfortunately, it is specially designed for age estimation with specific network architectures and is not applicable as a general KD method for image regression with a scalar response. Moreover, there is no KD framework for both types of tasks.

This section proposes a unified KD framework, termed \textit{cGAN-KD}, which is suitable for both image classification and regression (with a scalar response) tasks. The proposed framework can also fit into many state-of-the-art KD methods for image classification to improve their performances. In addition, we can blindly use the most precise heavyweight model as a teacher in cGAN-KD without worrying about the architectural difference between teacher and student.

To aid the reader, we summarize in \Cref{tab:definition_symbols} some essential notations with their definitions that appeared in this paper. These notations are also defined in detail near their first appearance.

\begin{table}[!h]
	\centering
	\caption{Definitions of some essential notations in this paper.}
	\begin{adjustbox}{width=1\textwidth}
		\begin{tabular}{c|l}
			\hline\hline
			Notation & Definition \\
			\hline
			$\bm{x}$ & an image at $C\times H\times W$ resolution and it may have a subscript, a superscript, or a tilde, e.g., $\tilde{\bm{x}}_i^g$  \\
			\hline
			$y$ & a class/regression label and it may have a subscript or superscript, e.g., $y_i^r$  \\
			\hline
			$\hat{y}$ & a predicted class/regression label and it may have subscript or superscript, e.g., $\hat{y}_i^r$.  \\
			\hline
			$p_r(\bm{x},y)$ & the density function of the joint distribution of real image-label pairs \\
			\hline
			$p_g(\bm{x},y)$ & the density function of the joint distribution of unprocessed fake image-label pairs  \\
			\hline
			$p_g^s(\bm{x},y)$ & the density function of the joint distribution of fake image-label pairs after applying \textbf{M1} \\
			\hline
			$\tilde{p}_g^\rho(\bm{x},y)$ & the density function of the joint distribution of fake image-label pairs after applying filtering \\
			\hline
			$\hat{p}_g^\rho(\bm{x},y)$ & the density function of the joint distribution of fake image-label pairs after applying replacement \\
			\hline
			$p_g^\rho(\bm{x},y)$ & the density function of the joint distribution of fake image-label pairs after applying \textbf{M2} \\
			\hline
			$D^r$ & a dataset of $N^r$ real image-label pairs  \\
			\hline
			$D^g$ & a dataset of infinite unprocessed fake image-label pairs  \\
			\hline
			$D^g_s$ & a dataset of $N^g$ fake image-label pairs after applying \textbf{M1}  \\
			\hline
			$\widetilde{D}^g_\rho$ & a dataset of $M^g$ fake image-label pairs after applying filtering \\
			\hline
			$\widehat{D}^g_\rho$ & a dataset of $M^g$ fake image-label pairs after applying replacement \\
			\hline
			$D^g_\rho$ & a dataset of $M^g$ fake image-label pairs after applying \textbf{M2}; it equals either $\widetilde{D}^g_\rho$ or $\widehat{D}^g_\rho$ \\
			\hline
			$D_{\text{aug}}$ & the augmented dataset in \textbf{M3}, i.e., $D_{\text{aug}} \coloneqq D^{r}\cup D_\rho^g$ \\
			\hline
			$\alpha$ & the filtering threshold in \textbf{M2} which is controlled by a hyper-parameter $\rho$ or a class label $c$ \\
			\hline
			$\rho$ & a hyper-parameter in $[0,1]$ to define the filtering threshold in \textbf{M2}, representing the $\rho$-th quantile of some errors \\
			\hline
			$\mathcal{L}(\cdot,\cdot)$ & a \textit{cross entropy} (CE) loss for classification or a \textit{squared error} (SE) loss for regression \\
			\hline
			$f$ & a predictor for classification or regression \\
			\hline
			$f^*$ & a theoretically optimal predictor \\
			\hline
			$f_t$ & the pre-trained teacher model in cGAN-KD \\
			\hline
			$f_s$ & the student model in cGAN-KD \\
			\hline
			$\mathcal{F}_s$ & the hypothesis space of $f_s$, i.e., a set of functions that can be represented by $f_s$ \\
			\hline
			$\widehat{\mathcal{R}}_{N^r+M^g}(\mathcal{F}_s)$  &  Empirical Rademacher complexity of $\mathcal{F}_s$ in terms of $N^r$ real samples and $M^g$ fake samples.  \\
			\hline
			$\mathcal{V}(f)$ & $\mathcal{V}(f) \coloneqq \mathbb{E}_{(\bm{x},y)\sim p_r(\bm{x},y)}\left[ \mathcal{L}(f(\bm{x}),y) \right]$ \\
			\hline
			$\widehat{\mathcal{V}}(f)$ & $\widehat{\mathcal{V}}(f) \coloneqq \frac{1}{N^r+M^g}\sum_{(\bm{x}_i,y_i)\in D_{\text{aug}}}\mathcal{L}(f(\bm{x}_i),y_i)$ \\
			\hline
			$f_s^{\circ}$ & the theoretical minimizer, i.e., $f_s^{\circ}=\arg\min_{f_s\in\mathcal{F}_s}\mathcal{V}(f_s)$ \\
			\hline
			$\hat{f}_s$ & the empirical minimizer, i.e., $\hat{f}_s=\arg\min_{f_s\in\mathcal{F}_s}\widehat{\mathcal{V}}(f_s)$ \\
			\hline\hline
		\end{tabular}%
	\end{adjustbox}
	\label{tab:definition_symbols}%
\end{table}%

\subsection{Problem formulation}\label{sec:cGAN-KD_problem_formulation}
Before we introduce cGAN-KD, let us formulate the KD task mathematically as follows. Assume we have a set of $N^r$ image-label pairs, i.e., 

\begin{equation*}
    D^r=\left\{ (\bm{x}_i^r,y_i^r) \ | \ i=1,\dots,N^r \right\},
\end{equation*}
which are randomly drawn from the actual image-label distribution with density function $p_r(\bm{x},y)$. We also have a teacher model $f_t$ and a student model $f_s$ which are trained on $D^r$. $f_t$ often has a smaller test error (i.e., higher precision) than $f_s$ does, i.e., 

\begin{equation*}
	\mathbb{E}_{(\bm{x},y)\sim p_r(\bm{x},y)}\mathcal{L}(f_t(\bm{x}),y)\leq \mathbb{E}_{(\bm{x},y)\sim p_r(\bm{x},y)}\mathcal{L}(f_s(\bm{x}),y),
\end{equation*}
where $\mathcal{L}$ is either the \textit{cross entropy} (CE) loss (i.e., Eq.\ \eqref{eq:cGAN-KD_student_loss_KD}) for classification or the \textit{squared error} (SE) loss for regression. The objective of KD is to reduce the test error of $f_s$ by using the knowledge learned by $f_t$.

\subsection{The workflow of cGAN-KD}\label{sec:cGAN-KD_overall_workflow}

As a preliminary of cGAN-KD, we need to train a cGAN on $D^r$. For image classification, we suggest adopting state-of-the-art class-conditional GANs such as BigGAN \citep{brock2018large}. For image regression with a scalar response, we should use CcGANs \citep{ding2021ccgan, ding2020continuous}. In the scenario with very few training data, we propose to apply DiffAugment \citep{zhao2020differentiable} to stabilize the cGAN training. 

After the cGAN training, the proposed KD framework can be applied. In Fig.\ \ref{fig:cGAN-KD_workflow}, we show the workflow of cGAN-KD, which includes three important modules denoted respectively by \textbf{M1} (\textit{subsampling}), \textbf{M2} (\textit{label adjustment}), and \textbf{M3} (\textit{data augmentation}). Firstly, we can sample infinitely many unprocessed fake samples from the trained cGAN, i.e.,

\begin{equation*}
    \nonumber
    D^g=\left\{ (\bm{x}_i^g,y_i^g) \ | \ i=1,\cdots,+\infty \right\},
\end{equation*}
where $(\bm{x}_i^g,y_i^g)$ is the $i$-th fake image-label pair. These fake samples are then subsampled by \textbf{M1} to drop low-quality ones and form a subset of $D^g$, i.e.,

\begin{equation*}
    D^g_s=\left\{ ({\bm{x}}_{(i)}^g,{y}_{(i)}^g) \ | \ ({\bm{x}}_{(i)}^g,{y}_{(i)}^g)\in D^g, \ i=1,\cdots,N^g \right\}.
\end{equation*}
The next module \textbf{M2} in the pipeline adjusts the labels of images in $D^g_s$ by a pre-trained, precise teacher model $f_t$. \textbf{M2} has two sequential sub-modules, the filtering sub-module and the replacement sub-module. The output of \textbf{M2} is a set of processed samples, i.e., 

\begin{equation*}
	D^g_\rho=
	\begin{cases}
		\widetilde{D}^g_\rho & \text{for classification,} \\
		\widehat{D}^g_\rho & \text{for regression.}
	\end{cases}
\end{equation*}
$\widetilde{D}^g_\rho$ and $\widehat{D}^g_\rho$ are defined respectively as follows:
 
\begin{equation*}
	\widetilde{D}^g_{\rho}=
	\begin{cases}
		\left\{ (\tilde{\bm{x}}_i^g,\tilde{y}_i^g) \ | \ (\tilde{\bm{x}}_i^g,\ \tilde{y}_i^g)\in D^g_s, CE(f_t(\tilde{\bm{x}}_i^g),\tilde{y}_i^g)\leq \alpha(\rho, \tilde{y}_i^g),\ i=1,\cdots,M^g \right\} & \text{for classification} \\
		\left\{ (\tilde{\bm{x}}_i^g,\tilde{y}_i^g) \ | \ (\tilde{\bm{x}}_i^g,\tilde{y}_i^g)\in D^g_s,  |f_t(\tilde{\bm{x}}_i^g)-\tilde{y}_i^g|\leq \alpha(\rho), \ i=1,\cdots,M^g \right\} & \text{for regression}
	\end{cases},
\end{equation*}
and
\begin{equation*}
	\widehat{D}^g_{\rho}=\left\{ (\tilde{\bm{x}}_i^g,\hat{y}_i^g) \ | \ (\tilde{\bm{x}}_i^g,\tilde{y}_i^g)\in \widetilde{D}^g_{\rho}, \hat{y}_i^g=f_t(\tilde{\bm{x}}_i^g), \ i=1,\cdots,M^g \right\},
\end{equation*}
where $CE(\cdot,\cdot)$ is the cross entropy loss, and $\alpha$ is a cut-off point defined in \Cref{sec:cGAN-KD_label_adjustment} and related to a positive hyper-parameter $\rho$ or a class label $\tilde{y}_i^g$. The variables in the parenthesis of $\alpha$ (i.e., $\rho$ and $\tilde{y}_i^g$) specify which quantity affects $\alpha$. The processed samples $D^g_\rho$ are then used to augment the training set $D^r$. Finally, \textbf{M3} trains the student model $f_s$ on the augmented training set $D^r\cup D^g_\rho$. The student model trained on $D^r\cup D^g_\rho$ is expected to perform better than the one trained on $D^r$. More details of the three modules are described in Sections \ref{sec:cGAN-KD_drop_samples} to \ref{sec:cGAN-KD_data_augmentation} and the evolution of fake sample datasets is shown in Fig.~S.4.13 in Supp.~S.4. Some visual illustrations and an ablation study of \textbf{M1}, \textbf{M2} and \textbf{M3} are also provided in Section 5.3.

\subsection{\textbf{M1}: Drop low-quality fake samples via subsampling}\label{sec:cGAN-KD_drop_samples}

Since low-quality samples may reduce prediction accuracy if they are used to augment the training set, \textbf{M1} (\textit{subsampling}) is adopted to drop these samples. The subsampling module implements cDR-RS \citep{ding2021subsampling} which performs rejection sampling to accept or reject a fake image-label pair $(\bm{x}^g,y^g)$ in terms of the density ratio of $\bm{x}^g$ conditioning on $y^g$. \citet{ding2021subsampling} shows that cDR-RS can effectively improve the overall image quality of both class-conditional GANs and CcGANs in the conditional image synthesis setting. Thus, cDR-RS is very suitable for dropping low-quality samples in the proposed cGAN-KD framework.

\subsection{\textbf{M2}: Distill knowledge via label adjustment}\label{sec:cGAN-KD_label_adjustment}

Assume that we have a fake image-label pair $(\tilde{\bm{x}}^g, \tilde{y}^g)$ generated from the previous module \textbf{M1}. The label $\tilde{y}^g$ is called the \textit{assigned label} of $\tilde{\bm{x}}^g$ in this paper. Please note that the assigned label $\tilde{y}^g$ may deviate from the \textit{actual label} of $\tilde{\bm{x}}^g$ (aka \textit{label inconsistency}) because of the imperfectness of cGAN's training and density ratio estimation. For example, CcGANs \citep{ding2021ccgan, ding2020continuous} can generate many fake facial images conditional on age 3, but some of them may actually come from the population of age 5. Besides the assigned and actual label, there is a third type of label for $\tilde{\bm{x}}^g$ called \textit{predicted label}. The predicted label denoted by $\hat{y}^g$ is defined as the prediction from the teacher model $f_t$ on $\tilde{\bm{x}}^g$, i.e., $\hat{y}^g=f_t(\tilde{\bm{x}}^g)$. The predicted label is assumed to be closer to the actual label than the assigned label, because discriminative learning (i.e., fitting $f_t$) is often easier than generative learning (i.e., fitting cGANs). Based on the assigned and predicted labels of $N^g$ fake images generated from the previous module \textbf{M1}, the two-stage module \textbf{M2} primarily aims to increase the label consistency of fake image-label pairs. \textbf{After applying \textbf{M2}, the relation between images and labels (``knowledge") learned by the teacher model $f_t$ is stored in the processed fake samples $D^g_\rho$.} Additionally, \textbf{M2} may also drop some unrealistic images so the overall visual quality can be further improved.

More specifically, \textbf{M2} includes two sequential sub-modules, a \textit{filtering} sub-module and a \textit{replacement} sub-module. The replacement sub-module is enabled only for regression problems. These two sub-modules are introduced as follows: \textbf{(1)} The \textit{filtering} sub-module computes the errors between assigned and predicted labels, and drops fake samples with errors larger than a cut-off point $\alpha$. Here, the errors are defined as the cross entropy (CE) loss for classification and mean absolute error (MAE) for regression. Two corresponding algorithms for classification and regression are summarized in \Cref{alg:cGAN-KD_filtering_module_classification} and \Cref{alg:cGAN-KD_filtering_module_regression}, respectively. The filtering threshold $\alpha$ equals the $\rho$-th quantile of fake samples' errors. A smaller $\rho$ implies that more samples are dropped. For classification, we compute one $\alpha$ for each class and conduct the filtering within each class. Differently, we have a global $\alpha$ to filter fake images with different labels in regression tasks. As for the selection of $\rho\in[0,1]$, we empirically suggest $\rho=0.9$ for classification and $\rho=0.7$ for regression. We recommend a smaller $\rho$ for regression because the label inconsistency problem is more severe for CcGANs than class-conditional GANs. After removing fake samples with large errors, the label consistency of the fake samples can be improved. Additionally, our empirical study also shows that a significant error between the assigned and predicted labels often implies poor visual quality of the corresponding fake images. Consequently, as a side effect, the filtering sub-module can also help improve the overall visual quality of fake images. \textbf{(2)} The subsequent \textit{replacement} sub-module is enabled for regression only. Similar to pseudo-labeling \citep{lee2013pseudo, arazo2020pseudo} in semi-supervised learning, it \textit{replaces} the assigned label $\tilde{y}^g$ with the predicted label $\hat{y}^g$. As shown in \citet{ding2021ccgan, ding2020continuous}, CcGANs often suffer from the label inconsistency problem, and the replacement sub-module can effectively alleviate this problem. This sub-module is not necessary for classification because most label-inconsistent samples have already been dropped after filtering.

\begin{algorithm}[!h]
	\footnotesize
	\SetAlgoLined
	
	\For{c from 1 to C}{
			Sample $N^g/C$ fake images from a trained cGAN via cDR-RS conditional on class $c$\;
        	Predict the labels of these fake images by the pre-trained teacher model $f_t$\;
        	Compute the \textbf{cross entropy (CE) loss} between the assigned and predicted labels. Note that, we use soft predicted labels (refer to \Cref{fig:hard_vs_soft_label}) to compute the error.\;
        	Sort these errors from smallest to largest and the $\rho$-th quantile of these errors is set as the filtering threshold (i.e., the cut-off point $\alpha(\rho, c)$) \;
        	Remove fake images pairs with errors larger than the filtering threshold $\alpha(\rho, c)$ \;
		}
	
	\caption{A filtering algorithm for classification with $C$ classes in \textbf{M2} with a hyper-parameter $\rho$ to adjust class labels and drop low-quality samples. We suggest $\rho=0.9$ for classification. Please note that, in classification, we calculate one filtering threshold $\alpha(\rho,c)$ per class and we conduct filtering within each class.}
	\label{alg:cGAN-KD_filtering_module_classification}
\end{algorithm}

\begin{algorithm}[!h]
	\footnotesize
	\SetAlgoLined
	Sample $N^g$ fake image-label pairs from a trained cGAN with cDR-RS\;
	Predict the labels of these fake images by the pre-trained teacher model $f_t$\;
	Compute the \textbf{mean absolute error (MAE)} between the assigned and predicted labels. \;
	Sort these errors from smallest to largest and the $\rho$-th quantile of these errors is set as the filtering threshold (i.e., the cut-off point $\alpha(\rho)$) \;
	Remove fake image-label pairs with errors larger than the filtering threshold $\alpha(\rho)$.
	
	\caption{A filtering algorithm for regression in \textbf{M2} with a hyper-parameter $\rho$ to adjust regression labels and drop low-quality samples. We suggest $\rho=0.7$ for regression. Unlike classification, in regression, we calculate a global filtering threshold $\alpha(\rho)$ to filter fake images with different labels.}
	\label{alg:cGAN-KD_filtering_module_regression}
\end{algorithm}

\subsection{\textbf{M3}: Transfer knowledge via data augmentation}\label{sec:cGAN-KD_data_augmentation}

The processed fake samples $D^g_\rho$ from the previous module are used to augment the original training set $D^{r}$, i.e., to give $D^r\cup D^g_\rho$. To transfer knowledge distilled from the pre-trained $f_t$, we train $f_s$ on the augmented dataset in \textbf{M3}. Please note that empirical studies in \Cref{sec:cGAN-KD_experiment} show that as $M^g$ increases, the test error of $f_s$ often does not stop decreasing until $M^g$ is larger than a certain threshold and then starts fluctuating over a small range. Since it is hard to obtain the optimal $M^g$ in practice and a hefty $M^g$ usually does not cause a significant adverse effect on precision, we suggest generating the maximum number of processed samples allowed by the computational budget. Two intuitive explanations for the effectiveness of such data augmentation are shown as follows:
\begin{itemize}[noitemsep,topsep=0pt,parsep=0pt,partopsep=0pt]
	\item As shown in Figs.\ 10 to 14 of \citet{brock2018large}, cGANs such as BigGAN can generate fake images that are unseen in the training set. In other words, cGANs may yield new information which may help improve the student's generalization performance. 
	\item Although cGANs may yield new information, the fake images from cGANs may be inconsistent with their assigned labels. We use the accurate teacher model to adjust the fake images' labels, so the image-label relation (``knowledge") learned by the teacher is stored in the fake samples. This learned knowledge is transferred to the student model via data augmentation.
\end{itemize}
	
Note that \textbf{M3} makes our method fundamentally different from many existing KD methods \citep{hinton2015distilling, mirzadeh2020improved, romero2014fitnets, zagoruyko2016paying, tung2019similarity, ahn2019variational, park2019relational, passalis2018learning, heo2019knowledge, kim2018paraphrasing, tian2019contrastive, xu2020knowledge, chen2021distilling, chen2021cross,wang2022semckd,chen2022knowledge}, because the distilled knowledge is transferred through samples instead of specially designed loss functions or training tasks. \textbf{M3} is also distinct from existing GAN-based data augmentation methods \citep{frid2018synthetic, sixt2018rendergan, wu2018conditional, zhu2018emotion, mariani2018bagan}, because these methods do not filter out unrealistic images or adjust the labels of label-inconsistent images. Consequently, these low-quality samples may cause negative effects on the supervised learning tasks, making these GAN-based data augmentation methods unstable.

\subsection{Advantages of cGAN-KD}\label{sec:cGAN-KD_advantages}

\subsubsection{A unified knowledge distillation framework for image classification and regression }\label{sec:cGAN-KD_unified_framework}

As the main advantage of cGAN-KD, all necessary steps in the workflow of cGAN-KD are applicable to both classification and regression (with a scalar response). \rev{Although some feature-based KD methods (e.g., FitNet \citep{romero2014fitnets}, AT \citep{zagoruyko2016paying}, and RKD \citep{park2019relational}) may be adjusted to fit the regression task by removing their logit-related components, empirical studies in \Cref{sec:exp_regression} show that such modification fails to result in stable performance, and it sometimes even makes students perform worse.} Furthermore, the theoretical analysis of cGAN-KD (see \Cref{sec:cGAN-KD_theory}) also has the same general formulation for both tasks. Thus, cGAN-KD is actually a unified KD framework.

\subsubsection{Compatibility with state-of-the-art KD methods}\label{sec:cGAN-KD_compatibility}

cGAN-KD distills and transfers knowledge via fake samples, and it does not require extra loss functions or network architecture changes. Thus, cGAN-KD can be combined with many state-of-the-art KD methods for image classification to improve their performance. To embed a state-of-the-art KD method into cGAN-KD, we just need to train the student model on the augmented training set with this KD method in \textbf{M3} but keep other procedures in \Cref{fig:cGAN-KD_workflow} unchanged.

\subsubsection{Architecture-invariance}\label{sec:cGAN-KD_arch_agnostic}

As shown by our experiments in \Cref{sec:cGAN-KD_experiment} and some papers \citep{mirzadeh2020improved, xu2020knowledge}, the architecture differences between a teacher model and a student model may influence the performance of some existing KD methods because these methods rely on logits or intermediate layers to transfer knowledge. Other KD methods such as SSKD \citep{xu2020knowledge} and SimKD \citep{chen2022knowledge} even require some adjustments to the teacher and student models' architectures. Conversely, since the proposed cGAN-KD framework distills and transfers knowledge via fake samples, there are no restrictions on the teacher and student models' architectures. The theoretical analysis in \Cref{sec:cGAN-KD_theory} also tells us to choose the most accurate teacher model for the label adjustment without worrying about the architecture differences, making cGAN-KD more flexible than other KD methods.

\section{Error bound of cGAN-KD}\label{sec:cGAN-KD_theory}

This section derives an error bound of cGAN-KD, reflecting the distance of a student model trained by cGAN-KD from the theoretically optimal predictor. This theoretical analysis illustrates how the cGAN-KD framework improves the precision of $f_s$, and it also helps guide our implementation of cGAN-KD in practice.

Before we move to the theoretical analysis, we first introduce some notations. Denote by $p_g(\bm{x},y)$ the density function of the distribution of unprocessed fake samples. Denote by $p^s_g(\bm{x},y)$ and $p^\rho_g(\bm{x},y)$ the density functions of the distributions of fake samples that are processed by \textbf{M1} and \textbf{M2} respectively. The evolution of fake data and their distributions is visualized in Fig.~S.4.13.. Additionally, we denote the augmented training dataset in \textbf{M3} as $D_{\text{aug}}$, i.e., $D_{\text{aug}} \coloneqq D^{r}\cup D_\rho^g$. Then, we define a theoretical loss of a predictor $f$ and its empirical approximation based on $D_{\text{aug}}$,

\begin{equation*}
	\begin{aligned}
		\mathcal{V}(f) & \coloneqq \mathbb{E}_{(\bm{x},y)\sim p_r(\bm{x},y)}\left[ \mathcal{L}(f(\bm{x}),y) \right],\\
		\widehat{\mathcal{V}}(f) & \coloneqq \frac{1}{N^r+M^g}\sum_{(\bm{x}_i,y_i)\in D_{\text{aug}}}\mathcal{L}(f(\bm{x}_i),y_i),
	\end{aligned}
\end{equation*} 
where $\mathcal{L}$ is either the CE loss for classification or the SE loss for regression. Without loss of generality, we assume $y\in[0,1]$ in regression tasks. Let $f^*$ be the \textit{optimal predictor} which minimizes $\mathcal{V}(f)$. We denote by $\mathcal{F}_s$ the \textit{hypothesis space} of $f_s$, i.e., a set of functions that can be represented by $f_s$. Note that $\mathcal{F}_s$ may not include $f^*$. Then, we define $f_s^{\circ}$ and $\hat{f}_s$ as

\begin{equation*}
	f_s^{\circ}=\arg\min_{f_s\in\mathcal{F}_s}\mathcal{V}(f_s),\quad \hat{f}_s=\arg\min_{f_s\in\mathcal{F}_s}\widehat{\mathcal{V}}(f_s).
\end{equation*}

\begin{figure}[!htbp]
	\centering
	\includegraphics[width=0.6\textwidth]{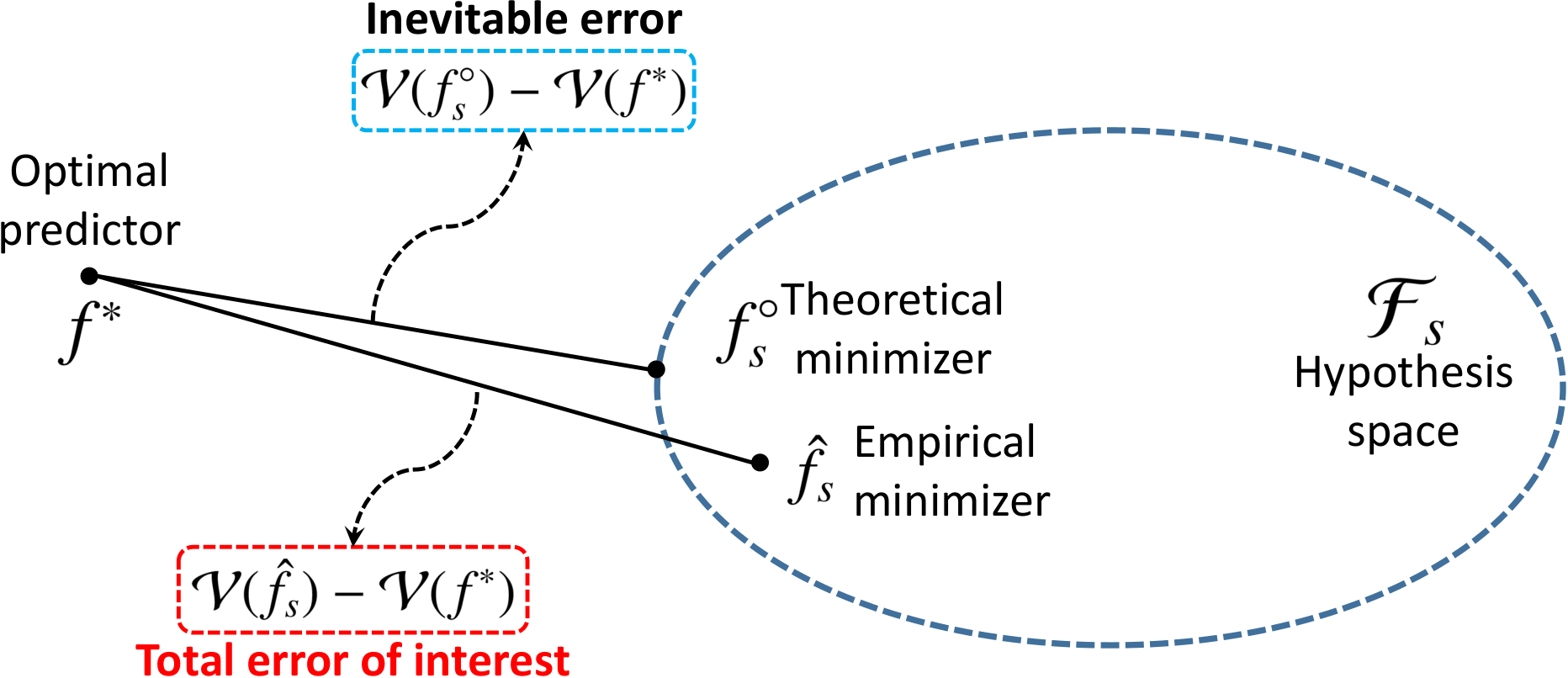}
	\caption{\textbf{An illustration of the error bound of cGAN-KD.} $\hat{f}_s$ is the minimizer we can obtain in practice, and we are interested in how far it deviates from the optimal predictor $f^*$, i.e., $\mathcal{V}(\hat{f}_s)-\mathcal{V}(f^*)$. We derive an upper bound of $\mathcal{V}(\hat{f}_s)-\mathcal{V}(f^*)$ in \Cref{thm:cGAN-KD_error_bound}. Please note that $\mathcal{V}(f_s^{\circ})-\mathcal{V}(f^*)$ is an inevitable error, because the hypothesis space may not cover $f^*$.  }
	\label{fig:illustrative_error_bound}
\end{figure}

If the architecture of the student model $f_s$ is determined, then the hypothesis space $\mathcal{F}_s$ is fixed. This hypothesis space may not cover the theoretically optimal predictor $f^*$. In this case, the training of the student model aims to minimize $\mathcal{V}(f_s)$ with respect to $f_s\in\mathcal{F}_s$, i.e., to get $f_s^{\circ}$. Unfortunately, obtaining $f_s^{\circ}$ is also inaccessible because the density function $p_r(\bm{x},y)$ is unknown and the expectation in $\mathcal{V}(f_s)$ is intractable. In the cGAN-KD framework, we approximate $\mathcal{V}(f_s)$ by $\widehat{\mathcal{V}}(f_s)$, and the minimizer we actually get in practice is $\hat{f}_s$. Therefore, \textbf{we are interested in how far $\hat{f}_s$ deviates from $f^*$}. We characterize this error via the theoretical loss $\mathcal{V}(f)$, i.e., $\mathcal{V}(\hat{f}_s)-\mathcal{V}(f^*)$. If $\hat{f}_s=f^*$, this error equals zero. An illustrative figure for this error is shown in \cref{fig:illustrative_error_bound}. Instead of providing an analytical form of this error, we derive its upper bound (i.e., error bound) in the form of a concentration inequality (refer to Theorem \ref{thm:cGAN-KD_error_bound}).

\begin{theorem}[Error Bound]
	\label{thm:cGAN-KD_error_bound}
	Suppose that 
	\begin{AL}
		\item (i.i.d. samples) $D^r\overset{i.i.d.}{\sim}p_r(\bm{x},y)$, $D^g_\rho\overset{i.i.d.}{\sim}p^\rho_g(\bm{x},y)$, and the augmented dataset is considered as i.i.d. samples from a mixture distribution, i.e., 
		
		\begin{equation}
			\label{eq:cGAN-KD_D_aug}
			D_{\text{aug}} \overset{i.i.d.}{\sim} \theta p_r(\bm{x},y) + (1-\theta) p^\rho_g(\bm{x},y) \eqqcolon p_{\theta}(\bm{x},y),
		\end{equation}
		where $\theta\in[0,1]$.

		\item (Measurability) $f_s$ is measurable for all $f_s\in\mathcal{F}_s$.

		\item (Distribution gap) There is a constant $C_{M1}>0$ such that 
		
		\begin{equation}
			\label{eq:cGAN-KD_W1_pr_pg}
			TV(p_r, p^\rho_g) = C_{M1} + \Theta(\mathbb{E}_{(\bm{x},y)\sim p_r(\bm{x},y)}\left[ \mathcal{L}(f_t(\bm{x}), y) \right]),
		\end{equation}
		where $TV$ denotes the total variation distance \citep{gibbs2002choosing} between the probability distributions of real samples and processed fake samples (with density functions $p_r(\bm{x},y)$ and $p^\rho_g(\bm{x},y)$ respectively); and $f(x) = \Theta(g(x))$ means  $f(x) = O(g(x))$ and $g(x) = O(f(x))$.

		\item (Boundedness) There exists a constant $0<C_{\mathcal{L}}<+\infty$, such that $\forall (\bm{x},y)$, $\mathcal{L}(f_s(\bm{x}),y)\leq C_{\mathcal{L}}$.
		
	\end{AL}
	Then, $\forall \delta\in(0,1)$, with probability at least $1-\delta$,
	\begin{align}
		& \mathcal{V}(\hat{f}_s)-\mathcal{V}(f^*) \nonumber\\
		\leq & 4 C_{\mathcal{L}}\widehat{\mathcal{R}}_{N^r+M^g}(\mathcal{F}_s) + 2 C_{\mathcal{L}}\sqrt{\frac{4}{N^r+M^g}\log\left( \frac{2}{\delta} \right)}\nonumber\\
		& + 4C_{\mathcal{L}}\left(1-\theta\right) \left(C_{M1}+\Theta\left(\mathbb{E}_{(\bm{x},y)\sim p_r(\bm{x},y)}\left[ \mathcal{L}(f_t(\bm{x}),y) \right]\right)\right) \nonumber\\
		& + \left( \mathcal{V}(f_s^\circ) - \mathcal{V}(f^*) \right),\label{eq:cGAN-KD_error_bound}
	\end{align}
	where $\widehat{\mathcal{R}}_{N^r+M^g}(\mathcal{F}_s)$ stands for the empirical Rademacher complexity \citep[Definition 3.1]{mohri2018foundations} of $\mathcal{F}_s$, which is defined on $N^r+M^g$ samples independently drawn from $p_{\theta}$. 
\end{theorem}	
	
	\begin{proof}
		The proof is in Supp.\ \ref{supp:proof_cGAN-KD_error_bound}.
	\end{proof}

	\begin{remark}[Rationale for \textbf{A3} and \textbf{A4}]
		\label{rmk:cGAN-KD_rationality_of_assumptions}
		In the cGAN-KD framework, processed fake images are used to augment the training set, so the distribution gap between $p_r(\bm{x},y)$ and $p_g^\rho(\bm{x},y)$ (measured by the total variation distance) should have a significant impact on the student model's performance. Thus, in \textbf{A3} of Theorem \ref{thm:cGAN-KD_error_bound}, we model the distribution gap by the summation of two components. The first component $C_{M1}$ stands for the divergence caused by the trained cGAN and the subsampling module. The second component is controlled by the generalization performance of $f_t$---the expected loss of the trained teacher model over the true data distribution.
		
		It is also worth discussing the rationale for \textbf{A4}. The two types of learning tasks considered in this work are the regression and classification, for which we use the squared loss $(f_s(\bm x)- y)^2$ and the cross entropy loss $\textstyle\sum_{c=1}^C\{- y_c \log p^s_c \}$ respectively. Take the regression task first. In our experiments on the regression datasets, the last layer of $f_s(\bm x)$ is the ReLU activation function \citep{fukushima1969visual,fukushima1982neocognitron}, so $f_s(\bm x)\geq 0$. Since $y\in (0,1)$, as long as $f_s(\bm x)$ is not an unstable predictor, it should not output arbitrarily large values, which implies $f_s(\bm x)$ can be bounded by a positive constant. Therefore, the squared loss is bounded and \textbf{A4} is satisfied. For the classification task, a sufficient condition to \textbf{A4} is that $p^s_c\geq\epsilon>0$ when $y_c=1$, representing that our classifier cannot produce $0$ probability for the true label, which is reasonable in practice.
	\end{remark}

	\begin{remark}[Illustration of Theorem \ref{thm:cGAN-KD_error_bound}]
		The four terms on the right side of Eq.\ \eqref{eq:cGAN-KD_error_bound} show that the error of $f_s$ has four components, and reducing them can improve the performance of $f_s$. The first and last terms are only relevant to the nature of $f_s$, so they are not influenced by $f_t$. If $f_s$ does not output arbitrarily extreme predictions (as discussed in Remark \ref{rmk:cGAN-KD_rationality_of_assumptions}), $C_{\mathcal{L}}$ stays at a moderate level, implying the first term is also small. The last term is inevitable because $\mathcal{F}_s$ may not include $f^*$. The second term diminishes if we set $M^g$ large. For the third term, $\theta\rightarrow 0$ as $M^g$ increases. Then, the third term is only controlled by the properties of $\mathcal{L}$ and the distribution gap. To reduce the distribution gap, we can either improve the cGAN model and the subsampling method, or choose $f_t$ to have better generalization performance. 
		
		
		Therefore, Theorem \ref{thm:cGAN-KD_error_bound}  implies when implementing cGAN-KD we should
		\begin{itemize}[noitemsep,topsep=0pt,parsep=0pt,partopsep=0pt]
		    \item use state-of-the-art cGANs and subsampling methods;
		    \item set $M^g$ large;
		    \item choose a teacher $f_t$ with the highest precision as possible.
		\end{itemize}
		
	\end{remark}

\section{Experiments}\label{sec:cGAN-KD_experiment}

{

This section aims to experimentally demonstrate the effectiveness of the proposed cGAN-KD framework in image classification and regression (with a scalar response) tasks. We conduct extensive experiments on four image datasets: CIFAR-100 \citep{krizhevsky2009learning} and ImageNet-100 \citep{cao2017hashnet} for image classification; and Steering Angle \citep{steeringangle, steeringangle2} and UTKFace \citep{utkface} for image regression. We compare the cGAN-KD framework against different types of KD methods on CIFAR-100 and ImageNet-100 (\Cref{sec:exp_classification}). Since the logits of teachers and students are unavailable in regression tasks, we only show the effectiveness of the cGAN-KD framework over some modified feature-based KD methods on Steering Angle and UTKFace (\Cref{sec:exp_regression}). In additional to these experiments, we conduct an ablation study to test the effects of different (sub-)modules in the cGAN-KD framework (\Cref{sec:exp_ablation}). Finally, some additional analyses about hyper-parameters, selection of teacher models, and running time and memory cost are also provided in Section 5.4. Please note that for image regression, as suggested by \citet{ding2021ccgan, ding2020continuous}, when training CcGANs, $f_t$, and $f_s$, regression labels are normalized to real numbers in $[0,1]$. Nevertheless, in the evaluation stage of $f_t$ and $f_s$, we compute \textit{mean absolute error} (MAE) on unnormalized regression labels.

}

\subsection{Classification: CIFAR-100 and ImageNet-100}\label{sec:exp_classification}

\subsubsection{Datasets}\label{sec:cls_datasets}
For classification, we conduct experiments on two datasets: CIFAR-100 and ImageNet-100. CIFAR-100 consists of 60,000 RGB images at $32\times32$ resolution uniformly spread across 100 classes. The overall number of training samples is 50,000 (500 for each class), and the remaining 10,000 samples (100 for each class) are for testing. ImageNet-100, as a subset of ImageNet \citep{imagenet_cvpr09}, has 128,503 RGB images at $128\times 128$ resolution from 100 classes. In our experiment, we randomly split ImageNet-100 into a training set and a test set, where 10,000 images are for testing (on average 100 images per class) and the rest are for training.

\subsubsection{The selection of teachers and students}\label{sec:cls_sel_tea_stu}
To select students and teachers for this experiment, some popular classifiers are trained from scratch on each dataset, and their Top-1 test accuracies are shown in Tables S.5.8 and S.6.11 in Appendix. Some light-weight neural networks (shown in Table 2 and Table 3) with low test accuracies are chosen as student models, and we aim to improve their performance. Some neural networks with high accuracies are chosen as the teacher models for the filtering step in cGAN-KD and other compared KD methods.

\subsubsection{Compared methods and implementation details}\label{sec:cls_candidates}

{

Compared KD methods for classification tasks are NOKD (i.e., no KD method is applied), BLKD \citep{hinton2015distilling}, FitNet \citep{romero2014fitnets}, AT \citep{zagoruyko2016paying}, PKT \citep{passalis2018learning}, FT \citep{kim2018paraphrasing}, SP \citep{tung2019similarity}, VID \citep{ahn2019variational}, RKD \citep{park2019relational}, AB \citep{heo2019knowledge}, CRD \citep{tian2019contrastive}, TAKD \citep{mirzadeh2020improved}, SSKD \citep{xu2020knowledge}, ReviewKD \citep{chen2021distilling}, SemCKD \citep{chen2021cross,wang2022semckd}, SimKD \citep{chen2022knowledge}, and the proposed cGAN-KD framework (including incorporating other KD methods into cGAN-KD, which is denoted by cGAN-KD + X).

On CIFAR-100, we implement BLKD, FitNet, AT, PKT, FT, SP, VID, RKD, AB, and CRD via \texttt{RepDistiller} (a GitHub repository provided by \citet{tian2019contrastive}).  ReviewKD, SemCKD, SimKD, and SSKD are implemented based on their official implementations. We implement TAKD based the provided algorithm in \citet{mirzadeh2020improved}, where $\lambda_{KD}=0.5$ and $T=5$ \citep{ruffy2019state}. Note that most of these KD methods do not support DenseNet \citep{huang2017densely} as the teacher due to its densely connection, but cGAN-KD and BLKD still work well. For ImageNet-100, compared with the CIFAR-100 experiment, we test fewer KD methods and teacher-student pairs due to limited computational resources. 

To implement the proposed cGAN-KD framework, we train two BigGAN models \citep{brock2018large} for CIFAR-100 and ImageNet-100, respectively. DiffAugment \citep{zhao2020differentiable} is also incorporated into the BigGAN training to improve the training stability. DenseNet121 and DenseNet161 \citep{huang2017densely} are chosen as the teacher models for filtering when implementing cGAN-KD due to their highest precision on CIFAR-100 and ImageNet-100, respectively. As we suggested in \Cref{sec:cGAN-KD_label_adjustment}, we let $\rho=0.9$ on both datasets. We generate $M^g=100,000$ processed fake samples for each dataset. Please note that we only combine cGAN-KD with six representative KD methods (i.e., BLKD, FitNet, VID, RKD, CRD, and SSKD) due to limited computational resources. When combining cGAN-KD with chosen KD methods (e.g., cGAN-KD+BLKD in \Cref{tab:cifar_KD_sim_results}), we have two types of teacher models: \textit{the primary teacher} and \textit{the secondary teacher}. The primary teacher is for filtering in cGAN-KD which is fixed as DenseNet121 (for CIFAR-100) or DenseNet161 (ImageNet-100), and the secondary teacher is for the implementation of existing KD methods.

All results of the CIFAR-100 experiment are reported in mean (standard deviation) \textbf{over 4 repetitions}. Differently, all results of the ImageNet-100 experiment are reported in \textbf{a single trial}. Other implementation details such as learning rate, batch size, weight decay, number of epochs, seed, and optimizer are shown in Sections S.5 and S.6 in Appendix.

}


\subsubsection{Experimental results}\label{sec:cls_results}

{
For CIFAR-100, Tables\ \ref{tab:cifar_KD_sim_results} and \ref{tab:cifar_KD_cross_results} show that among 21 teacher-student pairs, cGAN-KD-based methods perform the best in 18 pairs. For ImageNet-100, Table\ \ref{tab:imagenet_KD_results} shows that cGAN-KD+SSKD beats all compared methods in all teacher-student pairs.

From these results, we can see that cGAN-KD alone can effectively improve all students' performance on both classification datasets (i.e., cGAN-KD versus NOKD); however, we highly recommend combining it with other existing KD methods, e.g., cGAN-KD+BLKD and  cGAN-KD+SSKD. In Table\ \ref{tab:imagenet_KD_results}, after being combined with cGAN-KD, BLKD performs comparably to or even better than the state-of-the-art SSKD on ImageNet-100. Notably, incorporating cGAN-KD into training improves the state-of-the-art SSKD by an average of 1.32\% in test accuracy on ImageNet-100 across five teacher-student pairs. 

Please note that, as one of the recently proposed KD methods, ReviewKD requires that teacher and student network architectures have the same number of stages, and the feature map dimensions of teacher and student at each stage are consistent. Unfortunately, many teacher-student combinations in our experiment don't satisfy this requirement. Consequently, these teacher-student pairs are not supported by ReviewKD, and their corresponding results are marked by NA. Furthermore, although SimKD performs the best in 3 teacher-student pairs on CIFAR-100, it causes negative effects on some students' performance when teachers are ResNet110 and ResNet50 (marked in red), implying its instability. Additionally, SemCKD does not converge when teacher and student are ResNet110 and ResNet20, so the corresponding results are marked by NA. In many teacher-student pairs, SemCKD is able to beat other feature-based methods, but it fails to outperform SSKD. In fact, our extensive experiments show that, among many existing KD methods, SSKD is the most effective and stable one.

%

}

\begin{table}[!h]
	\centering
	\caption{ {  \textbf{CIFAR-100: Average Top-1 test accuracy (\%) of compared KD methods (KD between similar architectures) with standard deviation after the ``$\pm$" symbol.} Bold and underline denote the best and the second best results, respectively.} }
	\begin{adjustbox}{width=0.8\textwidth}
		
		\begin{tabular}{l|cccccc}
			\hline\hline
			\multicolumn{1}{l|}{\textbf{Teacher}} & \multicolumn{1}{c}{\begin{tabular}[c]{@{}c@{}} ResNet110\\ {(73.27)}\end{tabular}} & \multicolumn{1}{c}{\begin{tabular}[c]{@{}c@{}} ResNet32x4\\ {(79.11)}\end{tabular}} & \multicolumn{1}{c}{\begin{tabular}[c]{@{}c@{}} VGG13\\ {(74.85)}\end{tabular}} & \multicolumn{1}{c}{\begin{tabular}[c]{@{}c@{}} VGG19\\ {(73.88)}\end{tabular}} & \multicolumn{1}{c}{\begin{tabular}[c]{@{}c@{}} WRN40$\times$2\\ {(75.82)}\end{tabular}} & \multicolumn{1}{c}{\begin{tabular}[c]{@{}c@{}} ResNet32x4\\ {(79.11)}\end{tabular}} \\
			\multicolumn{1}{l|}{\textbf{Student}} & \multicolumn{1}{c}{ResNet20} & \multicolumn{1}{c}{ResNet20} & \multicolumn{1}{c}{VGG8} & \multicolumn{1}{c}{VGG8} & \multicolumn{1}{c}{WRN40$\times$1} & \multicolumn{1}{c}{ResNet8x4} \\
			\hline 
			NOKD & ${69.22 \pm 0.39}$ & ${69.22 \pm 0.39}$ & ${70.61 \pm 0.44}$ & ${70.61 \pm 0.44}$ & ${71.40 \pm 0.11}$ & ${72.74 \pm 0.07}$ \\
			
			\hdashline
			
			BLKD (2015)  & ${70.44 \pm 0.26}$ & ${69.15 \pm 0.33}$ & ${72.93 \pm 0.19}$ & ${72.03 \pm 0.19}$ & ${73.45 \pm 0.19}$ & ${73.44 \pm 0.29}$ \\
			FitNet (2015) & ${69.99 \pm 0.19}$ & ${70.03 \pm 0.27}$ & ${73.46 \pm 0.27}$ & ${71.96 \pm 0.09}$ & ${74.09 \pm 0.21}$ & ${74.96 \pm 0.06}$ \\
			AT (2017)  & ${70.95 \pm 0.29}$ &  ${70.13 \pm 0.43}$ &  ${73.44 \pm 0.18}$ &  ${71.11 \pm 0.21}$ &  ${74.12 \pm 0.25}$ &  ${74.88 \pm 0.26}$ \\
			PKT (2018)  & ${70.91 \pm 0.08}$ &  ${69.74 \pm 0.30}$ &  ${73.60 \pm 0.37}$ &  ${72.45 \pm 0.20}$ &  ${73.87 \pm 0.46}$ &  ${74.48 \pm 0.26}$ \\
			FT (2018) & ${70.97 \pm 0.14}$ &  ${70.23 \pm 0.19}$ &  ${73.14 \pm 0.36}$ &  ${72.09 \pm 0.20}$ &  ${73.88 \pm 0.23}$ &  ${74.88 \pm 0.19}$ \\
			SP (2019)  & ${70.56 \pm 0.06}$ &  ${69.39 \pm 0.36}$ &  ${73.33 \pm 0.12}$ &  ${71.72 \pm 0.22}$ &  ${73.81 \pm 0.11}$ &  ${73.92 \pm 0.22}$ \\
			VID (2019)  & ${70.71 \pm 0.18}$ &  ${69.86 \pm 0.20}$ &  ${73.44 \pm 0.30}$ &  ${72.19 \pm 0.15}$ &  ${73.80 \pm 0.28}$ &  ${74.53 \pm 0.16}$ \\
			RKD (2019) & ${70.61 \pm 0.35}$ &  ${69.46 \pm 0.25}$ &  ${73.34 \pm 0.36}$ &  ${71.79 \pm 0.27}$ &  ${73.66 \pm 0.26}$ &  ${74.06 \pm 0.15}$ \\
			AB (2019)  & ${70.48 \pm 0.23}$ &  ${69.91 \pm 0.14}$ &  ${73.26 \pm 0.26}$ &  ${71.73 \pm 0.29}$ &  ${74.19 \pm 0.12}$ &  ${74.47 \pm 0.16}$ \\
			CRD (2019)  & ${71.39 \pm 0.17}$ &  ${70.41 \pm 0.16}$ &  ${73.88 \pm 0.26}$ &  ${72.59 \pm 0.13}$ &  ${74.33 \pm 0.37}$ &  ${75.25 \pm 0.16}$ \\
			TAKD (2020)  & ${70.75 \pm 0.32}$ & ${69.74 \pm 0.42}$ & ${73.23 \pm 0.31}$ & ${71.73 \pm 0.24}$ & ${74.49 \pm 0.28}$ & ${73.47 \pm 0.32}$ \\
			SSKD (2020)  & ${70.92 \pm 0.20}$ &  ${71.10 \pm 0.24}$ &  ${74.52 \pm 0.23}$ &  ${73.27 \pm 0.17}$ &  ${75.56 \pm 0.25}$ &  ${75.70 \pm 0.17}$ \\
			ReviewKD (2021) & ${70.97 \pm 0.11}$ &  {NA} &  ${74.00 \pm 0.13}$ &  ${72.84 \pm 0.42}$ &  ${75.27 \pm 0.11}$ &  ${75.55 \pm 0.27}$ \\
			SemCKD (2021) & {NA} &  ${70.16 \pm 0.47}$ &  ${74.01 \pm 0.17}$ &  ${73.13 \pm 0.09}$ &  ${73.87 \pm 0.14}$ &  ${75.62 \pm 0.22}$ \\
			SimKD (2022) & $ \notes{68.94 \pm 0.32}$ &  $\bm{72.33 \pm 0.13}$ &  ${74.51 \pm 0.14}$ &  ${73.28 \pm 0.24}$ &  ${75.56 \pm 0.09}$ &  $\bm{77.28 \pm 1.08}$ \\
			\hline 
			\textbf{cGAN-KD} & $70.71 \pm 0.10$ & $70.71 \pm 0.10$ & $72.08 \pm 0.21$ & $72.08 \pm 0.21$ & $73.19 \pm 0.48$ & $72.90 \pm 0.11$ \\
			\textbf{cGAN-KD + BLKD} & ${71.30 \pm 0.34}$ &  ${70.73 \pm 0.12}$ &  ${74.49 \pm 0.13}$ &  ${73.74 \pm 0.15}$ &  ${75.22 \pm 0.14}$ &  ${73.71 \pm 0.25}$ \\
			\textbf{cGAN-KD + FitNet} & ${70.66 \pm 0.30}$ &  ${71.30 \pm 0.36}$ &  ${74.62 \pm 0.09}$ &  $\underline{73.92 \pm 0.37}$ &  ${75.41 \pm 0.19}$ &  ${74.36 \pm 0.25}$ \\
			\textbf{cGAN-KD + VID} & ${70.80 \pm 0.24}$ &  ${71.25 \pm 0.17}$ &  ${74.81 \pm 0.07}$ &  ${73.89 \pm 0.15}$ &  ${75.07 \pm 0.26}$ &  ${74.59 \pm 0.24}$ \\
			\textbf{cGAN-KD + RKD} & $\underline{71.63 \pm 0.06}$ &  ${71.08 \pm 0.28}$ &  ${74.62 \pm 0.21}$ &  ${73.60 \pm 0.12}$ &  ${75.50 \pm 0.18}$ &  ${74.22 \pm 0.21}$ \\
			\textbf{cGAN-KD + CRD} & $\bm{72.24 \pm 0.11}$ &  ${71.75 \pm 0.19}$ &  $\bm{75.05 \pm 0.14}$ &  $\bm{74.07 \pm 0.08}$ &  $\bm{75.98 \pm 0.29}$ &  ${75.53 \pm 0.11}$ \\
			\textbf{cGAN-KD + SSKD} & ${71.26 \pm 0.26}$ &  $\underline{72.29 \pm 0.23}$ &  $\underline{74.91 \pm 0.09}$ &  ${73.59 \pm 0.21}$ &  $\underline{75.71 \pm 0.19}$ & $\underline{76.65 \pm 0.16}$ \\
			\hline\hline
		\end{tabular}%
	
	\end{adjustbox}

	\label{tab:cifar_KD_sim_results}%
\end{table}%

\begin{table}[!h]
	\centering
	\caption{ {  \textbf{CIFAR-100: Average Top-1 test accuracy (\%) of compared KD methods (KD between different architectures) with standard deviation after the ``$\pm$" symbol.} Bold and underline denote the best and the second best results, respectively. } }
	\begin{adjustbox}{width=1\textwidth}
		\begin{tabular}{l|ccccccccc}
			\hline \hline
			\textbf{Teacher} & \multicolumn{1}{c}{\begin{tabular}[c]{@{}c@{}} WRN40$\times$2\\ {(75.82)}\end{tabular}} & \multicolumn{1}{c}{\begin{tabular}[c]{@{}c@{}} WRN40$\times$2\\ {(75.82)}\end{tabular}} & \multicolumn{1}{c}{\begin{tabular}[c]{@{}c@{}} ResNet32x4\\ {(79.11)}\end{tabular}} & \multicolumn{1}{c}{\begin{tabular}[c]{@{}c@{}} ResNet32x4\\ {(79.11)}\end{tabular}} & \multicolumn{1}{c}{\begin{tabular}[c]{@{}c@{}} ResNet32x4\\ {(79.11)}\end{tabular}} & \multicolumn{1}{c}{\begin{tabular}[c]{@{}c@{}} ResNet32x4\\ {(79.11)}\end{tabular}} & \multicolumn{1}{c}{\begin{tabular}[c]{@{}c@{}} ResNet50\\ {(79.51)}\end{tabular}} & \multicolumn{1}{c}{\begin{tabular}[c]{@{}c@{}} ResNet50\\ {(79.51)}\end{tabular}} & \multicolumn{1}{c}{\begin{tabular}[c]{@{}c@{}} ResNet50\\ {(79.51)}\end{tabular}} \\
			\textbf{Student} & \multicolumn{1}{c}{MobileNetV2} & \multicolumn{1}{c}{VGG8} & \multicolumn{1}{c}{MobileNetV2} & \multicolumn{1}{c}{VGG8} & \multicolumn{1}{c}{ShuffleNetV1} & \multicolumn{1}{c}{ShuffleNetV2} & \multicolumn{1}{c}{MobileNetV2} & \multicolumn{1}{c}{VGG8} & \multicolumn{1}{c}{ShuffleNetV1} \\
			\hline
			NOKD & ${64.59 \pm 0.34}$ &  ${70.61 \pm 0.44}$ &  ${64.59 \pm 0.34}$ &  ${70.61 \pm 0.44}$ &  ${71.47 \pm 0.28}$ &  ${72.63 \pm 0.49}$ &  ${64.59 \pm 0.34}$ &  ${70.61 \pm 0.44}$ &  ${71.47 \pm 0.28}$ \\
			\hdashline
			BLKD (2015)  & ${68.31 \pm 0.21}$ &  ${73.38 \pm 0.15}$ &  ${67.19 \pm 0.39}$ &  ${72.57 \pm 0.21}$ &  ${74.17 \pm 0.15}$ &  ${74.76 \pm 0.12}$ &  ${67.61 \pm 0.89}$ &  ${73.52 \pm 0.21}$ &  ${75.19 \pm 0.33}$ \\
			FitNet (2015) & ${68.63 \pm 0.15}$ &  ${73.46 \pm 0.11}$ &  ${67.57 \pm 0.28}$ &  ${73.47 \pm 0.22}$ &  ${75.73 \pm 0.43}$ &  ${76.64 \pm 0.34}$ &  ${67.41 \pm 0.92}$ &  ${72.97 \pm 0.32}$ &  ${75.16 \pm 0.40}$ \\
			AT (2017)   & ${68.76 \pm 0.11}$ &  ${73.22 \pm 0.34}$ &  ${67.03 \pm 0.24}$ &  ${72.06 \pm 0.42}$ &  ${75.79 \pm 0.18}$ &  ${75.98 \pm 0.21}$ &  ${65.97 \pm 0.42}$ &  ${73.83 \pm 0.19}$ &  ${75.80 \pm 0.22}$ \\
			PKT (2018)  & ${68.64 \pm 0.43}$ &  ${73.70 \pm 0.26}$ &  ${68.00 \pm 0.26}$ &  ${72.86 \pm 0.21}$ &  ${74.88 \pm 0.23}$ &  ${75.52 \pm 0.17}$ &  ${67.92 \pm 0.87}$ &  ${73.61 \pm 0.23}$ &  ${75.59 \pm 0.17}$ \\
			FT (2018)   & ${67.32 \pm 3.99}$ &  ${73.56 \pm 0.25}$ &  ${64.61 \pm 0.33}$ &  ${72.83 \pm 0.14}$ &  ${75.48 \pm 0.22}$ &  ${76.16 \pm 0.42}$ &  ${67.92 \pm 0.22}$ &  ${73.00 \pm 0.08}$ &  ${76.03 \pm 0.24}$ \\
			SP (2019)   & ${67.77 \pm 0.65}$ &  ${73.40 \pm 0.14}$ &  ${67.33 \pm 0.50}$ &  ${72.94 \pm 0.34}$ &  ${75.28 \pm 0.11}$ &  ${75.95 \pm 0.38}$ &  ${68.55 \pm 0.30}$ &  ${73.58 \pm 0.18}$ &  ${76.11 \pm 0.23}$ \\
			VID (2019)  & ${67.54 \pm 0.47}$ &  ${74.00 \pm 0.13}$ &  ${66.80 \pm 0.59}$ &  ${72.84 \pm 0.29}$ &  ${75.36 \pm 0.20}$ &  ${75.62 \pm 0.18}$ &  ${68.42 \pm 0.37}$ &  ${73.48 \pm 0.05}$ &  ${75.43 \pm 0.29}$ \\
			RKD (2019)   & ${67.77 \pm 0.77}$ &  ${73.36 \pm 0.20}$ &  ${67.22 \pm 0.19}$ &  ${72.22 \pm 0.13}$ &  ${74.66 \pm 0.16}$ &  ${75.36 \pm 0.38}$ &  ${67.72 \pm 0.65}$ &  ${73.39 \pm 0.23}$ &  ${75.50 \pm 0.38}$ \\
			AB (2019)   & {NA} &  {NA} &  {NA} &  {NA} &  ${75.38 \pm 0.48}$ &  ${76.00 \pm 0.05}$ &  ${68.66 \pm 0.67}$ &  ${73.50 \pm 0.20}$ &  {NA} \\
			CRD (2019)  & ${70.92 \pm 1.70}$ &  ${74.53 \pm 0.41}$ &  ${69.02 \pm 0.25}$ &  ${73.90 \pm 0.19}$ &  ${75.25 \pm 0.31}$ &  ${76.06 \pm 0.20}$ &  ${69.59 \pm 0.68}$ &  ${74.42 \pm 0.06}$ &  ${76.26 \pm 0.36}$ \\
			TAKD (2020) & ${68.67 \pm 0.43}$ &  ${73.85 \pm 0.12}$ &  ${67.29 \pm 0.36}$ &  ${72.62 \pm 0.22}$ &  ${74.41 \pm 0.29}$ &  ${74.86 \pm 0.03}$ &  ${68.37 \pm 0.47}$ &  ${73.43 \pm 0.17}$ &  ${75.14 \pm 0.06}$ \\
			SSKD (2020) & $\underline{71.57 \pm 0.13}$ &  ${75.30 \pm 0.16}$ &  ${69.94 \pm 1.88}$ &  ${75.48 \pm 0.21}$ &  $\underline{77.88 \pm 0.13}$ &  $\underline{78.47 \pm 0.09}$ &  $\underline{71.86 \pm 0.30}$ &  $\underline{75.48 \pm 0.10}$ &  $\underline{77.87 \pm 0.22}$ \\
			ReviewKD (2021) & {NA} &  {NA} &  {NA} &  {NA} &  ${76.78 \pm 0.45}$ &  ${77.14 \pm 0.32}$ &  ${66.35 \pm 0.25}$ &  {NA} &  {NA} \\
			SemCKD (2021) & ${69.30 \pm 0.25}$ &  ${74.48 \pm 0.08}$ &  ${68.56 \pm 0.37}$ &  ${74.83 \pm 0.31}$ &  ${76.56 \pm 0.12}$ &  ${77.50 \pm 0.22}$ &  ${67.62 \pm 0.54}$ &  ${73.80 \pm 0.33}$ &  ${75.77 \pm 0.26}$ \\
			SimKD (2022) & ${69.76 \pm 0.18}$ &  $\bm{75.98 \pm 0.16}$ &  ${68.31 \pm 0.18}$ &  $\underline{75.61 \pm 0.11}$ &  ${75.88 \pm 2.40}$ &  ${77.68 \pm 0.20}$ &  $\notes{63.48 \pm 0.14}$ &  $ \notes{66.97 \pm 0.24}$ &  $ \notes{65.47 \pm 0.12}$ \\
			\hline
			\textbf{cGAN-KD} & $68.35 \pm 0.22$ & $72.08 \pm 0.21$ & $68.35 \pm 0.22$ & $72.08 \pm 0.21$ & $74.60 \pm 0.48$ & $75.22 \pm 0.43$  & $68.35 \pm 0.22$ & $72.08 \pm 0.21$ & $74.60 \pm 0.48$ \\ 
			\begin{tabular}[l]{@{}l@{}} \textbf{cGAN-KD}\\ \textbf{ + BLKD}\end{tabular} & ${70.03 \pm 0.35}$ &  ${74.51 \pm 0.12}$ &  ${69.66 \pm 0.20}$ &  ${74.33 \pm 0.21}$ &  ${76.17 \pm 0.21}$ &  ${76.99 \pm 0.04}$ &  ${70.28 \pm 0.14}$ &  ${74.89 \pm 0.06}$ &  ${76.93 \pm 0.13}$ \\
			\begin{tabular}[l]{@{}l@{}} \textbf{cGAN-KD}\\ \textbf{ + FitNet}\end{tabular} & ${70.43 \pm 0.14}$ &  ${75.02 \pm 0.29}$ &  ${69.72 \pm 0.38}$ &  ${74.66 \pm 0.37}$ &  ${77.52 \pm 0.25}$ &  ${77.55 \pm 0.17}$ &  ${70.70 \pm 0.23}$ &  ${75.15 \pm 0.29}$ &  ${77.15 \pm 0.28}$ \\
			\begin{tabular}[l]{@{}l@{}} \textbf{cGAN-KD}\\ \textbf{ + VID}\end{tabular} & ${70.05 \pm 0.24}$ &  ${74.73 \pm 0.21}$ &  ${69.56 \pm 0.14}$ &  ${74.68 \pm 0.19}$ &  ${77.00 \pm 0.14}$ &  ${77.26 \pm 0.13}$ &  ${70.75 \pm 0.32}$ &  ${75.32 \pm 0.30}$ &  ${77.13 \pm 0.07}$ \\
			\begin{tabular}[l]{@{}l@{}} \textbf{cGAN-KD}\\ \textbf{ + RKD}\end{tabular} & ${70.59 \pm 0.29}$ &  ${74.65 \pm 0.13}$ &  ${70.00 \pm 0.12}$ &  ${74.23 \pm 0.21}$ &  ${76.30 \pm 0.12}$ &  ${77.25 \pm 0.21}$ &  ${70.73 \pm 0.24}$ &  ${75.03 \pm 0.15}$ &  ${77.02 \pm 0.32}$ \\
			\begin{tabular}[l]{@{}l@{}} \textbf{cGAN-KD}\\ \textbf{ + CRD}\end{tabular} & ${71.29 \pm 0.17}$ &  ${75.34 \pm 0.23}$ &  $\underline{70.50 \pm 0.21}$ &  ${75.31 \pm 0.03}$ &  ${77.27 \pm 0.16}$ &  ${77.61 \pm 0.27}$ &  ${71.38 \pm 0.10}$ &  ${75.20 \pm 0.23}$ &  ${77.50 \pm 0.15}$ \\
			\begin{tabular}[l]{@{}l@{}} \textbf{cGAN-KD}\\ \textbf{ + SSKD}\end{tabular} & $\bm{72.00 \pm 0.16}$ &  $\underline{75.60 \pm 0.21}$ &  $\bm{71.63 \pm 1.35}$ &  $\bm{76.58 \pm 0.27}$ &  $\bm{78.67 \pm 0.10}$ &  $\bm{78.99 \pm 0.07}$ &  $\bm{72.96 \pm 0.18}$ &  $\bm{76.12 \pm 0.25}$ &  $\bm{77.99 \pm 0.19}$ \\

			\hline \hline
			\textbf{Teacher} & \multicolumn{1}{c}{\begin{tabular}[c]{@{}c@{}} DenseNet121\\ {(79.98)}\end{tabular}} & \multicolumn{1}{c}{\begin{tabular}[c]{@{}c@{}} DenseNet121\\ {(79.98)}\end{tabular}} & \multicolumn{1}{c}{\begin{tabular}[c]{@{}c@{}} DenseNet121\\ {(79.98)}\end{tabular}} & \multicolumn{1}{c}{\begin{tabular}[c]{@{}c@{}} DenseNet121\\ {(79.98)}\end{tabular}} & \multicolumn{1}{c}{\begin{tabular}[c]{@{}c@{}} DenseNet121\\ {(79.98)}\end{tabular}} & \multicolumn{1}{c}{\begin{tabular}[c]{@{}c@{}} DenseNet121\\ {(79.98)}\end{tabular}} & & & \\
			\textbf{Student} & \multicolumn{1}{c}{MobileNetV2} & \multicolumn{1}{c}{ResNet20} & \multicolumn{1}{c}{VGG8} & \multicolumn{1}{c}{ResNet8x4} & \multicolumn{1}{c}{ShuffleNetV1} & \multicolumn{1}{c}{ShuffleNetV2} & & & \\
			\hline
			NOKD & $64.59 \pm 0.34$ & $69.22 \pm 0.39$ & $70.61 \pm 0.44$ & $72.74 \pm 0.07$ & $71.47 \pm 0.28$ & $72.63 \pm 0.49$ &       &       &  \\
			\cdashline{1-7}
			BLKD (2015)  & ${68.21 \pm 0.39}$ &  ${69.93 \pm 0.25}$ &  $\underline{73.45 \pm 0.31}$ &  $\underline{74.32 \pm 0.42}$ &  $\underline{75.59 \pm 0.15}$ &  $\underline{76.44 \pm 0.14}$ &       &       &  \\
			TAKD (2020)  & ${68.29 \pm 0.04}$ &  ${69.89 \pm 0.23}$ &  ${73.41 \pm 0.16}$ &  ${74.10 \pm 0.29}$ &  ${75.33 \pm 0.42}$ & ${75.98 \pm 0.37}$ &       &       &  \\
			\hline
			\textbf{cGAN-KD} & $\underline{68.35 \pm 0.22}$ & $\underline{70.71 \pm 0.10}$ & ${72.08 \pm 0.21}$ & ${72.90 \pm 0.11}$ & ${74.60 \pm 0.48}$ & ${75.22 \pm 0.43}$ & & \\
			\begin{tabular}[l]{@{}l@{}} \textbf{cGAN-KD}\\ \textbf{ + BLKD}\end{tabular} & $\bm{70.86 \pm 0.36}$ &  $\bm{71.68 \pm 0.39}$ &  $\bm{75.24 \pm 0.08}$ &  $\bm{74.81 \pm 0.25}$ &  $\bm{77.44 \pm 0.10}$ &  $\bm{78.29 \pm 0.20}$ &       &       &  \\
			\hline\hline
		\end{tabular}%
	\end{adjustbox}
	\label{tab:cifar_KD_cross_results}%
\end{table}%

\begin{table}[!h]
	\centering
	\caption{{ \textbf{ImageNet-100: Top-1 test accuracy (\%) of compared KD methods.} Bold and underline denote the best and the second best results, respectively. }}
	\begin{adjustbox}{width=0.6\textwidth}
		\begin{tabular}{l|ccccc}
			\hline \hline
			\textbf{Teacher} & \multicolumn{1}{c}{\begin{tabular}[c]{@{}c@{}} ResNet110\\ {(75.31)}\end{tabular}} & \multicolumn{1}{c}{\begin{tabular}[c]{@{}c@{}} WRN40$\times$2\\ {(77.67)}\end{tabular}} & \multicolumn{1}{c}{\begin{tabular}[c]{@{}c@{}} ResNet34\\ {(81.54)}\end{tabular}} & \multicolumn{1}{c}{\begin{tabular}[c]{@{}c@{}} VGG19\\ {(83.41)}\end{tabular}} & \multicolumn{1}{c}{\begin{tabular}[c]{@{}c@{}} VGG19\\ {(83.41)}\end{tabular}} \\
			\textbf{Student} & \multicolumn{1}{c}{ResNet20} & \multicolumn{1}{c}{WRN40$\times$1} & \multicolumn{1}{c}{WRN40$\times$1} & \multicolumn{1}{c}{VGG8} & \multicolumn{1}{c}{ShuffleNetV1} \\
			\hline
			NOKD & 65.25  & 69.98  & 69.98  & 77.36  & 75.02  \\
			\hdashline
			BLKD (2015)  & 65.35  & 70.79  & 69.36  & 80.15  & 77.07  \\
			FitNet (2015) & 66.20  & 71.46  & 71.38  & 80.26  & NA \\
			VID (2019)  & 66.76  & 71.38  & 71.34  & 81.03  & 77.41  \\
			RKD (2019)  & 66.45  & 71.37  & 69.81  & 80.70  & 77.46  \\
			CRD  (2019) & 66.79  & 72.13  & 70.39  & 81.17  & 77.65  \\
			SSKD (2020) & 67.05  & 72.92  & 71.44  & \underline{81.85}  & \underline{78.93}  \\
			SemCKD (2021) & NA & 71.46  & 71.79  & 80.74  & 77.72  \\
			SimKD (2022) & $ \notes{60.95}$  & 71.26  & 72.72  & 81.46  & 78.67  \\
			\hline
			\textbf{cGAN-KD}  & 67.15  & 72.17  & 72.17  & 78.21  & 76.89  \\
			\textbf{cGAN-KD + BLKD}  & 66.87  & 72.83  & 72.18  & 80.29  & 78.41  \\
			\textbf{cGAN-KD + FitNet}  & 66.73  & 73.53  & 72.63  & 81.16  & NA \\
			\textbf{cGAN-KD + VID} & 67.24  & 72.87  & 72.53  & 81.19  & 78.65  \\
			\textbf{cGAN-KD + RKD} & 67.66  & 73.10  & 72.21  & 80.93  & 78.70  \\
			\textbf{cGAN-KD + CRD} & \underline{67.91}  & \underline{73.18}  & \underline{73.06}  & 81.46  & 78.40  \\
			\textbf{cGAN-KD + SSKD} & \textbf{68.58} & \textbf{74.15} & \textbf{73.31} & \textbf{82.46} & \textbf{80.30} \\
			\hline\hline
		\end{tabular}%
	\end{adjustbox}
	\label{tab:imagenet_KD_results}%
\end{table}%

\subsection{Regression: Steering Angle and UTKFace}\label{sec:exp_regression}

\subsubsection{Datasets}\label{sec:reg_datasets}
This experiment is conducted on the Steering Angle and UTKFace datasets to show that cGAN-KD also performs very well in the image regression tasks with a scalar response variable. Steering Angle, a subset of an autonomous driving dataset \citep{steeringangle, steeringangle2}, includes 12,508 RGB images at $64\times 64$ resolution with 1,773 distinct steering angles in $[-88.13^\circ, 97.92^\circ]$ as labels. We split the range of angles into 246 disjoint unequal intervals, where each interval should have at least 40 instances. In each interval, 80\% instances are randomly selected for training, and the rest are for testing. UTKFace is an RGB human face image dataset with ages as regression labels. We use the processed UTKFace dataset \citep{ding2020continuous,ding2021ccgan}, which consists of 
14,760 RGB images with ages in [1, 60]. The number of images ranges from 50 to 1051 for different ages, and all images are of size $64\times64$. Among these images, 80\% are randomly selected for training for each age, and the rest are held out for testing.

\subsubsection{The selection of teachers and students}\label{sec:reg_sel_tea_stu}
Similar to classification, to select student and teacher models for this experiment, some popular neural networks are trained from scratch on each dataset, and their \textit{mean absolute errors} (MAE) on test sets are shown in Tables S.7.13 and S.8.15 in Appendix. Some light-weight neural networks (shown in \Cref{tab:SA_UK_DA_results}) with high test MAE are chosen as student models, and we aim to improve their accuracies. VGG19 and VGG11 \citep{simonyan2014very} are chosen as the teacher models for the label adjustment module in cGAN-KD on Steering Angle and UTKFace, respectively.

\subsubsection{Compared methods and implementation details}\label{sec:reg_candidates}
{
	
Note that there is no general KD method for image regression tasks with a scalar response. Thus, we compare cGAN-KD against modified FitNet, AT, and RKD, where we exclude any logit-related terms from the loss functions of these three feature-based KD methods. Compared KD methods in regression tasks include NOKD, FitNet, AT, RKD, cGAN-KD, and cGAN-KD + FitNet. Similar to our classification experiment, we implement FitNet, AT, and RKD by \texttt{RepDistiller} (a GitHub repository provided by \citet{tian2019contrastive}). 

For cGAN-KD-based methods, we adopt the SAGAN architecture \citep{zhang2019self} and train one CcGAN model (SVDL+ILI) with DiffAugment for each dataset. We let $\rho=0.7$ on both datasets. We generate 50,000 and 60,000 processed fake samples for the Steering Angle and UTKFace experiments, respectively.

All results of the regression experiments are reported in \textbf{a single trial}. Other implementation details such as learning rate, batch size, weight decay, number of epochs, seed, and optimizer are shown in Sections S.7 and S.8 in Appendix.

}

\subsubsection{Experimental results}\label{sec:reg_results}
{
	
The quantitative results of both experiments are shown in Table \ref{tab:SA_UK_DA_results}. For Steering Angle, cGAN-KD outperforms NOKD with a large margin. Notably, the test error of WRN16$\times$1 is reduced by $(5.74-1.79)/5.74\times 100\%=68.82\%$ on Steering Angle. FitNet and RKD perform surprisingly well on Steering Angle, and they outperform cGAN-KD under two teacher-student pairs (ResNet34-ShuffleNetV2 and VGG19-ShuffleNetV1). Thus, we also incorporate FitNet into cGAN-KD, i.e., cGAN-KD + FitNet. It beats all three existing KD methods in all teacher-student pairs. For UTKFace, cGAN-KD beats all existing KD methods with a large margin. For example, MobileNetV2's test error is reduced by $(7.16-4.84)/7.16\times 100\%=32.40\%$ by using cGAN-KD, but the three feature-based KD methods do not have significant effects on this dataset. cGAN-KD + FitNet is substantially better than FitNet but slightly worse than cGAN-KD, implying we should not combine cGAN-KD with a less effective KD method. Please also note that, unlike many existing KD methods, cGAN-KD even outperforms the teacher models in the UTKFace experiment, implying CcGAN may generate new information helpful for the regression task.  

}

\begin{table}[htbp]
	\centering
	\caption{ {   \textbf{The test mean absolute errors (MAE) of compared KD methods on Steering Angle and UTKFace.} The units of the test MAE for Steering Angle and UTKFace are degrees and years, respectively. Bold and underline denote the best and the second best results, respectively.} }
	\begin{adjustbox}{width=0.8\textwidth}
		\begin{tabular}{l|ccccccc}
			\hline\hline
			\multicolumn{8}{c}{\textbf{Steering Angle}} \\
			\hline
			\textbf{Teacher} & \begin{tabular}[c]{@{}c@{}} ResNet34 \\ {(1.30)}\end{tabular} & \begin{tabular}[c]{@{}c@{}} ResNet34 \\ {(1.30)}\end{tabular} & \begin{tabular}[c]{@{}c@{}} ResNet34 \\ {(1.30)}\end{tabular} & \begin{tabular}[c]{@{}c@{}} ResNet34 \\ {(1.30)}\end{tabular} & \begin{tabular}[c]{@{}c@{}} VGG19 \\ {(1.10)}\end{tabular} & \begin{tabular}[c]{@{}c@{}} VGG19 \\ {(1.10)}\end{tabular} & \begin{tabular}[c]{@{}c@{}} VGG19 \\ {(1.10)}\end{tabular} \\
			\textbf{Student} & ResNet20 & ShuffleNetV2 & WRN16$\times$1 & ResNet8x4 & WRN40$\times$1 & ShuffleNetV1 & ResNet56 \\
			\hline
			NOKD & 4.86 & 5.14 & 5.74 & 3.90 & 3.70 &  {3.47} &  {2.63} \\
			\hdashline
			FitNet (2015) &  {2.18} &  {2.23} &  {2.16} &  {2.69} &  {2.15} &  {2.14} &  {1.93} \\
			AT (2017) &  {2.75} &  \underline{2.04} &  {2.81} &  {2.32} &  {2.19} &  {2.32} &  {2.34} \\
			RKD (2019) &  {2.80} &  {2.43} &  {4.49} &  {2.69} &  {2.25} &  \underline{2.12} &  {2.44} \\
			\hline
			\textbf{cGAN-KD} &  \underline{1.54} &  {2.51} &  \textbf{1.79} &  \underline{2.15} &  \underline{1.52} &  {2.21} &  \underline{1.55} \\
			\textbf{cGAN-KD + FitNet} &  \textbf{1.53} &  \textbf{1.68} &  \underline{1.87} &  \textbf{1.82} &  \textbf{1.40} &  \textbf{2.05} &  \textbf{1.46} \\
			\hline\hline
			
			\multicolumn{8}{c}{\textbf{UTKFace}} \\
			\hline
			{\textbf{Teacher}} &  \begin{tabular}[c]{@{}c@{}} ResNet34 \\ {(5.29)}\end{tabular} &  \begin{tabular}[c]{@{}c@{}} ResNet34 \\ {(5.29)}\end{tabular} &  \begin{tabular}[c]{@{}c@{}} ResNet34 \\ {(5.29)}\end{tabular} &  \begin{tabular}[c]{@{}c@{}} ResNet34 \\ {(5.29)}\end{tabular} &  \begin{tabular}[c]{@{}c@{}} VGG11 \\ {(5.12)}\end{tabular} &  \begin{tabular}[c]{@{}c@{}} VGG11 \\ {(5.12)}\end{tabular} &  \begin{tabular}[c]{@{}c@{}} VGG11 \\ {(5.12)}\end{tabular} \\
			{\textbf{Student}} &  {WRN16$\times$1} &  {MobileNetV2} &  {ResNet56} &  {ShuffleNetV1} &  {ResNet20} &  {WRN40$\times$1} &  {ResNet8x4} \\
			\hline
			{NOKD} & 7.25  & 7.16  & 7.06  & 7.03  & 6.87  & 6.70  & 6.68  \\
			\hdashline
			{FitNet} (2015) & 7.04  & {6.86}  & {6.78}  & {6.93}  & 6.52  & 6.68  & 6.25  \\
			{AT} (2017) & 7.20  & 6.97  & 7.30  & 7.23  & 6.88  & 6.85  & 6.64  \\
			{RKD} (2019) & {6.71}  & 7.13  & 6.80  & 7.33  & {6.38}  & {6.39}  & {5.97}  \\
			\hline
			{\textbf{cGAN-KD}} & \textbf{5.01}  & \textbf{4.84}  & \textbf{4.90}  & \textbf{4.95}  & \textbf{4.92}  & \textbf{4.92}  & \textbf{4.76}  \\
			\textbf{cGAN-KD + FitNet} & \underline{5.63}  & \underline{5.65}  & \underline{5.52}  & \underline{5.58}  & \underline{5.23}  & \underline{5.24}  & \underline{4.96}  \\
			\hline\hline
		\end{tabular}%
	\end{adjustbox}
	\label{tab:SA_UK_DA_results}%
\end{table}%

\subsection{Ablation study: the effect of different (sub-)modules of cGAN-KD}\label{sec:exp_ablation}

\subsubsection{Some visual and quantitative illustration for \textbf{M1} and \textbf{M2}}\label{sec:exp_ablation_ill}

\begin{figure}[!h]
	\centering
	\includegraphics[width=0.75\textwidth]{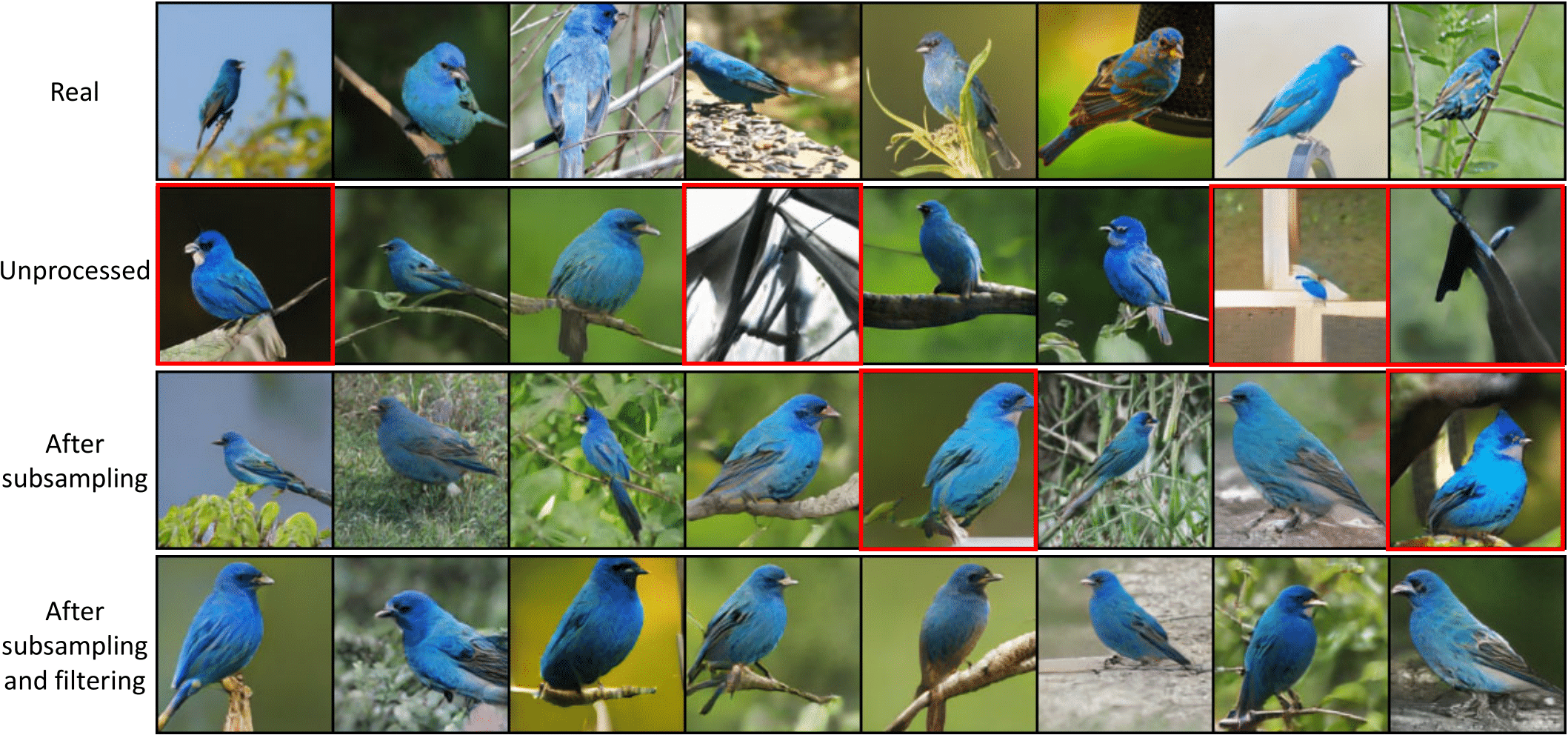} 
	\caption{ \rev{ \textbf{The subsampling and filtering (sub-)modules can effectively improve fake images' visual quality.} Some example images are shown here for the ``indigo bunting" class at $128\times 128$ resolution in the ImageNet-100 experiment. The first and second rows includes ten real images and ten unprocessed fake images, respectively. The third and fourth rows include fake images processed by subsampling only and subsampling+filtering, respectively. We observe many unrealistic unprocessed fake images in the second row (marked in red rectangles). The subsampling module (third row) can effectively remove most of them and the filtering sub-module (fourth row) can further improve the visual quality.}}
	\label{fig:ImageNet100_example_images_class2}
\end{figure}

\begin{figure}[!h]
	\centering
	\includegraphics[width=0.6\textwidth]{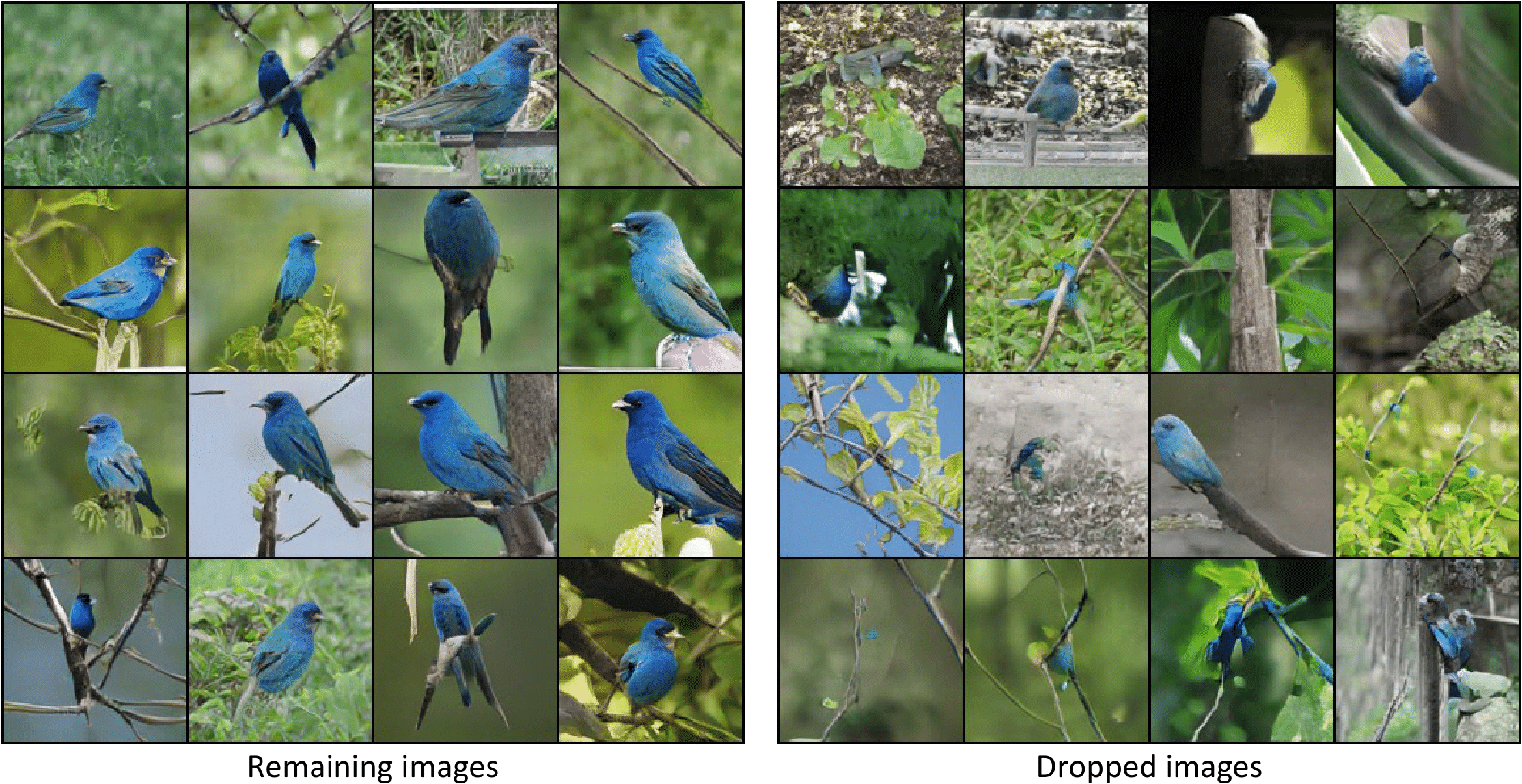} 
	\caption{ \textbf{Some example fake images processed by the filtering sub-module for the ``indigo bunting" class at $128\times 128$ resolution in the ImageNet-100 experiment (classification) in \Cref{sec:cGAN-KD_experiment}.} The left image grid shows some fake images are not dropped by the filtering sub-module, while the right grid includes fake images that are dropped. Both Figs. \ref{fig:ImageNet100_example_images_class2} and \ref{fig:ImageNet100_example_filtering_images_class2} show that, as a side effect, the filtering sub-module can effectively drop most unrealistic fake images, so the overall visual quality is improved. }
	\label{fig:ImageNet100_example_filtering_images_class2}
\end{figure}

\begin{figure}[!h]
	\centering
	\includegraphics[width=0.6\textwidth]{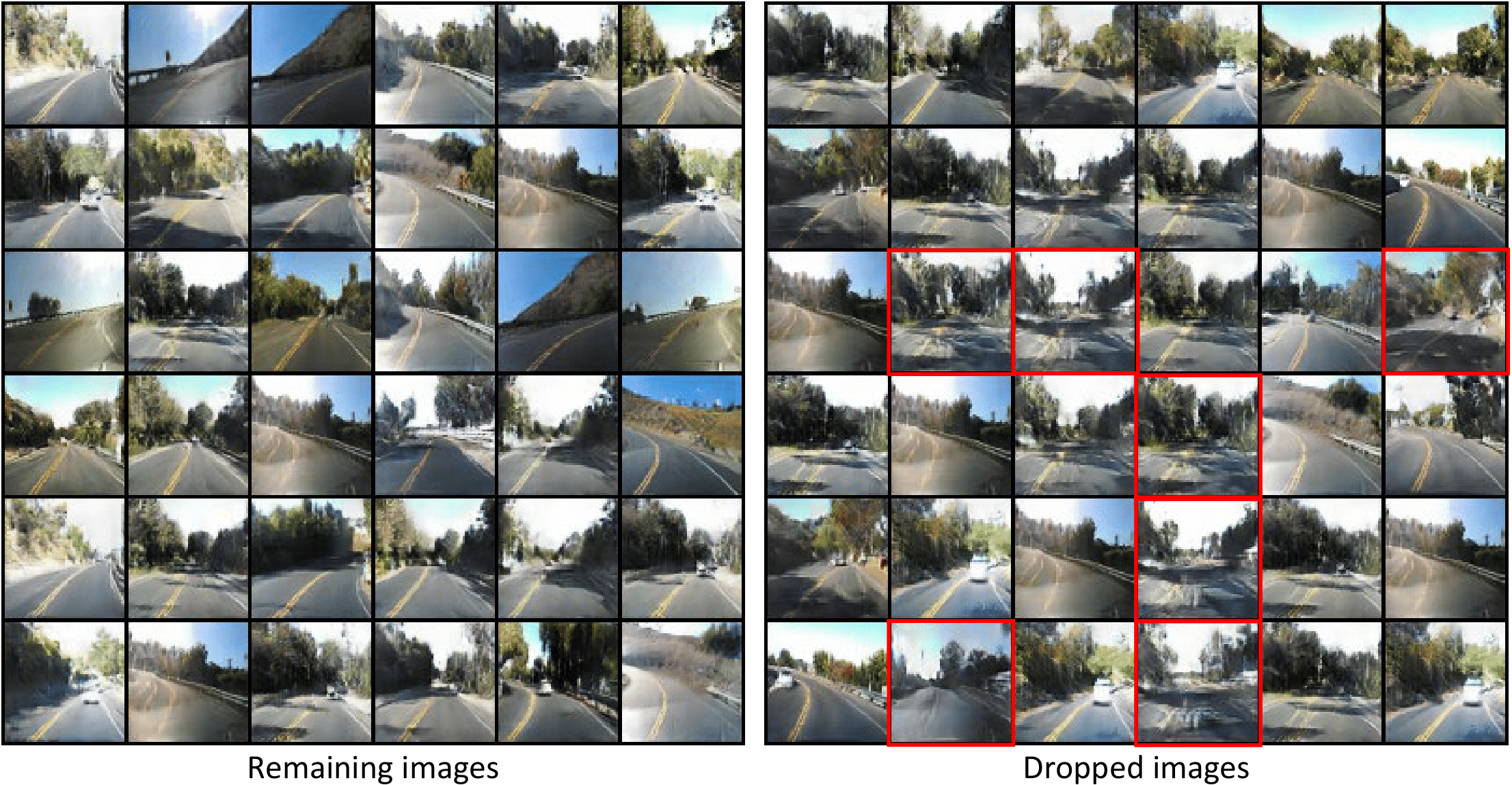} 
	\caption{\textbf{Some example fake images processed by the filtering sub-module at $64\times 64$ resolution in the Steering Angle experiment (regression).} The left image grid shows some fake images are not dropped by the filtering sub-module, while the right grid includes fake images that are dropped. The right grid shows that, as a side effect, the filtering sub-module can effectively drop many unrealistic fake images (marked in red rectangles). Some realistic images are also dropped in the right image grid because they are not consistent with their assigned labels. }
	\label{fig:SteeringAngle_example_filtering_images}
\end{figure}

\begin{figure}[!h]
	\centering
	\subfloat[][ImageNet-100 (``indigo bunting" class)]{
		\includegraphics[width=0.4\textwidth]{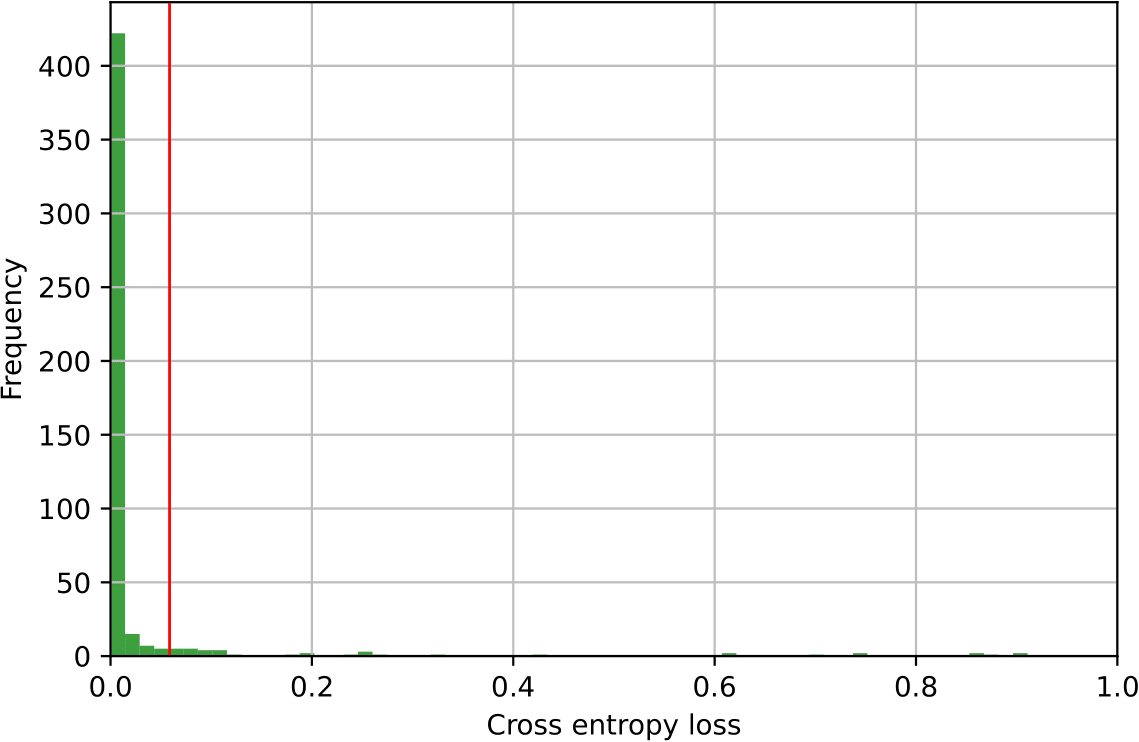}
		\label{fig:imagenet100_filtering_histogram_of_errors}}
	\subfloat[][Steering Angle]{
		\includegraphics[width=0.4\textwidth]{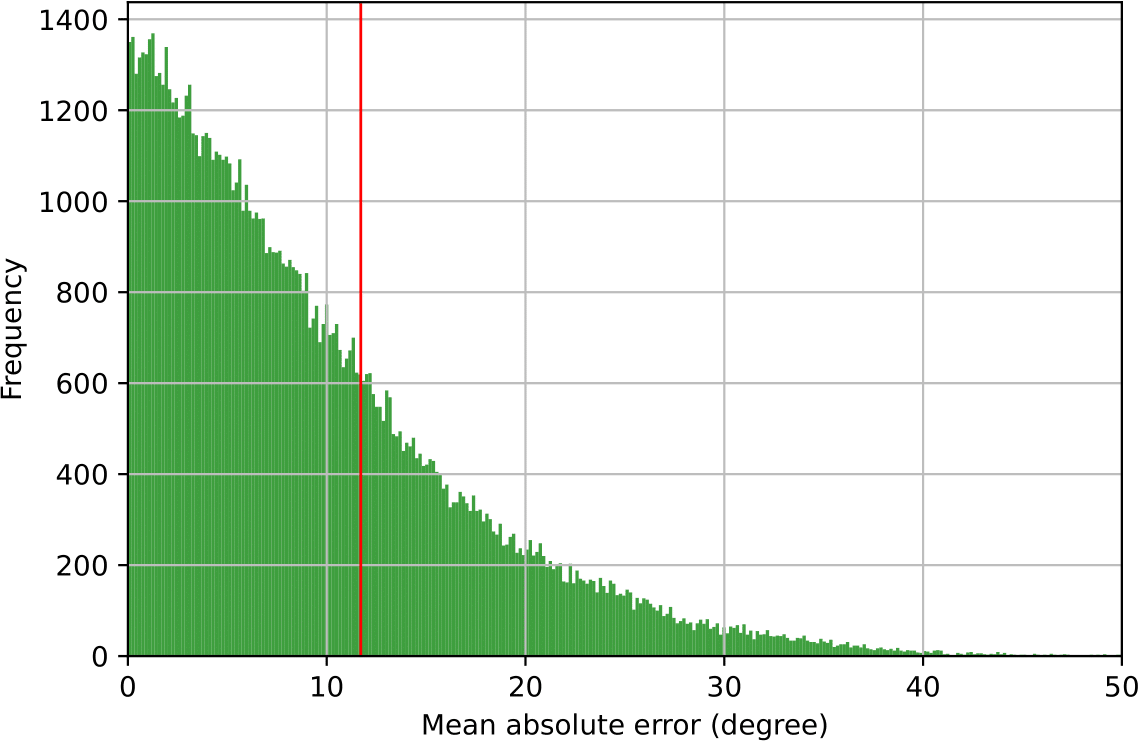}
		\label{fig:steeringangle_filtering_histogram_of_errors}}
	\caption{\textbf{The histograms of cross entropy losses and mean absolute errors in the ImageNet-100 (classification) and Steering Angle (regression) experiments.} The red vertical lines represent the filtering threshold $\alpha$ (with $\rho=0.9$ for ImageNet-100 and $\rho=0.7$ for Steering Angle). The fake samples with errors larger than the threshold are dropped. In Fig.\ \ref{fig:filtering_histogram_of_errors}(a), we filter 500 fake images generated via \textbf{M1} for the ``indigo bunting" class. Fig.\ \ref{fig:filtering_histogram_of_errors}(a) shows that, after the filtering, almost all remaining fake samples are label consistent, implying that it is unnecessary to apply replacement in classification. For the Steering Angle experiment, we filter 100,000 fake images generated by \textbf{M1} with a global $\alpha$. Fig.\ \ref{fig:filtering_histogram_of_errors}(b) shows that the tail distribution of errors in regression is much heavier than that in classification. Thus, we propose a smaller $\rho$ in regression, i.e., $\rho=0.7$. Although the filtering sub-module can effectively remove fake samples with very large errors, the errors of most remaining fake samples are still non-zero. Therefore, we need the replacement sub-module to further adjust their labels. }
	\label{fig:filtering_histogram_of_errors}
\end{figure}

{

As illustrated in \Cref{sec:cGAN-KD_drop_samples}, the subsampling module (\textbf{M1}) can effectively improve the visual quality of fake images. In \cref{fig:ImageNet100_example_images_class2}, we show some example fake images from the ``indigo bunting" class in the ImageNet-100 experiment in \Cref{sec:exp_classification}. We can see fake images directly generated from a trained cGAN may contain many unrealistic images (marked by red rectangles in the second row). But the subsampling module (third row) can effectively drop most of them. Besides the subsampling module, as described in \Cref{sec:cGAN-KD_label_adjustment} and visualized in \cref{fig:ImageNet100_example_images_class2}, the filtering sub-module can also enhance the visual quality. Some example fake images that are kept and dropped by filtering are shown in Figs.\ \ref{fig:ImageNet100_example_filtering_images_class2} and \ref{fig:SteeringAngle_example_filtering_images}, where most remaining images are high-quality while most dropped images have poor visual quality.  

\textbf{M2} consists of two sequential sub-modules, i.e., filtering and replacement. In addition to improving visual quality, the primary function of filtering is to increase the label consistency of fake images. Two illustrative figures and a table are shown in \cref{fig:filtering_histogram_of_errors} and \cref{tab:label_consistency_before_after_filtering}, respectively. From them, we can conclude that the filtering sub-module can effectively rule out fake images whose assigned labels are far from their predicted labels. Note that the predicted labels are assumed to be close to the actual labels of fake images, given that they are predictions from accurate teacher models. \cref{fig:filtering_histogram_of_errors} and \cref{tab:label_consistency_before_after_filtering} also show that, on CIFAR-100 and ImageNet-100, the remaining images' predicted labels are already consistent with their assigned labels after filtering, so the replacement sub-module is unnecessary for classification tasks. 

}

\begin{table}[!h]
	\centering
	\caption{\textbf{The label consistency of fake images before or after filtering in the CIFAR-100 and ImageNet-100 experiments.} We first generate 50,000 fake images (500 images per class) from \textbf{M1} in each experiment. These fake images are then filtered by a pre-trained teacher model (DenseNet121 for CIFAR-100 and DenseNet161 for ImageNet-100) with $\rho=0.9$. After filtering, there are 45,000 fake images left (450 images per class). Here, label consistency is the percentage of fake images whose assigned labels and predicted labels are equal. After filtering, almost all remaining images' predicted labels are consistent with their assigned labels. Therefore, the replacement sub-module is not necessary for classification tasks.}
	\begin{adjustbox}{width=0.35\textwidth}
		\begin{tabular}{cc|cc}
			\hline\hline
			\multicolumn{2}{c|}{CIFAR-100} & \multicolumn{2}{c}{ImageNet-100} \\
			\hline
			Before & After & Before & After \\
			\hline
			91.352\% & 97.802\% & 93.854\% & 99.064\% \\
			\hline\hline
		\end{tabular}%
	\end{adjustbox}
	\label{tab:label_consistency_before_after_filtering}%
\end{table}%

\subsubsection{Effects of different (sub-)modules on the prediction precision}\label{exp_ablation_pred}

\begin{figure}[!h]
	\centering
	\subfloat[][CIFAR-100]{
		\includegraphics[width=0.4\textwidth]{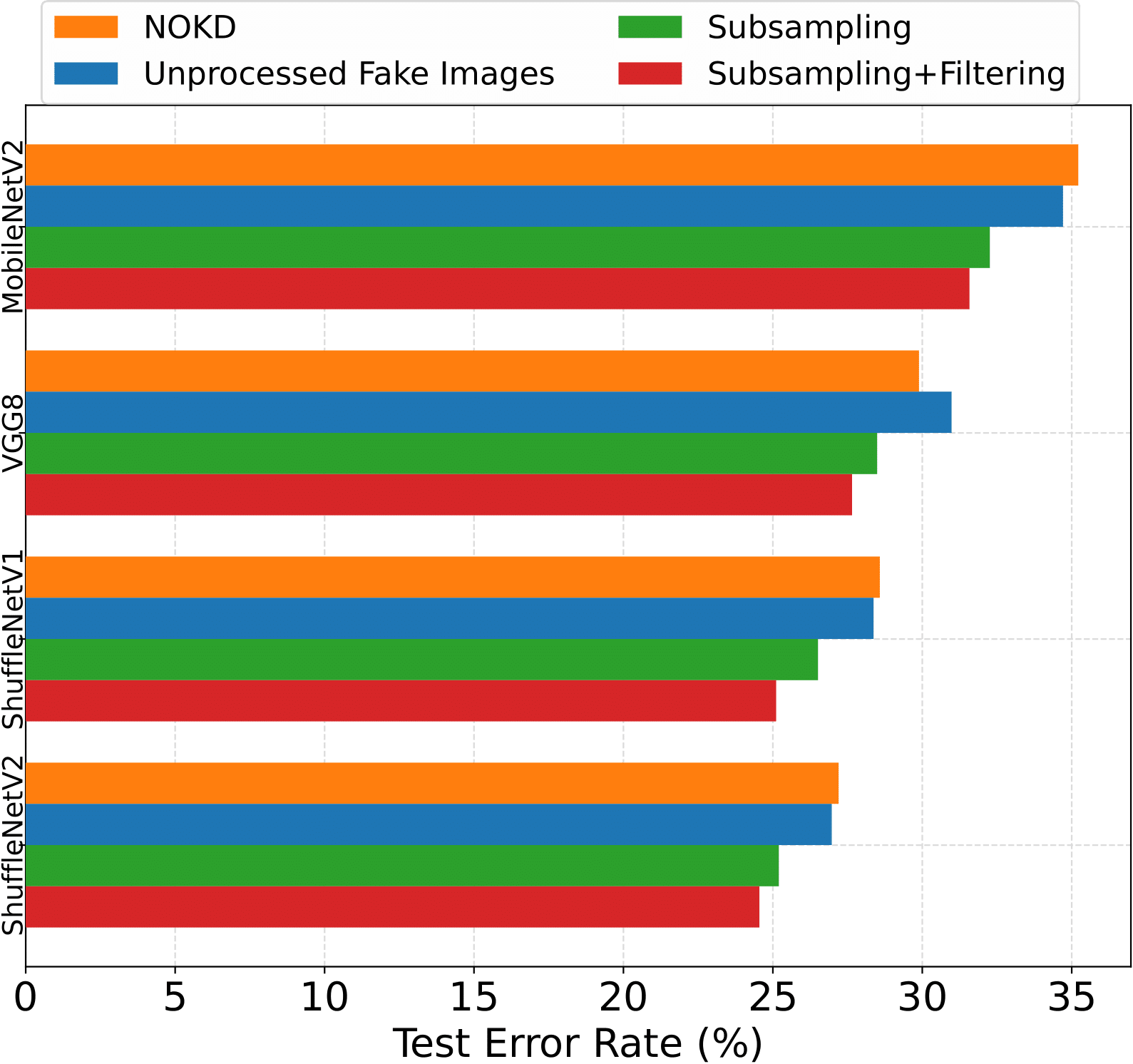}
		\label{fig:cifar_ablation}}
	\subfloat[][Steering Angle]{
		\includegraphics[width=0.4\textwidth]{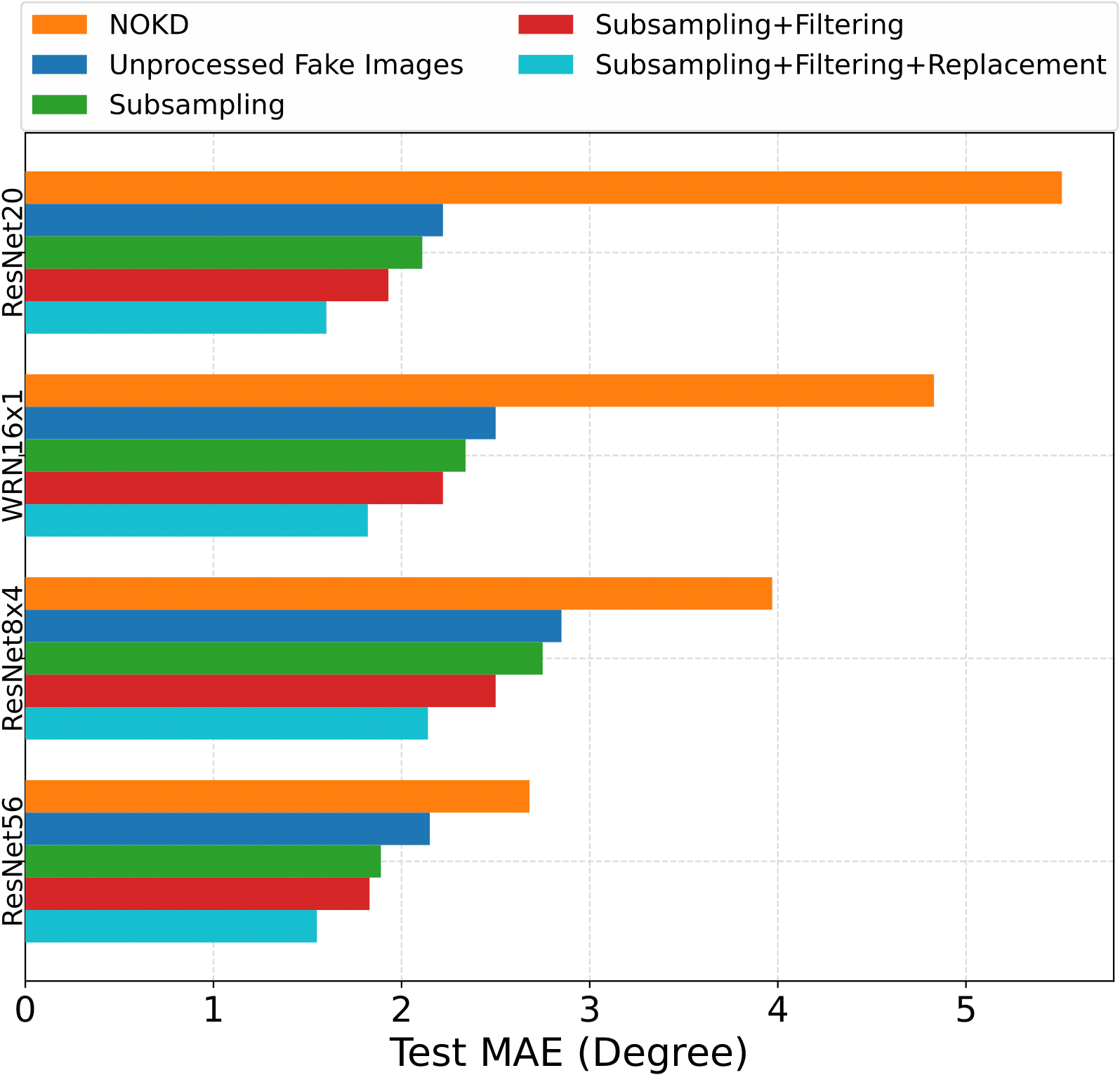}
		\label{fig:steeringangle_ablation}}
	\caption{\textbf{Ablation study: The effect of different (sub-)modules of cGAN-KD (subsampling, filtering, and replacement)}. The precision of student models gradually improves as we add these (sub-)modules into cGAN-KD sequentially, implying the combination of these modules leads to the highest precision. Please note that unprocessed fake samples may cause adverse effects if directly used in data augmentation, e.g., VGG8 in the CIFAR-100 experiment.}
	\label{fig:ablation}
\end{figure}

In this section, an ablation study is also designed to test the effectiveness of the subsampling, filtering, and replacement (for regression only) (sub-)modules in the cGAN-KD framework on CIFAR-100 and Steering Angle, aiming to show how cGAN-KD performs if these (sub-)modules are added into the framework one by one. In this study, other setups (e.g., $M^g$, $\rho$, and the teacher model) remain unchanged, and the results are reported based on a single trial. The quantitative result is visualized in \cref{fig:ablation}. We can see that the precision of student models gradually improves as we add these modules into cGAN-KD sequentially. The combination of these (sub-)modules leads to the highest precision. 

\subsection{Additional analysis}\label{sec:additional_analysis}

From Sections~\ref{sec:analysis_Mg} to \ref{sec:analysis_teacher}, we conduct some additional analyses to research the effects of $M^g$, $\rho$, and the selection of teacher in the cGAN-KD framework. Note that all results are reported in terms of a single trial due to time and facility constraints. We also provide a running time and memory cost comparison at the end of this section.

\subsubsection{Sensitivity analysis on the fake sample size $M^g$}\label{sec:analysis_Mg}
\begin{figure}[!h]
	\centering
	\subfloat[][CIFAR-100]{
		\includegraphics[width=0.4\textwidth]{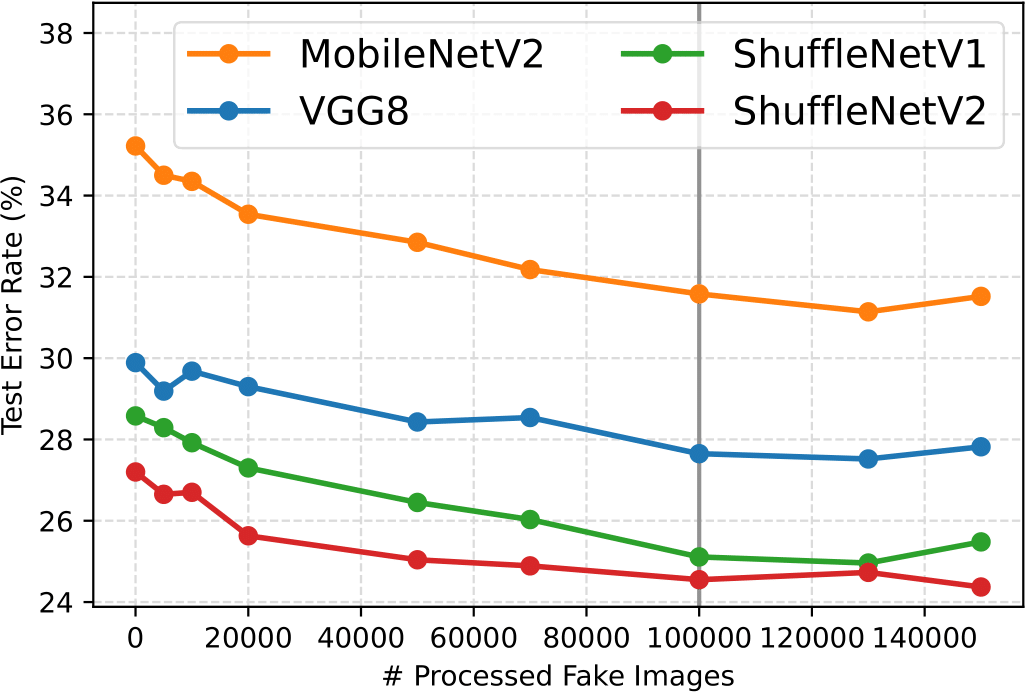}
		\label{fig:cifar_sensitivity_Mg}}
	\subfloat[][Steering Angle]{
		\includegraphics[width=0.4\textwidth]{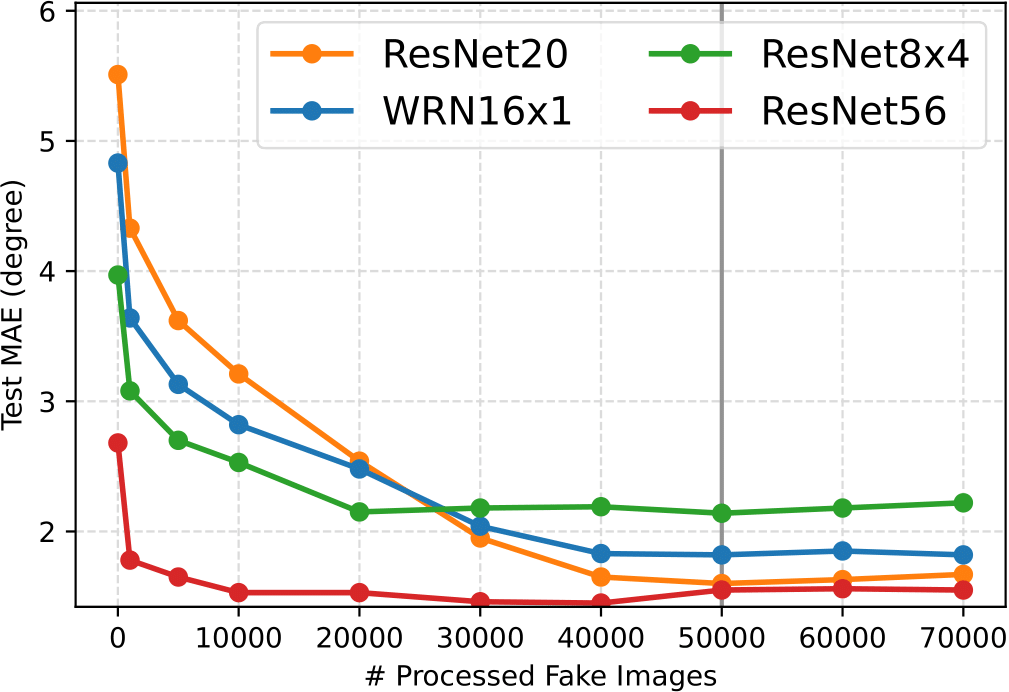}
		\label{fig:steeringangle_sensitivity_Mg}}
	\caption{\textbf{The effect of fake sample size $M^g$.} Gray lines indicate the $M^g$ we use in our experiments.}
	\label{fig:sensitivity_Mg}
\end{figure}

We conduct a sensitivity analysis on CIFAR-100 and Steering Angle to analyze the effect of different $M^g$. Compared with the main studies in Sections\ \ref{sec:exp_classification} and \ref{sec:exp_regression}, we only vary $M^g$ but keep other setups unchanged. The results of both datasets are shown in \Cref{fig:sensitivity_Mg}. We can conclude that more processed fake images stabilize the student models' performance without significantly decreasing the precision, which confirms the necessity of a large $M^g$. 

\subsubsection{Sensitivity analysis on the filtering quantile $\rho$}\label{sec:analysis_rho}

\begin{figure}[!h]
	\centering
	\subfloat[][CIFAR-100]{
		\includegraphics[width=0.4\textwidth]{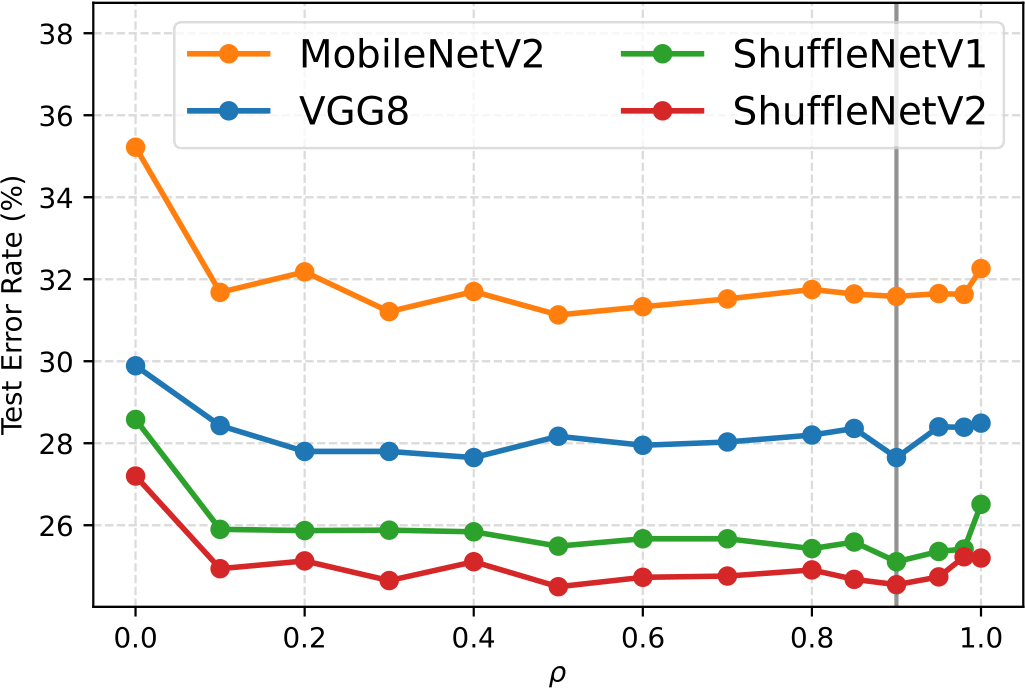}
		\label{fig:cifar_sensitivity_rho}}
	\subfloat[][Steering Angle]{
		\includegraphics[width=0.4\textwidth]{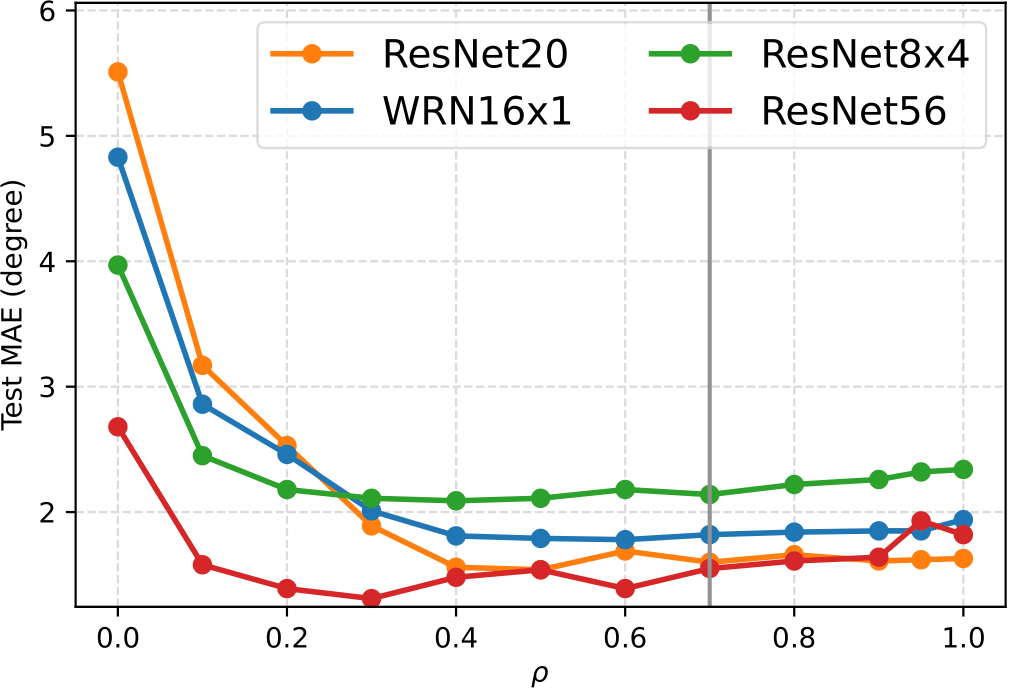}
		\label{fig:steeringangle_sensitivity_rho}}
	\caption{ {   \textbf{The effect of the filtering quantile $\rho$.} Gray lines indicate the $\rho$ we use in our experiments. $\rho=0$ corresponds to NOKD and $\rho=1$ implies no filtering. } }
	\label{fig:sensitivity_rho}
\end{figure}

{

The effect of $\rho$, another important hyper-parameter, is also analyzed in this section on CIFAR-100 and Steering Angle. Compared with the main studies in Sections\ \ref{sec:exp_classification} and \ref{sec:exp_regression}, we only vary $\rho$ but keep other setups unchanged, and visualize the results in \cref{fig:sensitivity_rho}. From \cref{fig:sensitivity_rho}(a), we can see that the performance of student models fluctuates in a small range when $\rho\in[0.2,0.98]$, and any $\rho$ that is not extremely small or large (i.e., close to 0 or 1) often leads to good performance on CIFAR-100. Similarly, from \cref{fig:sensitivity_rho}(b), we can see that a $\rho\in[0.3,0.8]$ often a good choice on the Steering Angle dataset. \cref{fig:sensitivity_rho} implies that the recommended $\rho$ in \Cref{sec:cGAN-KD_label_adjustment} (0.9 for classification and 0.7 for regression) often leads to desirable KD performance, and we generally don't need to carefully tune $\rho$ in practice. 

} 

\subsubsection{Sensitivity analysis on the teacher model's precision}\label{sec:analysis_teacher}
We also test the effect of the teacher model's precision. This study chooses neural networks with different precision as the teacher model in the cGAN-KD framework. \Cref{tab:sensitivity_teacher} shows that teachers with the highest accuracies often lead to the best KD results. Therefore, when implementing cGAN-KD, we should choose a teacher model with a precision as high as possible.

\begin{table}[!h]
	\centering
	\caption{\textbf{Sensitivity analysis of the effect of different teacher models in \textbf{M2} on CIFAR-100 and Steering Angle.} We show below the Top-1 test accuracy (\%) and test MAE (degree) of students under different teachers on CIFAR-100 and Steering Angle, respectively. The precision of different teachers on the test sets are shown in the parentheses.}
	\begin{adjustbox}{width=0.5\textwidth}
		\begin{tabular}{l|cccc}
			\hline \hline
			\multicolumn{5}{c}{\textbf{CIFAR-100}}\\
			\hline
			\diagbox{\textbf{Students}}{\textbf{Teachers}}  & \textbf{NOKD} & \begin{tabular}[l]{@{}c@{}} \textbf{WRN40$\times$2}\\ \textbf{(75.82)}\end{tabular}  & \begin{tabular}[l]{@{}c@{}} \textbf{ResNet18}\\ \textbf{(77.98)}\end{tabular} & \begin{tabular}[l]{@{}c@{}} \textbf{DenseNet121}\\ \textbf{(79.98)}\end{tabular} \\  \hline
			MobileNetV2 & 64.78  & 68.24  & 68.33  & 68.42  \\
			VGG8  & 70.11  & 71.63  & 72.29  & 72.35  \\
			ShuffleNetV1 & 71.42  & 73.83  & 74.43  & 74.89  \\
			ShuffleNetV2 & 72.80  & 74.63  & 74.94  & 75.45  \\
			\hline\hline
			\multicolumn{5}{c}{\textbf{Steering Angle}}\\
			\hline
			\diagbox{\textbf{Students}}{\textbf{Teachers}} & \textbf{NOKD} & \begin{tabular}[l]{@{}c@{}} \textbf{VGG8}\\ \textbf{(2.07)}\end{tabular} & \begin{tabular}[l]{@{}c@{}} \textbf{ResNet18}\\ \textbf{(1.71)}\end{tabular} & \begin{tabular}[l]{@{}c@{}} \textbf{VGG19}\\ \textbf{(1.06)}\end{tabular} \\
			\hline
			ResNet20 & 5.51  & 1.71  & 1.65  & 1.60  \\
			WRN16$\times$1 & 4.83  & 1.94  & 1.92  & 1.82  \\
			ResNet8x4 & 3.97  & 2.27  & 2.22  & 2.14  \\
			ResNet56 & 2.68  & 1.65  & 1.55  & 1.55  \\
			\hline\hline
		\end{tabular}%
	\end{adjustbox}
	\label{tab:sensitivity_teacher}%
\end{table}%

\subsubsection{Processing time and memory cost comparison}\label{sec:compare_cost}
\begin{figure}[!ht]
	\centering
	\subfloat[][CIFAR-100]{
		\includegraphics[width=0.76\textwidth]{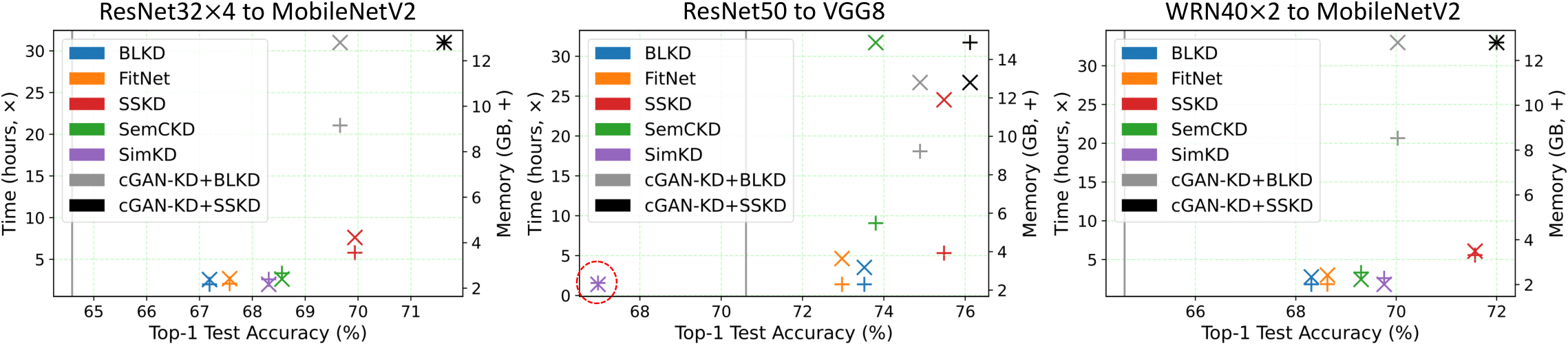}
		\label{fig:cifar_time_and_memory}}
	\\
	\subfloat[][Steering Angle]{
		\includegraphics[width=0.5\textwidth]{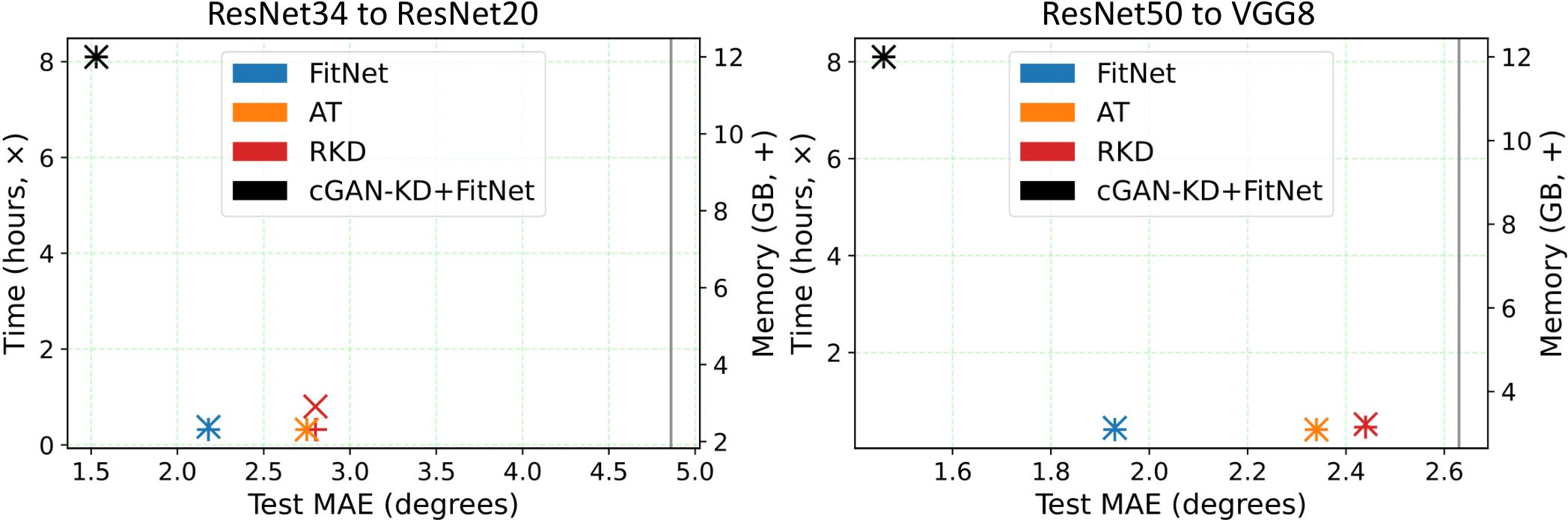}
		\label{fig:steeringangle_time_and_memory}}
	\subfloat[][UTKFace]{
		\includegraphics[width=0.5\textwidth]{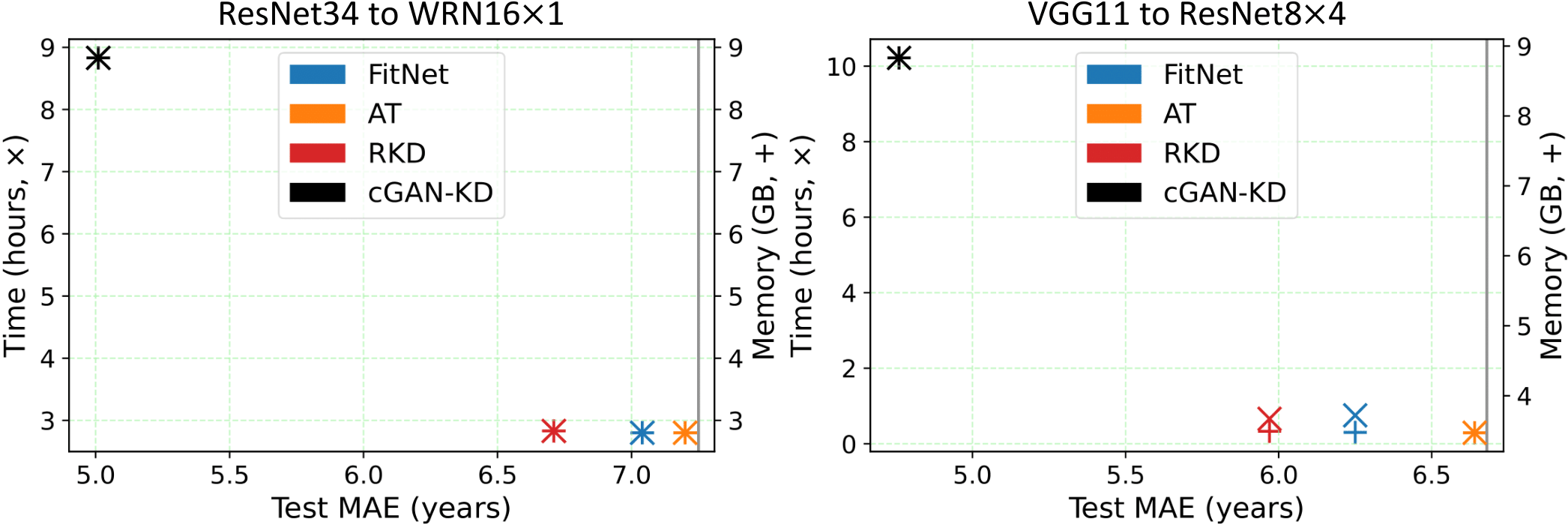}
		\label{fig:utkface_time_and_memory}}
	\caption{ { \textbf{Processing time (marked by ``$\times$") and memory cost (marked by ``$+$") comparison.} Gray lines represent the performance of NOKD.  } }
	\label{fig:time_and_memory}
\end{figure}

{

We also estimate the overall processing time and GPU memory required to implement cGAN-KD and some representative KD baselines on CIFAR-100, Steering Angle, and UTKFace in \Cref{fig:time_and_memory}, based on a single Tesla A100 GPU. Due to the implementation of cGANs, cDR-RS, and data augmentation, cGAN-KD-based methods need more processing time and GPU memory than other KD methods; however, cGAN-KD-based methods often lead to better KD performance. 


}

\section{Conclusion}\label{sec:cGAN-KD_conclusion}
As the first attempt, this paper proposes a unified knowledge distillation framework widely applicable for both classification and regression (with a scalar response) tasks. Fundamentally different from many existing knowledge distillation methods, we propose distilling and transferring knowledge from the teacher model to the student model through cGAN-generated samples, termed cGAN-KD. First, cGAN models are trained to generate a sufficient number of fake image samples. Then, high quality samples are obtained via subsampling and filtering procedures. Essentially, the knowledge is distilled by adjusting fake image labels utilizing the teacher model. Finally, the distilled knowledge is transferred to student models by training them on these knowledge-conveyed samples. The proposed framework is architecture-agnostic and it is compatible with existing state-of-the-art knowledge distillation models. We also derive the error bound of a student model trained in the cGAN framework for theoretical guidance. Extensive experiments demonstrate that the cGAN-KD incorporated methods can achieve state-of-the-art knowledge distillation performances for both classification and regression tasks.

\section*{Acknowledgments}
This work was supported by the Natural Sciences and Engineering Research Council of Canada (NSERC) under Grants CRDPJ 476594-14, RGPIN-2019-05019, and RGPAS2017-507965.

\bibliographystyle{model5-names}  
\bibliography{reference_cGAN_KD}

 \clearpage
 \newpage
 \appendix
 
 \section*{Supplementary Material}
 \addcontentsline{toc}{section}{Supplementary Material}

 \renewcommand{\thesection}{S.\arabic{section}} 
 \renewcommand{\thesubsection}{\thesection.\arabic{subsection}}
 \renewcommand\thefigure{\thesection.\arabic{figure}}
 \renewcommand\thetable{\thesection.\arabic{table}}
 \renewcommand{\theequation}{S.\arabic{equation}}
 \renewcommand{\thetheorem}{S.\arabic{theorem}} 
 \renewcommand{\thedefinition}{S.\arabic{definition}} 
 \renewcommand{\thelemma}{S.\arabic{lemma}} 
 \renewcommand{\theremark}{S.\arabic{remark}}
 
 \section{GitHub repository}\label{supp:cGAN-KD_codes}
 Please find some example codes for this paper at
 \begin{center}
 	\url{https://github.com/UBCDingXin/cGAN-based_KD}
 \end{center}

 \section{Resources for implementing cGANs, Subsampling and KD Methods}\label{supp:cGAN-KD_borrowed_codes}
 
 To implement TAKD, we refer to 
 \begin{center}
 	\url{https://github.com/imirzadeh/Teacher-Assistant-Knowledge-Distillation}
 \end{center}
 
 To implement SSKD, we refer to 
 \begin{center}
 	\url{https://github.com/xuguodong03/SSKD}
 \end{center}
 
 To implement ReviewKD, we refer to 
 \begin{center}
 	\url{https://github.com/dvlab-research/ReviewKD}
 \end{center}
 
 To implement SemCKD, we refer to 
 \begin{center}
 	\url{https://github.com/DefangChen/SemCKD}
 \end{center}
 
 To implement SimKD, we refer to 
 \begin{center}
 	\url{https://github.com/DefangChen/SimKD}
 \end{center}
 
 To implement other KD methods in our experiments, we refer to 
 \begin{center}
 	\url{https://github.com/HobbitLong/RepDistiller}
 \end{center}
 
 To implement BigGAN, we refer to
 \begin{center}
 	\url{https://github.com/ajbrock/BigGAN-PyTorch}
 \end{center}
 
 To implement CcGAN, we refer to
 \begin{center}
 	\url{https://github.com/UBCDingXin/improved_CcGAN}
 \end{center}
 
 To implement DiffAugment, we refer to 
 \begin{center}
 	\url{https://github.com/mit-han-lab/data-efficient-gans}
 \end{center}
 
 To implement cDR-RS, we refer to 
 \begin{center}
 	\url{https://github.com/UBCDingXin/cDR-RS}
 \end{center}

 \section{Proof of \Cref{thm:cGAN-KD_error_bound}}\label{supp:proof_cGAN-KD_error_bound}
 
 \begin{proof}
 	We first decompose $\mathcal{V}(\hat{f}_s)-\mathcal{V}(f^*) $ as follows
 	\begin{align}
 		& \mathcal{V}(\hat{f}_s)-\mathcal{V}(f^*) \nonumber\\
 		= & \mathcal{V}(\hat{f}_s) - \widehat{\mathcal{V}}(\hat{f}_s) + \widehat{\mathcal{V}}(\hat{f}_s) - \widehat{\mathcal{V}}(f_s^{\circ}) + \widehat{\mathcal{V}}(f_s^{\circ}) -
 		\mathcal{V}(f_s^{\circ})  + \mathcal{V}(f_s^{\circ}) - \mathcal{V}(f^*)\nonumber\\
 		& (\text{by } \widehat{\mathcal{V}}(\hat{f}_s) - \widehat{\mathcal{V}}(f_s^{\circ})\leq 0) \nonumber\\
 		\leq & \mathcal{V}(\hat{f}_s) - \widehat{\mathcal{V}}(\hat{f}_s) + \widehat{\mathcal{V}}(f_s^{\circ}) -
 		\mathcal{V}(f_s^{\circ}) + \mathcal{V}(f_s^{\circ}) - \mathcal{V}(f^*)\nonumber\\
 		\leq & 2\sup_{f_s\in\mathcal{F}_s}\left| \widehat{\mathcal{V}}(f_s) - \mathcal{V}(f_s) \right| + \mathcal{V}(f_s^{\circ}) - \mathcal{V}(f^*).
 		\label{eq:cGAN-KD_error_bound_first_decomposition}
 	\end{align}
 	The second term $\mathcal{V}(f_s^{\circ}) - \mathcal{V}(f^*)$ in Eq.\ \eqref{eq:cGAN-KD_error_bound_first_decomposition} is a non-negative number because the student model's hypothesis space $\mathcal{F}_s$ may not cover the optimal predictor $f^*$. In Eq.\ \eqref{eq:cGAN-KD_error_bound_first_decomposition}, $\sup_{f_s\in\mathcal{F}_s}\left| \widehat{\mathcal{V}}(f_s) - \mathcal{V}(f_s) \right|$ can be bounded as follows. Using the triangular inequality and \textbf{A4} (i.e., boundedness of $\mathcal{L}$) yields
 	
 	\begin{align}
 		& \sup_{f_s\in\mathcal{F}_s}\left| \widehat{\mathcal{V}}(f_s) - \mathcal{V}(f_s) \right| \nonumber\\
 		= & \sup_{f_s\in\mathcal{F}_s}\left| \frac{1}{N^r+M^g}\sum_{(\bm{x}_i,y_i)\in D_{\text{aug}}}\mathcal{L}(f_s(\bm{x}_i), y_i) - \mathbb{E}_{(\bm{x},y)\sim p_r(\bm{x},y)}\left[ \mathcal{L}(f_s(\bm{x}),y) \right] \right|\nonumber\\
 		& (\text{by the triangular inequality})\nonumber\\
 		\leq & \sup_{f_s\in\mathcal{F}_s}\left| \frac{1}{N^r+M^g}\sum_{(\bm{x}_i,y_i)\in D_{\text{aug}}}\mathcal{L}(f_s(\bm{x}_i), y_i) - \mathbb{E}_{(\bm{x},y)\sim p_{\theta}(\bm{x},y)}\left[ \mathcal{L}(f_s(\bm{x}),y) \right] \right|\nonumber\\ 
 		& + \sup_{f_s\in\mathcal{F}_s}\left| \mathbb{E}_{(\bm{x},y)\sim p_{\theta}(\bm{x},y)}\left[ \mathcal{L}(f_s(\bm{x}),y) \right] - \mathbb{E}_{(\bm{x},y)\sim p_r(\bm{x},y)}\left[ \mathcal{L}(f_s(\bm{x}),y) \right] \right|\nonumber\\
 		& (\text{by the Boundedness assumption \textbf{A4}})\nonumber\\
 		= & C_{\mathcal{L}} \sup_{f_s\in\mathcal{F}_s}\left| \frac{1}{N^r+M^g}\sum_{(\bm{x}_i,y_i)\in D_{\text{aug}}}\frac{1}{C_{\mathcal{L}}}\mathcal{L}(f_s(\bm{x}_i), y_i) - \mathbb{E}_{(\bm{x},y)\sim p_{\theta}(\bm{x},y)}\left[ \frac{1}{C_{\mathcal{L}}}\mathcal{L}(f_s(\bm{x}),y) \right] \right|\label{eq:cGAN-KD_error_bound_triangular_inequality_term_1}\\ 
 		& + \sup_{f_s\in\mathcal{F}_s}\left| \mathbb{E}_{(\bm{x},y)\sim p_{\theta}(\bm{x},y)}\left[ \mathcal{L}(f_s(\bm{x}),y) \right] - \mathbb{E}_{(\bm{x},y)\sim p_r(\bm{x},y)}\left[ \mathcal{L}(f_s(\bm{x}),y) \right] \right|
 		\label{eq:cGAN-KD_error_bound_triangular_inequality_term_2}.
 	\end{align}
 	For Eq.\ \eqref{eq:cGAN-KD_error_bound_triangular_inequality_term_1}, we apply the Rademacher bound \citep[Thm 7.7.1]{lafferty2010}, yielding that with at least probability $1-\delta$,
 	
 	\begin{align}
 		&  C_{\mathcal{L}} \sup_{f_s\in\mathcal{F}_s}\left| \frac{1}{N^r+M^g}\sum_{(\bm{x}_i,y_i)\in D_{\text{aug}}}\frac{1}{C_{\mathcal{L}}}\mathcal{L}(f_s(\bm{x}_i), y_i) - \mathbb{E}_{(\bm{x},y)\sim p_{\theta}(\bm{x},y)}\left[ \frac{1}{C_{\mathcal{L}}}\mathcal{L}(f_s(\bm{x}),y) \right] \right|\nonumber\\
 		\leq &   2C_{\mathcal{L}}\widehat{\mathcal{R}}_{N^r+M^g}(\mathcal{F}_s) + C_{\mathcal{L}}\sqrt{\frac{4}{N^r+M^g}\log\left( \frac{2}{\delta} \right)}.\label{eq:cGAN-KD_error_bound_triangular_inequality_term_1_bound}
 	\end{align}
 	Before we bound Eq.\ \eqref{eq:cGAN-KD_error_bound_triangular_inequality_term_2}, we first review the definition of the total variation distance \citep{gibbs2002choosing} between any two distributions $P$ and $Q$, i.e., 
 	
 	\begin{equation}
 		TV(P,Q) \coloneqq \frac{1}{2}\sup_{|g|\leq 1} \left|\int g \mathrm{d} P -  \int g \mathrm{d} Q   \right|  . \nonumber
 	\end{equation}
 	where $g$ is a measurable function.
 	Thus,
 	
 	\begin{align}
 		& TV(p_r, \theta p_r + (1-\theta) p_g^\rho)\nonumber\\
 		= & \frac{1}{2}\sup_{|g|\leq 1} \left| \mathbb{E}_{(\bm x, y)\sim p_\theta(\bm x, y)} \left[ g(\bm x, y)\right]- \mathbb{E}_{(\bm x, y)\sim p_r(\bm x, y)} \left[ g(\bm x, y)\right] \right| \nonumber\\
 		= & \frac{1}{2}\sup_{|g|\leq 1} \left\{(1- \theta) \cdot \left| \mathbb{E}_{(\bm x, y)\sim p_g^\rho(\bm x, y)} \left[ g(\bm x, y)\right] - \mathbb{E}_{(\bm x, y)\sim p_r(\bm x, y)} \left[ g(\bm x, y)\right] \right| \right\} \nonumber\\
 		= & (1-\theta) \cdot TV(p_r, p_g^\rho).\label{eq:cGAN-KD_error_bound_TV_property}
 	\end{align}
 	Since $f_s$ is measurable (by \textbf{A2}) and $\mathcal{L}$ is continuous, $\mathcal{L}(f_s(\bm{x}),y)$ is also measurable. Let $\mathcal{L}(f_s(\bm{x}),y) / C_{\mathcal{L}}$ be $g(\bm{x},y)$ in Eq.\ \eqref{eq:cGAN-KD_error_bound_TV_property}, then by \textbf{A3} (i.e., the distribution gap between $p_r$ and $p_g^\rho$), we have
 	
 	\begin{align}
 		&  \sup_{f_s\in\mathcal{F}_s}\left| \mathbb{E}_{(\bm{x},y)\sim p_{\theta}(\bm{x},y)}\left[ \mathcal{L}(f_s(\bm{x}),y) \right] - \mathbb{E}_{(\bm{x},y)\sim p_r(\bm{x},y)}\left[ \mathcal{L}(f_s(\bm{x}),y) \right] \right| \nonumber\\
 		= &  2C_{\mathcal{L}}(1-\theta)\left( C_{M1} + \Theta\left( \mathbb{E}_{(\bm{x},y)\sim p_r(\bm{x},y)}\left[ \mathcal{L}(f_t(\bm{x}), y) \right] \right) \right).\label{eq:cGAN-KD_error_bound_triangular_inequality_term_2_bound}
 	\end{align}
 	Combining Eqs.\ \eqref{eq:cGAN-KD_error_bound_triangular_inequality_term_1_bound} and \eqref{eq:cGAN-KD_error_bound_triangular_inequality_term_2_bound}, we can get
 	
 	\begin{align}
 		&  \sup_{f_s\in\mathcal{F}_s}\left| \widehat{\mathcal{V}}(f_s) - \mathcal{V}(f_s) \right| \nonumber\\
 		\leq & 2C_{\mathcal{L}}\widehat{\mathcal{R}}_{N^r+M^g}(\mathcal{F}_s) + C_{\mathcal{L}}\sqrt{\frac{4}{N^r+M^g}\log\left( \frac{2}{\delta} \right)} \nonumber\\
 		&  + 2C_{\mathcal{L}}(1-\theta)\left( C_{M1} + \Theta\left( \mathbb{E}_{(\bm{x},y)\sim p_r(\bm{x},y)}\left[ \mathcal{L}(f_t(\bm{x}), y) \right] \right) \right).\label{eq:cGAN-KD_error_bound_first_decomposition_1st_term_bound}
 	\end{align}
 	
 	Finally, incorporating Eq.\ \eqref{eq:cGAN-KD_error_bound_first_decomposition_1st_term_bound} into Eq.\ \eqref{eq:cGAN-KD_error_bound_first_decomposition}, we obtain the inequality (i.e., Eq.\ \eqref{eq:cGAN-KD_error_bound}) in Theorem \ref{thm:cGAN-KD_error_bound}, which completes the proof.
 	
 \end{proof}

 \section{Evolution of fake data and their distributions after applying \textbf{M1} and \textbf{M2}}\label{supp:dataset_evolution}
 \begin{figure}[!h]
 	\centering
 	\includegraphics[width=0.8\textwidth]{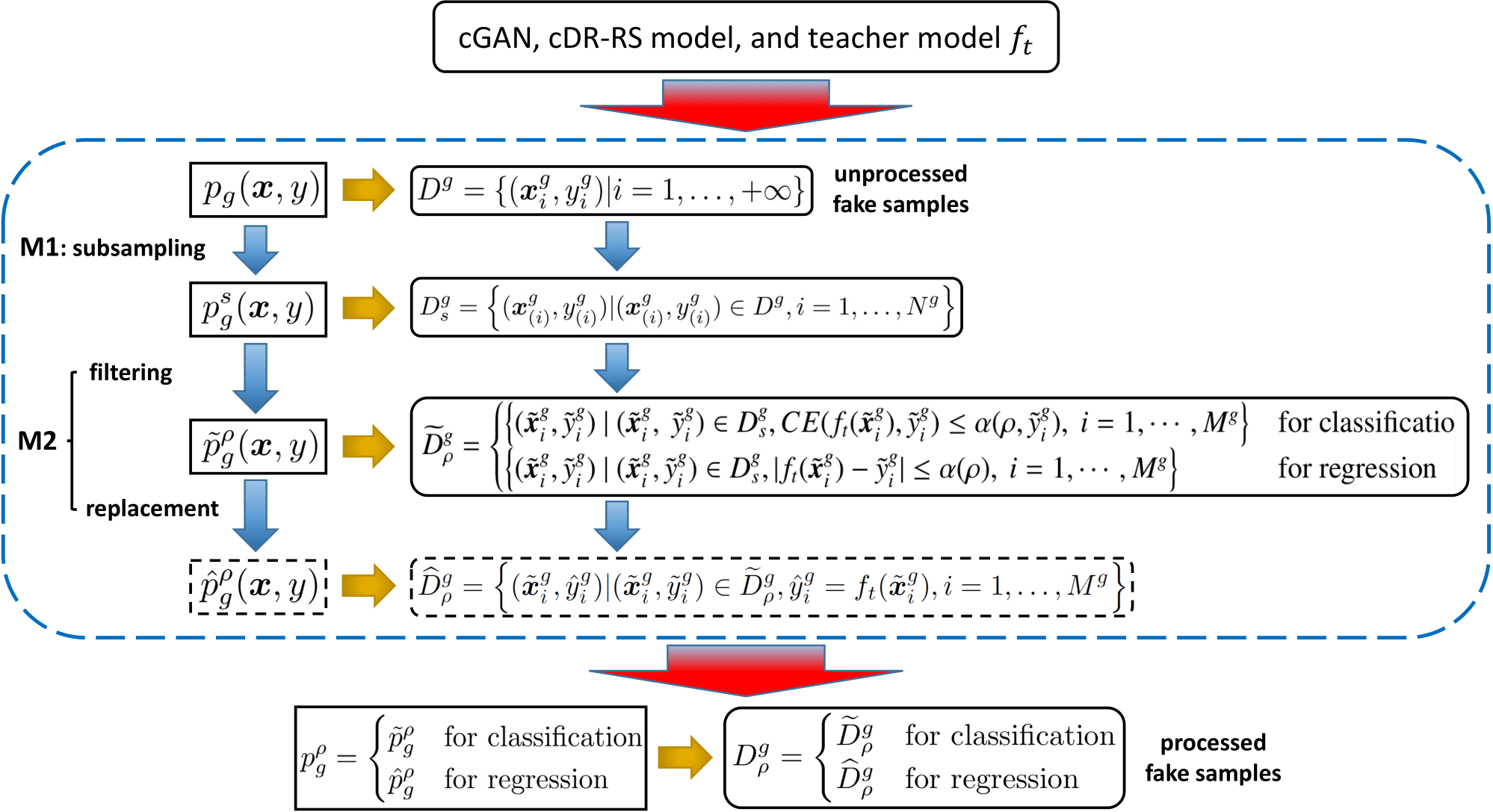}
 	\caption{\textbf{Evolution of fake data and their distributions after applying \textbf{M1} and \textbf{M2}.} Fake datasets are denoted by $D$ with or without hat, tilde, superscripts or subscripts, e.g., $D^g_s$. The density functions of fake data's distributions are denoted by $p(\bm{x},y)$ with or without hat, tilde, superscripts or subscripts, e.g., $p_g^s(\bm{x},y)$. The replacement sub-module is enabled only for regression problems. The number of samples after filtering is smaller than or equal to that after subsampling ($M^g\leq N^g$). The filtering threshold $\alpha$ is related to a hyper-parameter $\rho$ or a class label. It is defined as the $\rho$-th quantile of errors between predicted and assigned labels of fake images. For classification, we have one $\alpha$ for one class, so that we conduct filtering within each class. For regression, we have a global $\alpha$ to filter all fake images. Please refer to \Cref{sec:cGAN-KD_label_adjustment} for the definition of $\alpha$ and the selection of $\rho$.}
 	\label{fig:cGAN-KD_evolution_of_fake_distribution}
 \end{figure}

 \section{More details of experiments on the CIFAR-100 dataset}\label{supp:cGAN-KD_details_of_cifar10}
 
 We first train some popular neural networks from scratch on the training set of CIFAR-100. Following \citet{xu2020knowledge}, all these neural networks are trained for 240 epochs with the SGD optimizer, initial learning rate 0.05 (but 0.01 for ShuffleNet and MobileNetV2; decays at epoch 150, 180, and 210 with factor 0.1), weight decay $5\times 10^{-4}$, and batch size 64. The number of parameters, inference speed, and Top-1/5 test accuracies of these neural networks are shown in \Cref{tab:cifar100_performance_all}. MobileNetV2, ResNet20, VGG8, WRN40$\times$1, ShuffleNetV1, ResNet8x4, ShuffleNetV2, and WRN16$\times$2 are chosen as students due to their low Top-1 test accuracies. DenseNet121 is chosen as the teacher model $f_t$ in cGAN-KD due to its highest Top-1 test accuracy. Some of these neural networks' checkpoints are used in the implementation of existing KD methods. 
 
 To implement BLKD and TAKD, we set $\lambda_{KD}=0.5$ and $T=5$ following \citet{ruffy2019state}. In TAKD, the precision of a good TA model is usually the average of those of the teacher and student models \citep{mirzadeh2020improved}. Based on this principle, the TA models for all teacher-student pairs are chosen and shown in \Cref{tab:cifar100_teacher_assistants_for_TAKD}. To implement SSKD, we follow the default setups in \citet{xu2020knowledge} and the corresponding GitHub repository. To implement ReviewKD, we follow the default setups in \citet{chen2021distilling} and the corresponding GitHub repository. To implement SemCKD \citep{chen2021cross,wang2022semckd} and SimKD \citep{chen2022knowledge}, we use their official codes and default setups. We use the setups suggested by CRD \citep{tian2019contrastive} to implement other KD methods.
 
 As for the implementation of cGAN-KD-based methods, we first train a BigGAN for 2,000 epochs with a batch size 512. DiffAugment \citep{zhao2020differentiable} is enabled in the GAN training with the strongest transformation combination (Color + Translation + Cutout). Then, we implement cDR-RS to subsample fake images with the setups suggested by \citet{ding2021subsampling}. We choose DenseNet121 to filter fake images with $\rho=0.9$. We generate 1000 processed fake images for each class (100,000 fake samples in total), which are then used to augment the training set. 
 
 {\setlength{\parindent}{0cm} \textbf{cGAN-KD alone and cGAN-KD + X (excluding SSKD):}} We first initialize student models with their NOKD checkpoints. Then, we train students on the augmented dataset for 240 epochs with the SGD optimizer, initial learning rate 0.01 (decays at epoch 150, 180, and 210 with factor 0.1), weight decay $5\times 10^{-4}$, and batch size 128. 
 
 {\setlength{\parindent}{0cm} \textbf{cGAN-KD+SSKD:}} Initialize student models with their SSKD checkpoints. Then, we train students on the augmented dataset for 240 epochs with the SGD optimizer, initial learning rate 0.01 (decays at epoch 150, 180, and 210 with factor 0.1), weight decay $1\times 10^{-4}$, and batch size 128. 
 
 We repeat all KD experiments four times and report the average results. The random seeds used in this experiment are shown in \Cref{tab:cifar100_seeds}. Please refer to our codes for more detailed experimental setup.

 \begin{table}[!h]
 	\centering
 	\caption{Test accuracy, number of parameters, and inference speed of different neural networks on CIFAR-100. The inference speed is measured by processing 10,000 images with batch size 64 on a single RTX 2080TI. Since DenseNet121 has the highest Top-1 test accuracy, it is chosen as the teacher model $f_t$ in cGAN-KD. }
 	\begin{adjustbox}{width=0.7\textwidth}
 		\begin{tabular}{l|rccc}
 			\hline\hline
 			\textbf{Models} & \textbf{\# Params} & \begin{tabular}[l]{@{}c@{}} \textbf{Inference speed}\\ \textbf{(images/second)}\end{tabular}  & \begin{tabular}[l]{@{}c@{}} \textbf{Test Accuracy $\uparrow$}\\ \textbf{(Top-1)}\end{tabular} & \begin{tabular}[l]{@{}c@{}} \textbf{Test Accuracy $\uparrow$}\\ \textbf{(Top 5)}\end{tabular} \\
 			\hline
 			VGG8 & 3,965,028 & 5217  & 70.11  & 90.82  \\
 			VGG11 & 9,277,284 & 5011  & 71.64  & 90.49  \\
 			VGG13 & 9,462,180 & 4571  & 74.85  & 92.32  \\
 			VGG19 & 20,086,692 & 3778  & 73.88  & 91.77  \\
 			ResNet20 & 278,324 & 4542  & 68.83  & 91.10  \\
 			ResNet56 & 861,620 & 3374  & 72.67  & 92.41  \\
 			ResNet110 & 1,736,564 & 2322  & 73.27  & 92.71  \\
 			ResNet8x4 & 1,233,540 & 4692  & 72.77  & 93.10  \\
 			ResNet32x4 & 7,433,860 & 2340  & 79.11  & 94.64  \\
 			ResNet18 & 11,220,132 & 2192  & 77.98  & 94.04  \\
 			ResNet34 & 21,328,292 & 2384  & 78.94  & 94.65  \\
 			ResNet50 & 23,705,252 & 1876  & 79.51  & 95.02  \\
 			WRN16$\times$2 & 703,284 & 4821  & 73.02  & 92.94  \\
 			WRN40$\times$1 & 569,780 & 3956  & 71.35  & 92.02  \\
 			WRN40$\times$2 & 2,255,156 & 3895  & 75.82  & 93.53  \\
 			\textbf{DenseNet121} & 7,048,548 & 1421  & \textbf{79.98} & 95.04  \\
 			DenseNet169 & 12,643,172 & 1216  & 79.54  & 95.19  \\
 			DenseNet201 & 18,277,220 & 1021  & 79.89  & 95.48  \\
 			DenseNet161 & 26,681,188 & 763   & 79.60  & 95.10  \\
 			ShuffleNetV1 & 949,258 & 1509  & 71.42  & 91.04  \\
 			ShuffleNetV2 & 1,355,528 & 3249  & 72.80  & 91.45  \\
 			MobileNetV2 & 812,836 & 3414  & 64.78  & 88.47  \\
 			\hline\hline
 		\end{tabular}%
 	\end{adjustbox}
 	\label{tab:cifar100_performance_all}%
 \end{table}%

 \begin{table}[!h]
 	\centering
 	\caption{The teacher assistants for TAKD in the CIFAR-100 experiment in \Cref{sec:exp_classification}. The teacher assistants' performance is often in the middle of the corresponding teacher-student combination.}
 	\begin{adjustbox}{width=1\textwidth}
 		\begin{tabular}{ccc|ccc|ccc}
 			\hline\hline
 			Teacher & \textbf{Assistant} & Student & Teacher & \textbf{Assistant} & Student & Teacher & \textbf{Assistant} & Student \\
 			\hline
 			{ResNet110} & {WRN40$\times$1} & {ResNet20} & {WRN40$\times$2} & {VGG8} & {MobileNetV2} & {DenseNet121} & {ResNet56} & {MobileNetV2} \\
 			{ResNet32x4} & {ResNet110} &  {ResNet20} &  {WRN40$\times$2} &  {ResNet8x4} &  {VGG8} &  {DenseNet121} &  {VGG13} &  {ResNet20} \\
 			{VGG13} &  {VGG11} &  {VGG8} &  {ResNet32x4} &  {VGG11} &  {MobileNetV2} &  {DenseNet121} &  {VGG13} &  {VGG8} \\
 			{VGG19} &  {VGG11} &  {VGG8} &  {ResNet32x4} &  {VGG13} &  {VGG8} &  {DenseNet121} &  {WRN40$\times$2} &  {ResNet8x4} \\
 			{WRN40$\times$2} &  {WRN16$\times$2} &  {WRN40$\times$1} &  {ResNet32x4} &  {WRN40$\times$2} &  {ShuffleNetV1} &  {DenseNet121} &  {WRN40$\times$2} &  {ShuffleNetV1} \\
 			{ResNet32x4} &  {ResNet110} &  {ResNet8x4} &  {ResNet32x4} &  {WRN40$\times$2} &  {ShuffleNetV2} &  {DenseNet121} &  {WRN40$\times$2} &  {ShuffleNetV2} \\
 			&       &       &  {ResNet50} &  {ResNet56} &  {MobileNetV2} &       &       &  \\
 			&       &       &  {ResNet50} &  {VGG13} &  {VGG8} &       &       &  \\
 			&       &       &  {ResNet50} &  {WRN40$\times$2} &  {ShuffleNetV1} &       &       &  \\
 			\hline\hline
 		\end{tabular}%
 	\end{adjustbox}
 	\label{tab:cifar100_teacher_assistants_for_TAKD}%
 \end{table}%
 
 \begin{table}[!h]
 	\centering
 	\caption{The seed setups in different folders of our GitHub repository for the CIFAR-100 experiment. \texttt{./BigGAN} and \texttt{./make\_fake\_datasets} train the BigGAN model and implement the subsampling and filtering modules. \texttt{./RepDistiller} implements all other KD methods except TAKD, SSKD, ReviewKD, SemCKD, and SimKD. }
 	\begin{adjustbox}{width=0.5\textwidth}
 		\begin{tabular}{l|cccc}
 			\hline\hline
 			Folder & Run 1 & Run 2 & Run 3 & Run 4 \\
 			\hline
 			\texttt{./BigGAN} & 0     & 1     & 2     & 3 \\
 			\texttt{./make\_fake\_datasets} & 2021  & 2022  & 2023  & 2024 \\
 			\hline
 			\texttt{./RepDistiller} & 2021  & 2022  & 2023  & 2024 \\
 			\texttt{./TAKD}  & 2021  & 2022  & 2023  & 2024 \\
 			\texttt{./SSKD}  & 2021  & 2022  & 2023  & 2024 \\
 			\texttt{./ReviewKD} & 2021  & 2022  & 2023  & 2024 \\
 			\texttt{./SemCKD} & 2021  & 2022  & 2023  & 2024 \\
 			\texttt{./SimKD} & 2021  & 2022  & 2023  & 2024 \\
 			\hline\hline
 		\end{tabular}%
 	\end{adjustbox}
 	\label{tab:cifar100_seeds}%
 \end{table}%

 \section{More details of experiments on the ImageNet-100 dataset}\label{supp:cGAN-KD_details_of_imagenet}
 
 We first train some popular neural networks from scratch on the training set of ImageNet-100. Following \citet{xu2020knowledge}, all these neural networks are trained for 240 epochs with the SGD optimizer, initial learning rate 0.05 (but 0.01 for ShuffleNet and MobileNetV2; decays at epoch 150, 180, and 210 with factor 0.1), weight decay $5\times 10^{-4}$, and batch size 128. The number of parameters, inference speed, and Top-1/5 test accuracies of these neural networks are shown in \Cref{tab:imagenet100_performance_all}. ResNet20, WRN40$\times$1, WRN16$\times$2, ResNet8x4, ResNet56, MobileNetV2, ShuffleNetV1, and VGG8 are chosen as students due to their low Top-1 test accuracies. DenseNet161 is chosen as the teacher model $f_t$ in cGAN-KD due to its highest Top-1 test accuracy. Some of these neural networks' checkpoints are used in the implementation of existing KD methods. 
 
 To implement SSKD, we follow the default setups in \citet{xu2020knowledge} and the corresponding Github repository. We use the setups suggested by CRD \citep{tian2019contrastive} to implement other KD methods. Due to limited computational resources, we test fewer KD methods and teacher-student pairs in this experiment.
 
 As for the implementation of cGAN-KD-based methods, we train a BigGAN for 96,000 iterations with the BigGAN-deep architecture \citep{brock2018large} and a batch size 1,024 following \citet{ding2021subsampling}. We borrow the checkpoint of this BigGAN from \citet{ding2021subsampling}. DiffAugment \citep{zhao2020differentiable} is enabled in the GAN training with the strongest transformation combination (Color + Translation + Cutout). Then, we implement cDR-RS to subsample fake images with the setups suggested by \citet{ding2021subsampling}. We choose DenseNet161 to filter fake images with $\rho=0.9$. We generate 1000 processed fake images for each class (100,000 fake samples in total), which are then used to augment the training set. 
 
 {\setlength{\parindent}{0cm} \textbf{cGAN-KD alone and cGAN-KD + X (excluding SSKD):}} We first initialize student models with their NOKD checkpoints. Then, we train students on the augmented dataset for 240 epochs with the SGD optimizer, initial learning rate 0.01 (decays at epoch 150, 180, and 210 with factor 0.1), weight decay $5\times 10^{-4}$, and batch size 256. 
 
 {\setlength{\parindent}{0cm} \textbf{cGAN-KD+SSKD:}} Initialize student models with their SSKD checkpoints. Then, we train students on the augmented dataset for 240 epochs with the SGD optimizer, initial learning rate 0.01 (decays at epoch 150, 180, and 210 with factor 0.1), weight decay $1\times 10^{-4}$, and batch size 256. 
 
 The random seeds used in this experiment are shown in \Cref{tab:imagenet100_seeds}. Please refer to our codes for more detailed experimental setup.
 
 \begin{table}[!h]
 	\centering
 	\caption{Test accuracy, number of parameters, and inference speed of different neural networks on ImageNet-100. The inference speed is measured by processing 10,000 images with batch size 64 on a single RTX 2080TI. Since DenseNet161 has the highest Top-1 test accuracy, it is chosen as the teacher model $f_t$ in cGAN-KD.}
 	\begin{adjustbox}{width=0.7\textwidth}
 		\begin{tabular}{l|rccc}
 			\hline\hline
 			\textbf{Models} & \textbf{\# Params} & \begin{tabular}[l]{@{}c@{}} \textbf{Inference speed}\\ \textbf{(images/second)}\end{tabular}  & \begin{tabular}[l]{@{}c@{}} \textbf{Test Accuracy $\uparrow$}\\ \textbf{(Top-1)}\end{tabular} & \begin{tabular}[l]{@{}c@{}} \textbf{Test Accuracy $\uparrow$}\\ \textbf{(Top 5)}\end{tabular} \\
 			\hline
 			VGG8  & 13,455,460 & 1546  & 77.36  & 92.93  \\
 			VGG11 & 18,767,716 & 1690  & 81.72  & 94.29  \\
 			VGG13 & 18,952,612 & 1169  & 83.03  & 95.18  \\
 			VGG19 & 29,577,124 & 864   & 83.41  & 95.32  \\
 			ResNet20 & 278,660 & 2042  & 65.25  & 87.47  \\
 			ResNet56 & 861,956 & 1733  & 73.20  & 91.72  \\
 			ResNet110 & 1,736,900 & 1428  & 75.31  & 92.03  \\
 			ResNet8x4 & 1,234,212 & 2098  & 73.19  & 91.38  \\
 			ResNet32x4 & 7,434,532 & 1855  & 81.97  & 94.71  \\
 			ResNet18 & 11,221,476 & 1945  & 80.22  & 93.65  \\
 			ResNet34 & 21,329,636 & 1826  & 81.54  & 94.47  \\
 			ResNet50 & 23,706,596 & 1576  & 83.33  & 95.28  \\
 			WRN16$\times$2 & 703,652 & 2057  & 72.09  & 91.06  \\
 			WRN40$\times$1 & 570,148 & 1901  & 69.98  & 89.81  \\
 			WRN40$\times$2 & 2,255,524 & 1900  & 77.67  & 93.37  \\
 			DenseNet121 & 7,050,020 & 1278  & 83.22  & 95.14  \\
 			DenseNet169 & 12,644,644 & 1091  & 82.93  & 94.86  \\
 			DenseNet201 & 18,278,692 & 960   & 82.66  & 95.15  \\
 			\textbf{DenseNet161} & 26,683,396 & 732 & \textbf{84.37} & 95.48 \\
 			ShuffleNetV1 & 950,338 & 1426  & 75.02  & 92.07  \\
 			ShuffleNetV2 & 1,356,608 & 1751  & 77.12  & 92.72  \\
 			MobileNetV2 & 812,836 & 1805  & 74.55  & 91.81  \\
 			\hline\hline
 		\end{tabular}%
 	\end{adjustbox}
 	\label{tab:imagenet100_performance_all}%
 \end{table}%

 \begin{table}[!h]
 	\centering
 	\caption{The seed setups in different folders of our GitHub repository for the ImageNet-100 experiment. \texttt{./make\_fake\_datasets} generates fake images via a pre-trained BigGAN from \citet{ding2021subsampling} and implements the subsampling and filtering modules. \texttt{./RepDistiller} implements all other KD methods except SSKD, SemCKD, and SimKD.}
 	\begin{adjustbox}{width=0.7\textwidth}
 		\begin{tabular}{l|ccccc}
 			\hline\hline
 			& \texttt{./make\_fake\_datasets} & \texttt{./RepDistiller} & \texttt{./SSKD} & \texttt{./SemCKD} & \texttt{./SimKD} \\
 			\hline
 			Seed  & 2021  & 2021  & 0  & 2021  & 2021 \\
 			\hline\hline
 		\end{tabular}%
 	\end{adjustbox}
 	\label{tab:imagenet100_seeds}%
 \end{table}%

 \section{More details of experiments on the Steering Angle dataset}\label{supp:cGAN-KD_details_of_steeringangle}
 
 In this experiment, all networks are trained for 350 epochs with the SGD optimizer, initial learning rate 0.01 (decays at epoch 150 and 250 with factor 0.1), weight decay $5\times 10^{-4}$, and batch size 128. 
 
 To determine teacher and student models, we first train some popular networks from scratch and their test errors are shown in Table \ref{tab:steeringangle_performance_all}. ResNet20, MobileNetV2, WRN16$\times$1, ResNet8x4, WRN40$\times$1, ShuffleNetV1, and ResNet56 are chosen as students due to their high test MAEs. VGG19 is chosen as the teacher model $f_t$ in cGAN-KD due to its lowest test MAE. Some of these neural networks' checkpoints are used in the implementation of existing KD methods. 
 
 We use the SAGAN architecture \citep{zhang2019self}, SVDL+ILI, and hinge loss in the CcGAN training. The CcGAN model is trained for 20,000 iterations with batch size 512, $\kappa=1123.760$, and $\sigma=0.028$. DiffAugment \citep{zhao2020differentiable} is enabled in the GAN training with the strongest transformation combination (Color + Translation + Cutout). The rest setups are consistent with the official implementation of CcGANs \citep{ding2020continuous}. We follow most setups in \citet{ding2021subsampling} to implement cDR-RS, but we disable the filtering scheme in cDR-RS. The reason is the label adjustment module in cGAN-KD functions similarly to the filtering scheme, and the filtering scheme in cDR-RS often leads to much longer training time and sampling time. We choose VGG19 to filter fake images with $\rho=0.7$. We generate 50,000 processed fake images, which are then used to augment the training set. 
 
 The random seeds used in this experiment are shown in \Cref{tab:steeringangle_seeds}. Please refer to our codes for more detailed training and testing setups.
 
 \begin{table}[!h]
 	\centering
 	\caption{Test MAE, number of parameters, and inference speed of different neural networks on Steering Angle. The inference speed is measured by processing 10,000 images with batch size 64 on a single RTX 2080TI.}
 	\begin{adjustbox}{width=0.5\textwidth}
 		\begin{tabular}{l|rcc}
 			\hline\hline
 			\textbf{Models} & \textbf{\# Params} & \begin{tabular}[l]{@{}c@{}} \textbf{Inference speed}\\ \textbf{(images/second)}\end{tabular}  & \begin{tabular}[l]{@{}c@{}} \textbf{Test MAE $\downarrow$}\\ \textbf{(degree)}\end{tabular} \\
 			\hline
 			VGG8  & 5,228,033 & 2745  & 2.07  \\
 			VGG11 & 10,540,289 & 2615  & 1.38  \\
 			VGG13 & 10,725,185 & 2172  & 1.43  \\
 			VGG16 & 16,037,441 & 2137  & 1.49  \\
 			\textbf{VGG19} & 21,349,697 & 2061  & \textbf{1.06} \\
 			ResNet20 & 570,609 & 2541  & 5.51  \\
 			ResNet56 & 1,153,905 & 2102  & 2.68  \\
 			ResNet110 & 2,028,849 & 1635  & 2.87  \\
 			ResNet8x4 & 1,604,641 & 2737  & 3.97  \\
 			ResNet32x4 & 7,804,961 & 2135  & 2.69  \\
 			ResNet18 & 11,700,929 & 2615  & 1.71  \\
 			ResNet34 & 21,809,089 & 2117  & 1.41  \\
 			ResNet50 & 24,814,657 & 1478  & 1.61  \\
 			WRN16$\times$1 & 473,281 & 2636  & 4.83  \\
 			WRN16$\times$2 & 1,022,017 & 2750  & 4.51  \\
 			WRN40$\times$1 & 862,145 & 2247  & 3.87  \\
 			WRN40$\times$2 & 2,573,889 & 2317  & 3.31  \\
 			DenseNet121 & 7,737,537 & 1412  & 1.37  \\
 			DenseNet169 & 13,595,841 & 1212  & 1.54  \\
 			DenseNet201 & 19,335,361 & 1046  & 1.55  \\
 			DenseNet161 & 27,858,721 & 849   & 1.35  \\
 			ShuffleNetV1 & 1,611,487 & 2172  & 3.67  \\
 			ShuffleNetV2 & 2,044,125 & 2158  & 4.86  \\
 			MobileNetV2 & 3,144,961 & 2278  & 3.06  \\
 			\hline\hline
 		\end{tabular}%
 	\end{adjustbox}
 	\label{tab:steeringangle_performance_all}%
 \end{table}%

 \begin{table}[!h]
 	\centering
 	\caption{The seed setups in different folders of our GitHub repository for the Steering Angle experiment. \texttt{./cGAN-KD} generates fake images from a CcGAN and implements the subsampling and label adjustment modules. \texttt{./cGAN-KD} also implements cGAN-KD. \texttt{./RepDistiller} implements feature-based KD methods.}
 	\begin{adjustbox}{width=0.3\textwidth}
 		\begin{tabular}{l|cc}
 			\hline\hline
 			& \texttt{./cGAN-KD} & \texttt{./RepDistiller} \\
 			\hline
 			Seed  & 2020  & 2020  \\
 			\hline\hline
 		\end{tabular}%
 	\end{adjustbox}
 	\label{tab:steeringangle_seeds}%
 \end{table}%

 \section{More details of experiments on the UTKFace dataset}\label{supp:cGAN-KD_details_of_utkface}
 
 In this experiment, all networks are trained for 350 epochs with the SGD optimizer, initial learning rate 0.01 (decays at epoch 150 and 250 with factor 0.1), weight decay $5\times 10^{-4}$, and batch size 128. 
 
 To determine teacher and student models, we first train some popular networks from scratch and their test errors are shown in Table \ref{tab:utkface_performance_all}. ResNet20, MobileNetV2, WRN16$\times$1, ResNet8x4, WRN40$\times$1, ShuffleNetV1, and ResNet56 are chosen as students due to their high test MAEs. VGG19 is chosen as the teacher model $f_t$ in cGAN-KD due to its lowest test MAE. Some of these neural networks' checkpoints are used in the implementation of existing KD methods. 
 
 We use the SAGAN architecture \citep{zhang2019self}, SVDL+ILI, and hinge loss in the CcGAN training. The CcGAN model is trained for 40,000 iterations with batch size 512, $\kappa=900$, and $\sigma=0.043$. DiffAugment \citep{zhao2020differentiable} is enabled in the GAN training with the strongest transformation combination (Color + Translation + Cutout). The rest setups are consistent with the official implementation of CcGANs \citep{ding2020continuous}. We follow most setups in \citet{ding2021subsampling} to implement cDR-RS, but we disable the filtering scheme in cDR-RS. The reason is the label adjustment module in cGAN-KD functions similarly to the filtering scheme, and the filtering scheme in cDR-RS often leads to much longer training time and sampling time. We choose VGG11 to filter fake images with $\rho=0.7$. We generate 60,000 processed fake images, which are then used to augment the training set. 
 
 The random seeds used in this experiment are shown in \Cref{tab:utkface_seeds}. Please refer to our codes for more detailed training and testing setups.
 
 \begin{table}[!h]
 	\centering
 	\caption{Test MAE, number of parameters, and inference speed of different neural networks on UTKFace. The inference speed is measured by processing 10,000 images with batch size 64 on a single RTX 2080TI.}
 	\begin{adjustbox}{width=0.5\textwidth}
 		\begin{tabular}{l|rcc}
 			\hline\hline
 			\textbf{Models} & \textbf{\# Params} & \begin{tabular}[l]{@{}c@{}} \textbf{Inference speed}\\ \textbf{(images/second)}\end{tabular}  & \begin{tabular}[l]{@{}c@{}} \textbf{Test MAE $\downarrow$}\\ \textbf{(year)}\end{tabular} \\
 			\hline
 			VGG8  & 5,228,033 & 2745  & 5.28  \\
 			\textbf{VGG11} & 10,540,289 & 2615 & \textbf{5.12} \\
 			VGG13 & 10,725,185 & 2172  & 5.16  \\
 			VGG16 & 16,037,441 & 2137  & 5.25  \\
 			VGG19 & 21,349,697 & 2061  & 5.32  \\
 			ResNet20 & 570,609 & 2541  & 6.87  \\
 			ResNet56 & 1,153,905 & 2102  & 7.06  \\
 			ResNet110 & 2,028,849 & 1635  & 6.77  \\
 			ResNet8x4 & 1,604,641 & 2737  & 6.68  \\
 			ResNet32x4 & 7,804,961 & 2135  & 6.36  \\
 			ResNet18 & 11,700,929 & 2615  & 5.62  \\
 			ResNet34 & 21,809,089 & 2117  & 5.29  \\
 			ResNet50 & 24,814,657 & 1478  & 5.91  \\
 			WRN16$\times$1 & 473,281 & 2636  & 7.25  \\
 			WRN16$\times$2 & 1,022,017 & 2750  & 6.69  \\
 			WRN40$\times$1 & 862,145 & 2247  & 6.70  \\
 			WRN40$\times$2 & 2,573,889 & 2317  & 6.86  \\
 			DenseNet121 & 7,737,537 & 1412  & 5.34  \\
 			DenseNet169 & 13,595,841 & 1212  & 5.65  \\
 			DenseNet201 & 19,335,361 & 1046  & 5.61  \\
 			DenseNet161 & 27,858,721 & 849   & 5.42  \\
 			ShuffleNetV1 & 1,611,487 & 2172  & 7.03  \\
 			ShuffleNetV2 & 2,044,125 & 2158  & 6.87  \\
 			MobileNetV2 & 3,144,961 & 2278  & 7.16  \\
 			\hline\hline
 		\end{tabular}%
 	\end{adjustbox}
 	\label{tab:utkface_performance_all}%
 \end{table}%

 \begin{table}[!h]
 	\centering
 	\caption{The seed setups in different folders of our GitHub repository for the UTKFace experiment. \texttt{./cGAN-KD} generates fake images from a CcGAN and implements the subsampling and label adjustment modules. \texttt{./cGAN-KD} also implements cGAN-KD. \texttt{./RepDistiller} implements feature-based KD methods.}
 	\begin{adjustbox}{width=0.3\textwidth}
 		\begin{tabular}{l|cc}
 			\hline\hline
 			& \texttt{./cGAN-KD} & \texttt{./RepDistiller} \\
 			\hline
 			Seed  & 2020  & 2020  \\
 			\hline\hline
 		\end{tabular}%
 	\end{adjustbox}
 	\label{tab:utkface_seeds}%
 \end{table}%

\end{document}